\documentclass[10pt,journal,compsoc]{IEEEtran}

\ifCLASSOPTIONcompsoc
  \usepackage[nocompress]{cite}
\else
  \usepackage{cite}
\fi

\usepackage[utf8]{inputenc} 
\usepackage[T1]{fontenc}    
\usepackage{url}            
\usepackage{booktabs}       
\usepackage{amsfonts}       
\usepackage{nicefrac}       
\usepackage{microtype}      
\usepackage{xcolor}         
\usepackage{colortbl}
\usepackage{amsmath}
\usepackage{physics}
\usepackage{float}
\usepackage{multirow, tikz, makecell, verbatim, booktabs, array}
\usepackage{wrapfig}
\usepackage{cleveref}
\usepackage{graphicx}
\usepackage{subfigure}
\usepackage{kotex}
\usepackage{float} 

\usepackage[normalem]{ulem}
\useunder{\uline}{\ul}{}
\newcommand{\etal}{\emph{et al}.\@ }
\newcolumntype{C}{>{\centering\arraybackslash}p{4.5em}}
\newcolumntype{D}{>{\centering\arraybackslash}p{6em}}

\makeatletter
\newcommand\notsotiny{\@setfontsize\notsotiny{6.31415}{7.1828}}
\makeatother

\usepackage{amssymb}
\usepackage{pifont}
\newcommand{\cmark}{\textcolor{green}{\ding{51}}}%
\newcommand{\xmark}{\textcolor{red}{\ding{55}}}%

\def\genbox#1#2#3#4#5#6{
    \leavevmode\raise#4bp\hbox to#5bp{\vrule height#5bp depth0bp width0bp
    \pdfliteral{q .5 w \csname #2COLOR\endcsname\space RG
                       \csname #3PDF\endcsname{#5}{#6} S Q
             \ifx1#1 q \csname #2COLOR\endcsname\space rg 
                       \csname #3PDF\endcsname{#5}{#6} f Q\fi}\hss}}
\def\trianbox   #1#2{\genbox{#1}{#2}  {trian}    {0}   {6}    {3}}

\begin{document}

\title{Bridging Implicit and Explicit Geometric Transformation for Single-Image View Synthesis}

\author{Byeongjun~Park*,
        Hyojun~Go*,
        and~Changick~Kim,~\IEEEmembership{Senior~Member,~IEEE}
\IEEEcompsocitemizethanks{\IEEEcompsocthanksitem B. Park and C. Kim are with the School
of Electrical Engineering, Korea Advanced Institute of Science and Technology (KAIST), Daejeon 34141, Republic of Korea. (E-mail: \{pbj3810, changick\}@kaist.ac.kr)
\IEEEcompsocthanksitem H. Go is with Twelvelabs, Seoul, Republic of Korea. \protect\\
(E-mail: gohyojun15@gmail.com)
\IEEEcompsocthanksitem B. Park and H. Go contributed equally to this work.
\IEEEcompsocthanksitem Corresponding author: Changick Kim.}}%

\newcommand{\todoblue}[1]{\textcolor{blue}{\textbf{#1}}}
\newcommand{\hyojun}[1]{\todoblue{\textbf{hyojun:} #1}}



\IEEEtitleabstractindextext{%
\begin{abstract}
    Creating novel views from a single image has achieved tremendous strides with advanced autoregressive models, as unseen regions have to be inferred from the visible scene contents.
    Although recent methods generate high-quality novel views, synthesizing with only one explicit or implicit 3D geometry has a trade-off between two objectives that we call the "seesaw" problem: 1) preserving reprojected contents and 2) completing realistic out-of-view regions.
    Also, autoregressive models require a considerable computational cost. 
    In this paper, we propose a single-image view synthesis framework for mitigating the seesaw problem while utilizing an efficient non-autoregressive model.
    Motivated by the characteristics that explicit methods well preserve reprojected pixels and implicit methods complete realistic out-of-view regions, we introduce a loss function to complement two renderers.
    Our loss function promotes that explicit features improve the reprojected area of implicit features and implicit features improve the out-of-view area of explicit features.
    With the proposed architecture and loss function, we can alleviate the seesaw problem, outperforming autoregressive-based state-of-the-art methods and generating an image $\approx$100 times faster.
    We validate the efficiency and effectiveness of our method with experiments on RealEstate10K and ACID datasets.
\end{abstract}
\begin{IEEEkeywords}
Single-Image View Synthesis, Transformer.
\end{IEEEkeywords}
}

\maketitle

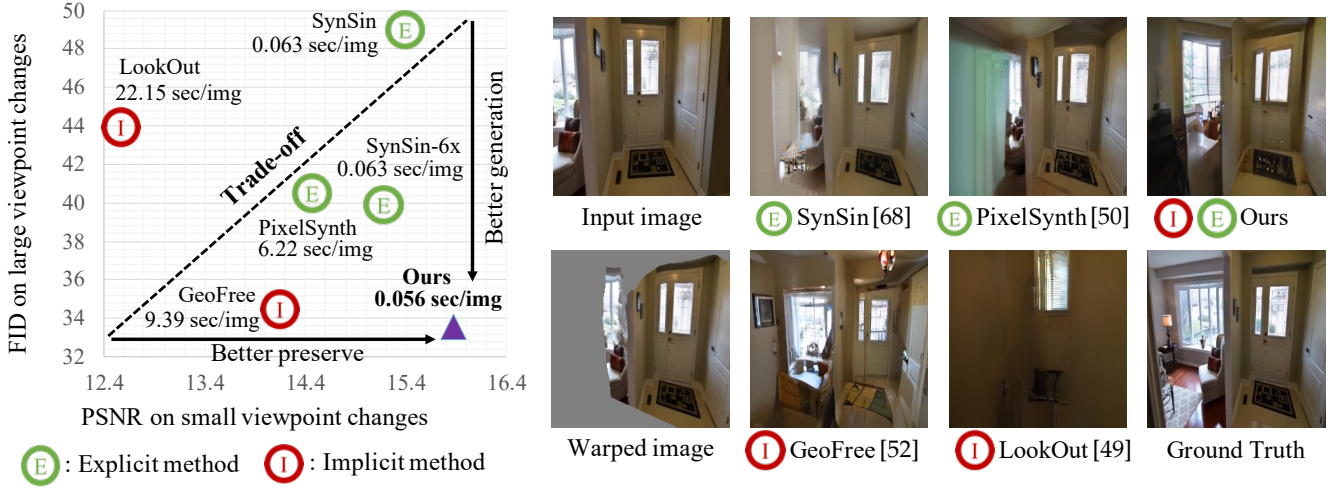
\begin{figure*}[ht]
    \centering
    \begin{tikzpicture}
    \draw (0, 0) node[inner sep=0] {\includegraphics[width=\textwidth]{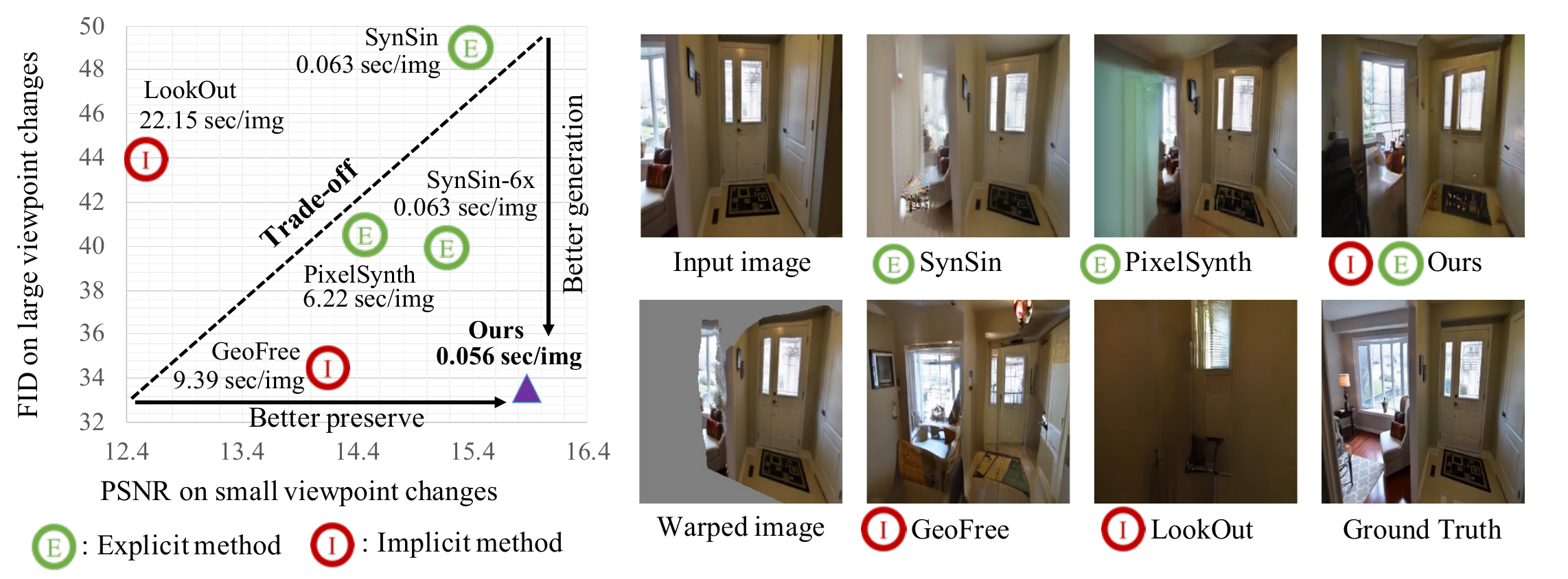}};
    \draw (2.8, 0.31) node {\small{\cite{wiles2020synsin}}};
    \draw (2.9, -2.79) node {\small{\cite{rombach2021geometry}}};
    \draw (5.7 , 0.31) node {\small{\cite{rockwell2021pixelsynth}}};
    \draw (5.73 , -2.79) node {\small{\cite{ren2022look}}};
    \end{tikzpicture}

    \vspace{-0.5cm}
    
    \caption{\textbf{Seesaw problem of explicit and implicit methods.} Explicit methods well preserve warped contents but sacrifice to fill unseen pixels ($\uparrow$ PSNR on small view change, $\uparrow$ FID on large view change). Implicit methods amply fill unseen pixels but fall short of preserving seen contents ($\downarrow$ PSNR on small view change, $\downarrow$ FID on large view change). Note that LookOut~\cite{ren2022look} focuses on long-term novel view synthesis, so it has degenerated in balancing both objectives. Our proposed framework alleviates this seesaw problem and generates an image faster than the state-of-the-art methods.}
    \vspace{-0.4cm}
    \label{fig:seesaw_problem}   
\end{figure*}

\ifCLASSOPTIONcompsoc
\IEEEraisesectionheading{\section{Introduction}\label{sec:introduction}}
\else
\section{Introduction}
\label{sec:introduction}
\fi

\IEEEPARstart{S}{ingle-image} view synthesis is the task of generating novel view images from a given single image~\cite{chen2019monocular,liu2021infinite,watson2023novel,yu2021pixelnerf,wiles2020synsin,hu2021worldsheet,rombach2021geometry,rockwell2021pixelsynth,ren2022look, li2022infinitenature}.
It can enable the movement of the camera from a photograph and bring an image to 3D, which is significant for various computer vision applications such as image editing and animating.
To perform the realistic view synthesis in these applications, we can expect that the novel view image has to consist of existing objects and unseen new objects from the reference viewpoint.
Therefore, for synthesizing high-quality novel views, the following two goals should be considered: 1) preserving 3D transformed seen contents of a reference image—represented by the horizontal axis in Fig.~\ref{fig:seesaw_problem}, and 2) generating semantically compatible pixels for filling the unseen region—illustrated by the vertical axis in Fig.~\ref{fig:seesaw_problem}.
To achieve two goals, explicit and implicit methods have been proposed.

With the recent success of differentiable geometric transformation methods~\cite{bao2019depth, niklaus2020softmax}, explicit methods~\cite{chen2019monocular, wiles2020synsin, hou2021novel, zhou2016view, liu2021infinite, li2022infinitenature} leverage such 3D inductive biases to guide the view synthesis network to preserve 3D transformed contents, and various generative models are applied to complete the unseen regions.
Since explicit methods directly learn the pixel correspondence between the reference image and the novel view image, convolutional neural networks are often utilized to extract the discriminative local scene representation.
Therefore, they can produce high-quality novel view images in small view changes, where the content of the reference viewpoint still occupies a large portion.
However, the global scene representation is not sufficiently trained, resulting in the image quality having been degraded for large view changes due to a lack of ability to generate pixels of the unseen region.
To deal with this problem, outpainting with the autoregressive model is exploited to fill unseen regions~\cite{rockwell2021pixelsynth}, but generating photo-realistic images remains a challenge for explicit methods.

On the other side, implicit methods~\cite{tatarchenko2016multi, rombach2021geometry,ren2022look} less enforce 3D inductive biases and let the model learn the required 3D geometry for view synthesis.
Based on the powerful autoregressive transformer~\cite{esser2021taming}, recent implicit methods~\cite{rombach2021geometry,ren2022look} learn the 3D geometry from reference images and camera parameters by leveraging the global self-attention for all pixels.
Implicitly learned 3D geometry allows the model to synthesize diverse and realistic novel view images, but the only use of the global scene representation has degenerated to represent detailed correspondences between the reference image and the novel view image.
Due to the lack of discriminative local scene representation and the reduction in 3D inductive biases, implicit methods fail to preserve the reprojected contents.

To sum up, previous single-image view synthesis methods suffer from a trade-off between two objectives: 1) preserve seen contents and 2) generate semantically compatible unseen regions.
Figure~\ref{fig:seesaw_problem} highlights the trade-off that explicit methods well preserve seen contents while sacrificing the generation of unseen regions and vice versa for implicit methods.
We term this trade-off the \textit{seesaw problem} and advocate for a synergistic approach that combines the strengths of both explicit and implicit methods.
Moreover, recent methods often depend on autoregressive models, which generate individual pixels sequentially.
This sequential generation hampers view synthesis speed, impeding real-time image animation.
Therefore, we refocus on an efficient non-autoregressive model for single-image view synthesis.

In this paper, we present a non-autoregressive framework for alleviating the seesaw problem.
To this end, we aim to design the architecture that encodes both global and local scene representation and the loss function that bridges implicit and explicit geometric transformations.
Specifically, our encoder is built on recent transformer architectures for point cloud representations~\cite{lee2019set, park2021fast}, allowing the network to consider the three-dimensional camera movement by encoding the 3D spatial locations of each point.
We also design two parallel renderers that explicitly or implicitly learn geometric transformations from point cloud representations.
To bridge explicit and implicit transformations, we propose a novel loss function that motivates explicit features to improve seen pixels of implicit features and implicit features to improve unseen pixels of explicit features.
Interestingly, we observe that the proposed loss makes two renderers embed discriminative features and allows the model to use both renderers in a balanced way.
With the proposed architecture and the loss function, we can merge the pros of both explicit and implicit methods, alleviating the seesaw problem.

We validate the effectiveness of our framework with experiments on the indoor dataset RealEstate10K~\cite{zhou2018stereo}, the outdoor dataset ACID~\cite{liu2021infinite} and the urban-scene dataset SWORD~\cite{khakhulin2022stereo}.
We summarize the following contributions:

\begin{itemize}
    \item We design a two-branch attention module for point cloud representation, enabling the network to consider 3D camera movement and represent both detailed semantics and the entire scene context.
    \item We design two parallel render blocks that explicitly or implicitly learn geometric transformations and introduce a novel loss function to combine the strengths of two render blocks.
    \item Our non-autoregressive framework facilitates real-time applications and is over $\approx$100 times faster than existing autoregressive methods.
    \item Experiments on different view changes in indoor and outdoor datasets show that our method alleviates the seesaw problem and performs superior results in preserving seen contents and generating unseen contents.
\end{itemize}
\begin{figure*}[t!]
  \centering
  \includegraphics[width=\linewidth]{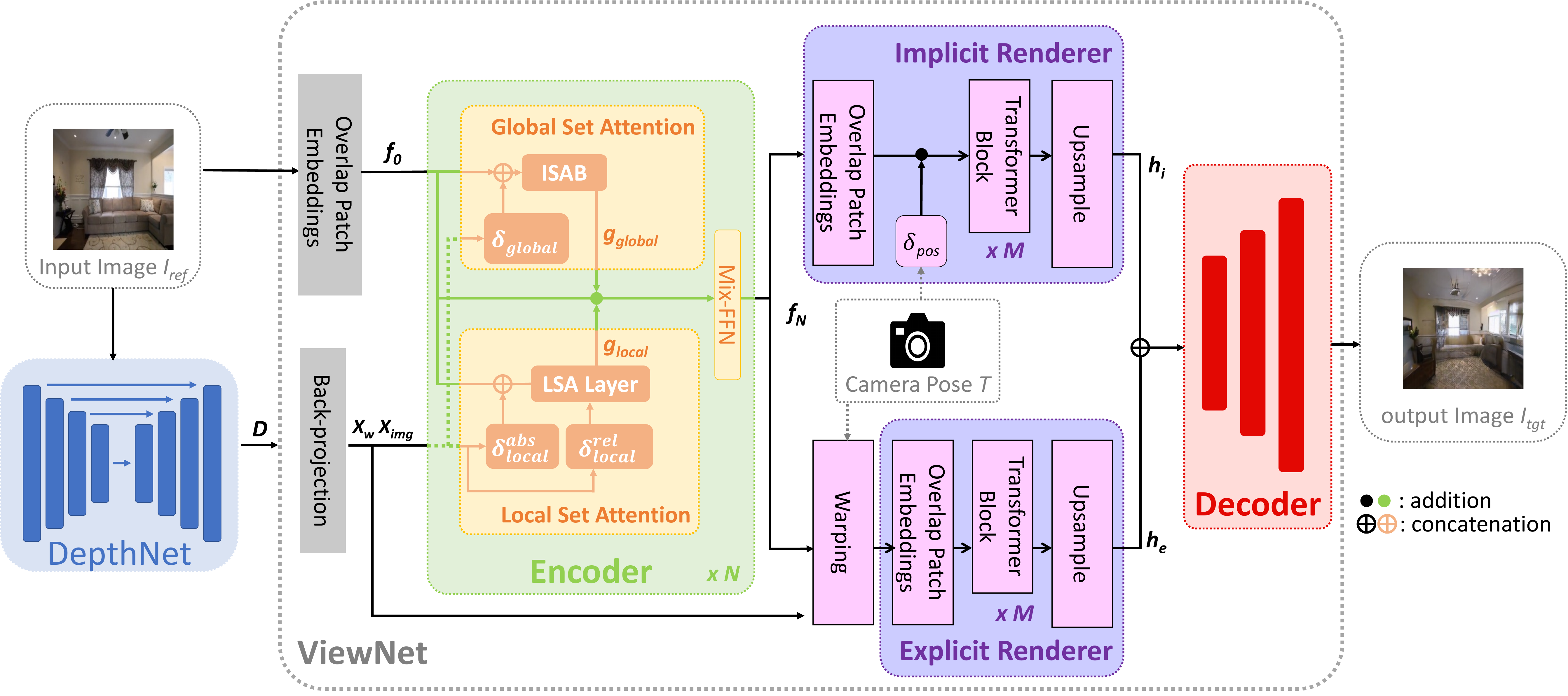}
  \vspace{-0.7cm}
  \caption{
  \textbf{An overview of network architecture.} Our network takes a reference image $I_{ref}$ and a relative camera pose $T$ as inputs. The depth estimation network (DepthNet) first predicts a depth map $D$, and the view synthesis network (ViewNet) generates a target image $I_{tgt}$ from $I_{ref}$, $D$ and $T$. Specifically, $D$ is used for calculating the 3D world coordinate $X_{w}$ and the normalized image coordinate $X_{img}$ at the reference viewpoint, which are passed through various positional encoding layers in the encoder (e.g., $\delta_{global}$, $\delta_{local}^{abs}$ and $\delta_{local}^{rel}$) to provide the scene structure representations. Encoded features $f_{N}$ are then transformed by both Implicit Renderer and Explicit Renderer with $T$. Finally, two transformed feature map, $h_{i}$ and $h_{e}$, are concatenated to generate $I_{tgt}$ by the decoder.
  }
  \label{fig_network}
  \vspace{-0.5cm}
\end{figure*}

\section{Related Works}
\subsection{Neural Representations for Novel View Synthesis}

Novel view synthesis aims to generate novel view images given multiple images from different viewpoints of a scene.
To consider the arbitrary viewpoints in three-dimension, early methods often utilize the multi-view geometry combined with image-based rendering methods to synthesize realistic novel views~\cite{chen1993view,debevec1996modeling,gortler1996lumigraph,levoy1996light,seitz2006comparison,zitnick2004high,debevec1998efficient}.
Recently, deep neural networks have been used for rendering~\cite{meshry2019neural,hedman2018deep,martin2018lookingood,novotny2019perspectivenet} and several representation for view synthesis, such as multi-plane image~\cite{srinivasan2019pushing, zhou2018stereo,flynn2019deepview}, point cloud~\cite{aliev2020neural}, depth~\cite{srinivasan2017learning}, and voxel~\cite{lombardi2019neural,olszewski2019transformable,sitzmann2019deepvoxels}.
However, these methods focus on modeling the scene surface with geometric proxies, limiting the network to representing simple scenes with low geometric complexity.
On the other hand, coordinate-based neural representations~\cite{chen2019learning, mescheder2019occupancy, mildenhall2020nerf} have achieved outstanding results in modeling the complex scene as implicit scene representations.
Specifically, Neural Radiance Fields (NeRF)~\cite{mildenhall2020nerf} have been proposed to model the scene as a continuous volumetric field with neural networks, rendering photo-realistic novel views of complex scenes with higher resolution.
Some methods incorporate the geometric prior with NeRF representations, accelerating the inference speed by avoiding unnecessary ray sampling in the empty space~\cite{xu2022point} and enabling the novel view synthesis from limited image sets~\cite{wang2021ibrnet, chen2021mvsnerf, yu2021pixelnerf}.
Note that these methods combine explicit and implicit neural scene representations to model the view-dependent effects only for seen contents and interpolate given multiple images to synthesize novel views.
In contrast, our method aims at extrapolating a single image by bridging explicit and implicit geometric transformations to achieve high-quality performance on both preserving seen contents and generating unseen contents.

\subsection{Single-Image View Synthesis}

Single-image view synthesis is more challenging than general novel view synthesis since a single input image is only available~\cite{chen2019monocular,liu2021infinite,yu2021pixelnerf,wiles2020synsin,hu2021worldsheet,rombach2021geometry,rockwell2021pixelsynth,ren2022look, li2022infinitenature}. 
Explicit methods directly inject 3D inductive biases into models.
For example, SynSin~\cite{wiles2020synsin} uses 3D point cloud features with estimated depth from the model, projects to novel viewpoints, and refines unseen pixels with recent generative models~\cite{brock2018large}.
SynSin works well for small viewpoint changes but has poor performance on large viewpoint changes due to the lack of generating unseen pixels.
To deal with this issue, PixelSynth~\cite{rockwell2021pixelsynth} exploits the autoregressive outpainting model~\cite{razavi2019generating} with 3D point cloud representation. 
Also, InfNat~\cite{liu2021infinite} and InfZero~\cite{li2022infinitenature} propose the temporal autoregressive model for natural scenes.
Despite using the slow autoregressive model, it cannot generate unseen pixels as well. 
For an implicit method, Rombach~\etal\cite{rombach2021geometry} propose a powerful autoregressive transformer.
By less enforcing 3D inductive biases, this approach can generate realistic view synthesis and complete the unseen region without explicit 3D geometry. 
However, it takes a long inference time due to the autoregressive model and fails to preserve seen contents of a reference image.
We bridge these implicit and explicit methods as a non-autoregressive architecture, which can outperform autoregressive approaches with fast inference.

\subsection{Transformer for Point Cloud Representation}

The transformer and self-attention have brought a breakthrough in natural language processing~\cite{vaswani2017attention,devlin2018bert} and computer vision~\cite{dosovitskiy2020image}.
Inspired by this success, transformer and self-attention networks have been widely applied for point cloud recognition tasks and achieved remarkable performance gains.
Early methods utilize global attention for all of the point clouds, resulting in a large amount of computation and inapplicable for large-scale 3D point cloud~\cite{xie2018attentional, liu2019point2sequence, yang2019modeling}.
Lee \etal~\cite{lee2019set} propose the SetTransformer module suitable for point cloud due to permutation-invariant, which uses inducing point methods and reduces computational complexity from quadratic to linear in the number of elements.
Also, local attention methods are utilized to enable scalability~\cite{park2021fast,zhao2021point,guo2021pct}.
Notably, among local attention methods, Fast Point Transformer~\cite{park2021fast} uses voxel hashing-based architecture and achieves both remarkable performance and computational efficiency.
Global attention may dilute important content by excessive noises as most neighbors are less relevant, and local attention may not have sufficient context due to their scope.
Therefore, our approaches use both global and local attention to deal with 3D point cloud representation.

\section{Methodology}

\subsection{Overview}

Given a reference image $I_{ref}$ and a relative camera pose $T$, the goal of single-image view synthesis is to create a target image $I_{tgt}$ with keeping visible contents of $I_{ref}$ and completing realistic out-of-view pixels.
To achieve this, we focus on mitigating the seesaw problem between explicit and implicit methods in terms of the network architecture and the loss function.
Figure~\ref{fig_network} describes an overview of our network architecture.
The network consists of two sub-networks, the depth estimation network (\textbf{DepthNet}) and the view synthesis network (\textbf{ViewNet}).
Note that the pre-trained DepthNet generates depth map $D$, which is used for ViewNet.
We design a simple view synthesis network built on architectural innovations of recent transformer models.
Specifically, we exploit 3D point cloud representation to consider the relationship between the geometry-aware camera pose information and the input image.

\subsection{Depth Estimation Network (DepthNet)}
We train the depth estimation network for explicit 3D geometry since ground-truth depths are not available.
Following Monodepth2~\cite{godard2017unsupervised}, our DepthNet is trained in a self-supervised manner from monocular video sequences.
Because a ground-truth relative camera pose between images is available, we substitute the pose estimation network with the ground-truth relative pose.
Given a reference image $I_{ref}$, a ground-truth image for the target viewpoint $I_{gt}$, and output depth maps $D_{ref}$ of $I_{ref}$, we formulate the loss function $L_{depth}$ with the reprojection loss $L_{rep}$ and the edge-aware depth smoothness loss $L_{sm}$ to train the DepthNet as:
\begin{equation}
L_{rep} = \frac{\alpha}{2} \left( {1 - SSIM} \right) + \left( 1 - \alpha \right){ ||I_{ref} - I'_{gt} ||}_1,
\end{equation}
\begin{equation}
L_{sm} = {|\partial_{x}D^{*}_{ref}|e^{-|\partial_{x}I_{ref}|} + |\partial_{y}D^{*}_{ref}|e^{-|\partial_{y}I_{ref}|}},
\end{equation}
\begin{equation}
L_{depth} = L_{rep} + \lambda_{sm} L_{sm}.
\end{equation}
Here, $I'_{gt}$ is a geometrically transformed $I_{gt}$ to the reference viewpoint, and we fix $\alpha = 0.85$ and $\lambda_{sm} = 1e^{-3}$.
Also, $SSIM$ is the structure similarity \cite{wang2004image} between $I_{ref}$ and $I'_{gt}$, and $D^{*}_{ref}$ is the mean-normalized inverse depth from \cite{wang2018learning}.
Similar to Monodepth2, we train the network on minimum reprojection loss across the neighbor frames and apply the auto-masking.
After training DepthNet, we fix it during training ViewNet.

\subsection{Encoder: Global and Local Set Attention Block}

The encoder aims to extract scene representations from a feature point cloud of a reference image.
To deal with point clouds,  we design a Global and Local Set Attention (GLSA) block which simultaneously extracts overall contexts and detailed semantics.
For efficient input size of transformers, $I_{ref} \in \mathbb{R}^{{H} \times {W} \times 3}$ is encoded into $f_{0} \in \mathbb{R}^{\frac{H}{4} \times \frac{W}{4} \times C}$ by an overlapping patch embedding~\cite{xie2021segformer}, where $C$ denotes the channel dimension.
Then, the homogeneous coordinates $p$ of a pixel in $f_0$ are mapped into normalized image coordinates $X_{img}$ as $X_{img}^p=K^{-1}_{\downarrow}p$, where $K_{\downarrow}$ denotes the camera intrinsic matrix of $f_0$.
Finally, 3D world coordinates of $p$ are calculated with depth map $D$ as $X_{w}^p=D(p)X_{img}^p$.
Our global set attention block takes 3D world coordinates to encode the spatial position of each point, and our local set attention block takes both homogeneous coordinates and 3D world coordinates to efficiently encode relative positions.
Note that the channel dimension $C$ is set to 256, and we set all positional encoding layers embedded into 32 channels.
Our encoder architecture is $N$ stacked GLSA block, and $i$-th GLSA block receives $f_{i-1}, X_{img}$ and $X_{w}$ and outputs $f_i$ with Mix-FFN~\cite{xie2021segformer}.
Specifically, the $i$-th output point feature of the encoder $f_{i}$ formulated with a global set attention $g_{global}^{i}$ and a local set attention $g_{local}^{i}$ as:
\begin{equation}
\begin{gathered}
    f_{i} = \text{MLP}(\text{GELU}(\text{CONV}_{3 \times 3}(\text{MLP}(X_{i})))) + X_{i}, \\
    \text{where} \quad X_{i} = f_{i-1} + g_{global}^{i} + g_{local}^{i}.
\end{gathered}
\end{equation}

\subsubsection{Global Set Attention} 
We utilize Induced Set Attention Block (ISAB)~\cite{lee2019set} to extract global set attention between the feature point clouds.
We first define a multi-head attention block (MAB) as:
\begin{equation}
\label{eq:mab}
    \text{MAB}(X, Y) = \text{LayerNorm}(H + rFF(H)),
\end{equation}
\begin{equation}
    H = \text{LayerNorm}(X + \text{Attention}(X, Y, Y)),
\end{equation}
\begin{equation}
    \text{Attention}(Q, K, V) = \text{Softmax}(\frac{QK^{T}}{\sqrt{d_{head}}})V,
\end{equation}
where rFF denotes any row-wise feed-forward layer, and we use the same rFF in \cite{lee2019set}.
Then, using two MABs and $m$ inducing points $I \in \mathbb{R}^{m \times C}$, we define the induced set attention block for $n$ points as:

\begin{equation}
\begin{gathered}
    \text{ISAB}_{m}(X) = \text{MAB}(X, G) \in \mathbb{R}^{n \times C}, \\
    \text{where} \quad G = \text{MAB}(I, X) \in \mathbb{R}^{m \times C}.
\end{gathered}
\end{equation}
With the positional encoder $\delta_{global}$ and the vector concatenation operator $\oplus$, global attention of $i$-th GLSA bock is represented as: 
\begin{equation}
    g_{global}^i(p) = ISAB(f_{i}(p) \oplus \delta_{global}(X_{w}^p)).
\end{equation}
Note that we compute the global set attention for $n = \frac{H}{4} \cdot \frac{W}{4}$ points and fix the number of inducing points $m$ to $32$.

\begin{figure}[t!]
    \centering
    \subfigure[Coordinate Decomposition]{\includegraphics[width=0.215\textwidth]{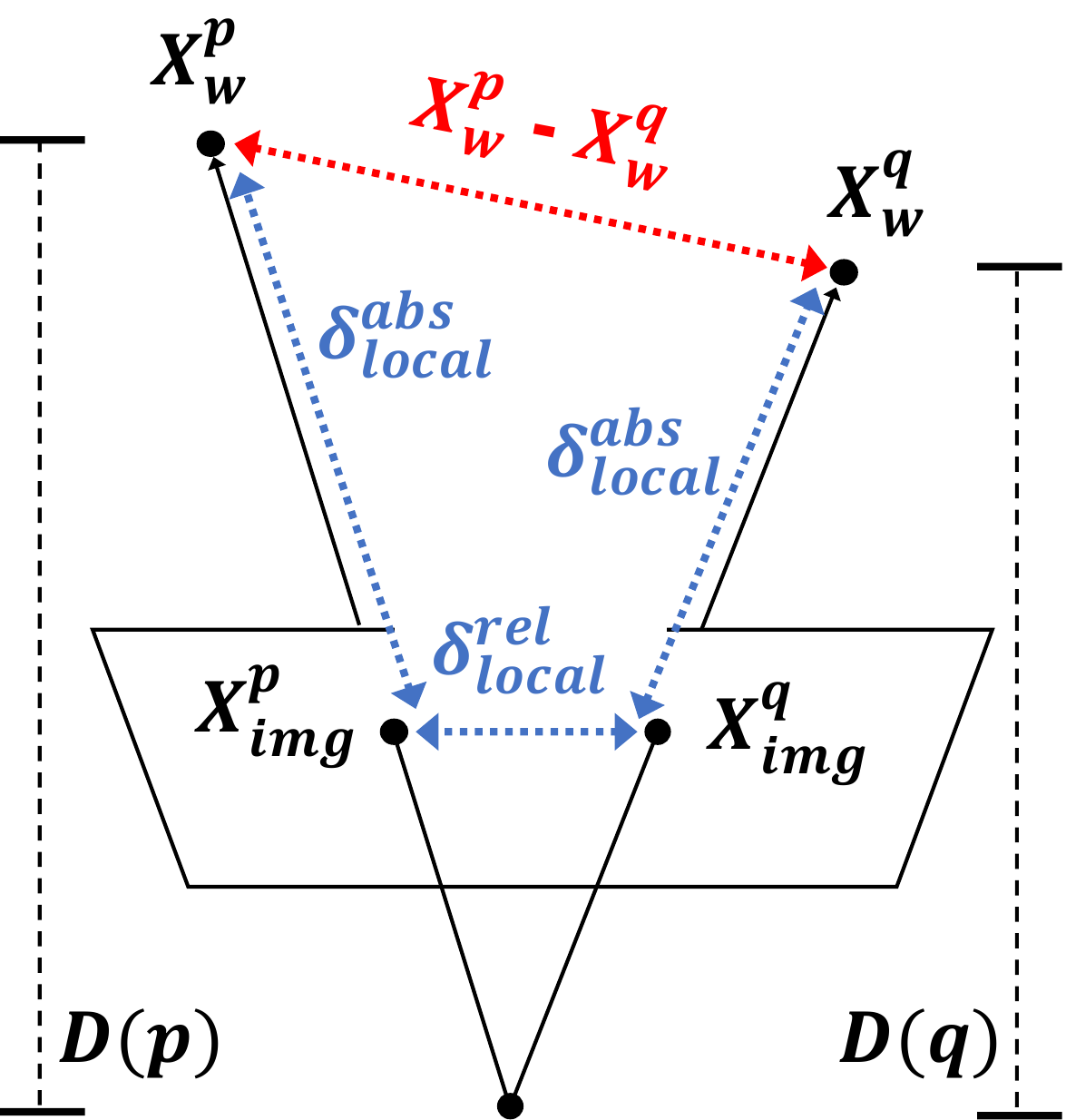}}
    \subfigure[Diagram of Local Set Attention]{\includegraphics[width=0.265\textwidth]{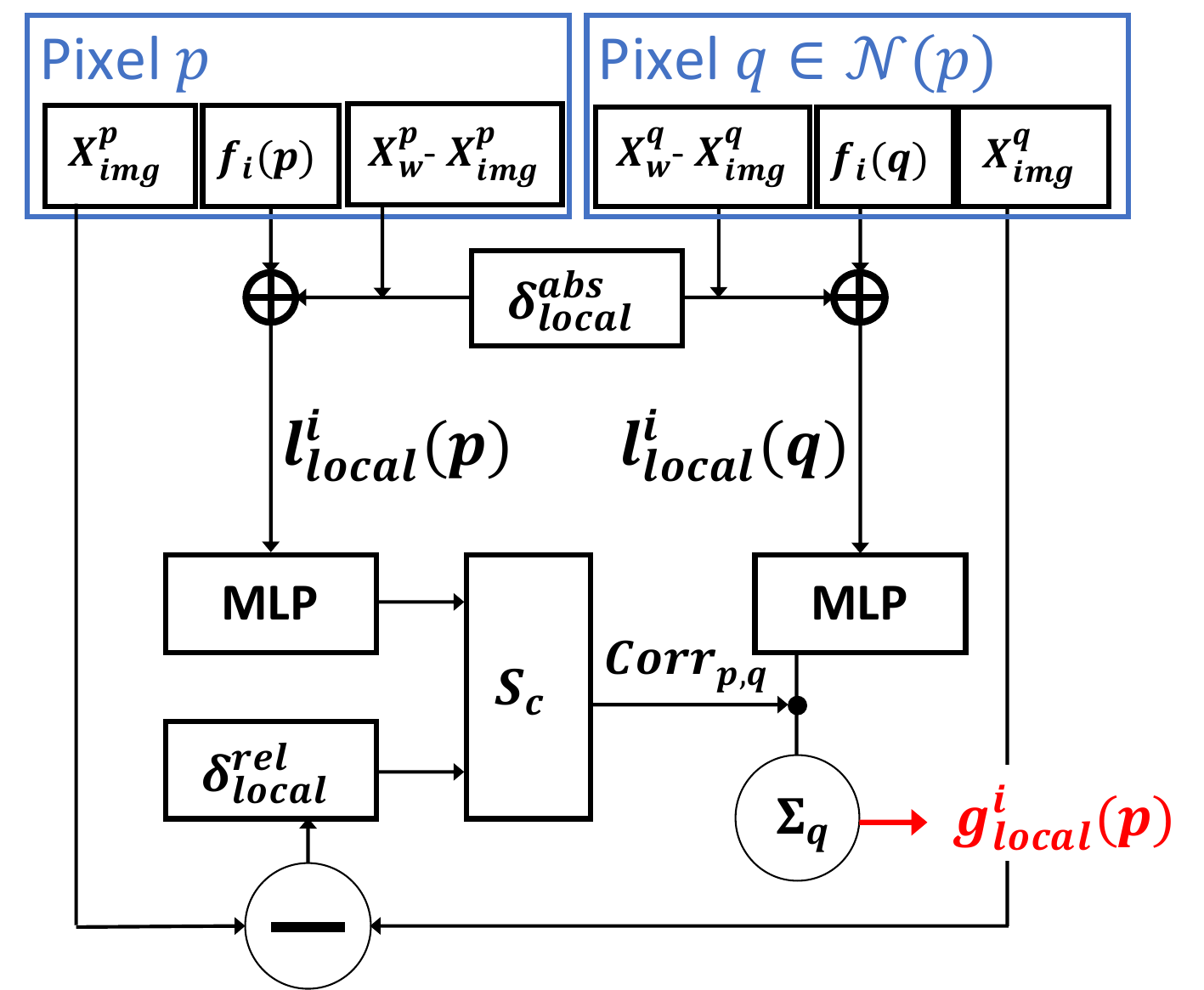}}

    \vspace{-0.4cm}
    
    \caption{\textbf{Illustration of Local Set Attention Block.} (a) A relative position in 3D world coordinates (a red dotted line) is decomposed into three relative positions (blue dotted lines). (b) Decomposed relative positions are applied for the corresponding positional encoding layer to output local set attention $g_{local}^i(p)$.}
    \vspace{-0.4cm}
    \label{fig:local_attn}
\end{figure}

\subsubsection{Local Set Attention}
Figure~\ref{fig:local_attn} shows our local set attention block. 
We use a modified Lightweight Self-Attention (LSA) layer~\cite{park2021fast} for the set attention in $r \times r$ local window of each pixel point. Unlike the decomposing relative position of voxels in~\cite{park2021fast}, we decompose the relative position of 3D world coordinates between neighbor pixels $X_{w}^p - X_{w}^q$ using normalized image coordinates as:
\begin{equation}
    \label{eq:decompose}
    (X_{w}^p - X_{img}^p) - (X_{w}^q - X_{img}^q) + (X_{img}^p - X_{img}^q),
\end{equation}
where $q \in \mathcal{N}(p)$ is a neighbor set of homogeneous coordinates in a $r \times r$ window centered at $p$.
With the decomposition in Eq.~\ref{eq:decompose}, we can divide the relative positional encoding into a continuous positional encoding $\delta_{local}^{abs}$ and a discretized positional encoding $\delta_{local}^{rel}$.
Then, the computation procedures for local set attention $g_{local}^i$ of $i$-th GLSA block is similar to original LSA layer as:
\begin{equation}
    l_{local}^i(p) = f_{i}(p) \oplus \delta_{local}^{abs}(X_{w}^p - X_{img}^p),
\end{equation}
\begin{equation}
    Corr_{p, q} = S_{C}(\psi(l_{local}^i(p)), \delta_{local}^{rel}({X_{img}^p - X_{img}^q)}),
\end{equation}
\begin{equation}
    g_{local}^i(p) = \Sigma_{q \in \mathcal{N}(p)} ({Corr_{p, q}} \phi(l_{local}^i(q))),
\end{equation}
where $\psi$ and $\phi$ are MLP-layers, and $S_{c}(a, b) = \frac{a \cdot b}{\norm{a}\norm{b}}$ computes the cosine similarity between $a$ and $b$.
As pixel coordinates of $p$ and $q$ are all integer, the encoding of $X_{img}^p - X_{img}^q$ is hashed over $r^{2}-1$ values, resulting in a space complexity reduction from $\mathcal{O}(HW \cdot r^{2} \cdot C)$ to $\mathcal{O}(HW \cdot C) + \mathcal{O}(r^{2} \cdot C)$.
We fix local window size $r = 5$ considering the previous point transformer networks where Point Transformer~\cite{zhao2021point} uses 32 neighbors, and Fast Point Transformer~\cite{park2021fast} set local window size as 3 or 5.

\subsection{Rendering Module} 
Given the scene representations of the reference image, the rendering module learns 3D transformation from the reference viewpoint to the target viewpoint.
Based on our observations of implicit and explicit methods, we design the Explicit Renderer(ER) and the Implicit Renderer(IR) connected in parallel to bypass the seesaw problem.
The structure of the two renderers is similar; they consist of an overlapping patch embedding, transformer module~\cite{radford2019language}, and ResNet blocks~\cite{he2016deep} with upsampling layers.
Note that the overlapping patch embedding and upsampling layers are designed for downsampling and upsampling the input feature with a factor of 4, respectively.
Also, for both renderers, we use the MAB($Z, Z$) described in Eq.~\ref{eq:mab} as transformer blocks for input feature $Z$, with MiX-FFN~\cite{xie2021segformer} as the feed-forward layer.
The major difference between the two renderers is how the relative camera pose $T$ is used for the geometric transformation.

\subsubsection{Explicit Renderer (ER)}
Given the rotation matrix $R$ and translation vector $t$ of relative camera pose $T$, $p$ can be reprojected to the homogeneous coordinates of target viewpoint $p^\prime$ as $p^\prime = K_{\downarrow}RX_{w}^p + t$.
The output of encoder $f_N$ is warped by splatting operation~\cite{niklaus2020softmax} with optical flow from $p$ to $p^\prime$.
Then, warped $f_N$ goes through the explicit renderer to produce the explicit feature map $h_e$.

\subsubsection{Implicit Renderer (IR)}
Unlike the explicit renderer, the implicit renderer uses the camera parameter itself. 
Instead of embedding 3x4 camera extrinsic matrix, we use seven key parameters to embed pose information; Translation vector $t$ and axis-angle notation $(\frac{\textbf{u}}{\norm{\textbf{u}}}, \theta)$ to parameterize rotation matrix $R$.
Axis-angle notation consists of \textit{ normalized axis}, i.e., a normalized vector along the axis is not changed by the rotation, and \textit{angle}, i.e., the amount of rotation about that axis.
We use a standard method that defines the eigenvector \textbf{u} of the rotation matrix by using the property that $R-R^{T}$ is a skew-symmetric matrix as:
\vspace{-0.2cm}
\begin{equation}
\vspace{-0.2cm}
\begin{gathered}
    {[\textbf{u}]}_{X} \equiv (R-R^{T}), \\
    i.e., \ \textbf{u} = [r_{32} - r_{23}, \ r_{13} - r_{31}, \ r_{21} - r_{12}]^{T},
\end{gathered}
\end{equation}
where $r_{ij}$ is the element of $R$ located at the $i$-th row and the $j$-th column. We can also calculate the rotation angle $\theta$ from the relationship between the norm of eigenvector $\norm{\textbf{u}}$ and the trace of the rotation matrix $tr(R)$. Following the existing theorem~\cite{shepperd1978quaternion, diebel2006representing}, the rotation angle $\theta$ is derived as:
\begin{equation}
    \theta = \arctan (\frac{\norm{\textbf{u}}}{tr(R)-1}).
\end{equation}

With a translation vector $t$, seven pose parameters (i.e., $(\frac{\textbf{u}}{\norm{\textbf{u}}}, \theta, t)$) are processed into $\delta_{pos}$, and then added to all output tokens of the overlapping patch embedding layer.
Taken together, the output feature of our encoder $f_N$ passes through the implicit renderer and outputs the implicit feature map $h_i$.

\subsection{Decoder}
Two feature maps from \textit{ER} and \textit{IR}, which are denoted as $h_{e}$ and $h_{i}$, are then concatenated before the decoder. 
We use a simple CNN-based decoder by gradually upsampling the concatenated feature map with four ResNet blocks. 
Instead of generating pixels in an auto-regressive manner, we directly predict all pixels in the one-path, resulting in more than 110 times faster than the state-of-the-art autoregressive methods~\cite{rockwell2021pixelsynth, rombach2021geometry, ren2022look} in generating images.

\subsection{Loss Design for ViewNet}

Following the previous single-image view synthesis methods~\cite{wiles2020synsin, rockwell2021pixelsynth}, we also use the $\ell_1$-loss, perceptual loss~\cite{park2019semantic} and adversarial loss to learn the network.
Specifically, we compute $\ell_1$-loss and perceptual loss between $I_{tgt}$ and the ground-truth image $I_{gt}$ at the target viewpoint.
Also, we use the global and local discriminators~\cite{iizuka2017globally} with the Projected GAN~\cite{sauer2021projected} structure and the hinge loss~\cite{lim2017geometric}.
We observe that our methods improve the generation performance even through these simple network structural innovations.
Furthermore, we introduce a transformation similarity loss $L_{ts}$ to complement two output feature maps $h_{e}$ and $h_{i}$.

\begin{figure}[ht]
  \centering
  \includegraphics[width=0.9\linewidth]{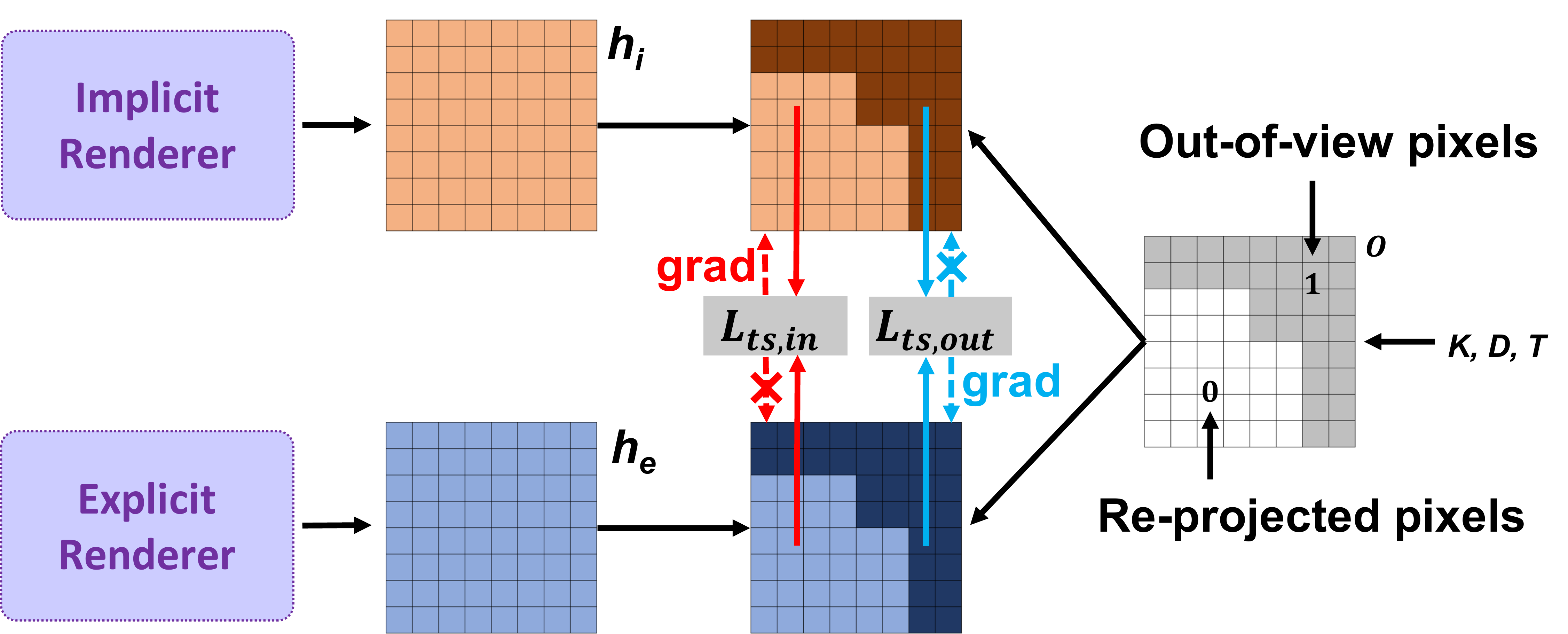}
  \vspace{-0.2cm}
  \caption{\textbf{An overview of our transformation similarity loss.} Two transformed features, $h_{i}$ and $h_{e}$, are complemented each other by the transformation similarity loss. Specifically, we first derive out-of-view mask $\textbf{O}$ from $K$, $D$ and $T$. By using \textbf{O}, two transformation similarity loss, i.e., $L_{ts, in}$ and $L_{ts, out}$, are applied to encourage the discriminability of $h_{i}$ and $h_{e}$, respectively. To guide the another renderer as intended, we allow the back-propagated gradients of $L_{ts, in}$ only to the reprojected regions of $h_{i}$, and those of $L_{ts, out}$ only to the out-of-view regions of $h_{e}$.}
  \vspace{-0.4cm}
  \label{fig_tsloss}
\end{figure}

\begin{table*}[ht]
    \caption{\textbf{Quantitative results on RealEstate10K and ACID.} Image quality is measured by PSNR-all, PSNR-vis, SSIM, LPIPS, and FID for three types of view changes, i.e., \textit{Small}, \textit{Medium} and \textit{Large}. For both datasets, best results in each metric are in \textbf{bold}, and second best are {\ul underlined}.}
   \vspace{-3mm}
    \centering
    \setlength\tabcolsep{3pt}
    \renewcommand{\arraystretch}{1.1}
    \resizebox{1.0\linewidth}{!}{
        \begin{tabular}{llcccccccccccc}
        \toprule
        \multirow{2}{*}{Dataset} & \multirow{2}{*}{Methods} &  \multicolumn{4}{c}{Small} & \multicolumn{4}{c}{Medium} & \multicolumn{4}{c}{Large} \\
        \arrayrulecolor{gray}\cmidrule(lr){3-6} \cmidrule(lr){7-10} \cmidrule(lr){11-14}
         & & PSNR-(all/vis)$\uparrow$ & SSIM$\uparrow$ & LPIPS$\downarrow$ & FID$\downarrow$   & PSNR-(all/vis)$\uparrow$  & SSIM$\uparrow$ & LPIPS$\downarrow$ & FID$\downarrow$ &PSNR-(all/vis)$\uparrow$ & SSIM$\uparrow$ & LPIPS$\downarrow$ & FID$\downarrow$ \\
         \arrayrulecolor{black}\midrule
        \multirow{8}{*}{RealEstate10K~\cite{zhou2018stereo}} & MV-3D~\cite{tatarchenko2016multi} & 11.12 / 11.16 & 0.364 & 0.597 & 258.75  & 10.90 / 10.75  & 0.361 & 0.600 & 248.55 & 10.80 / 10.70 & 0.348 & 0.614 & 249.24   \\
        & ViewApp~\cite{zhou2016view} & 12.51 / 12.39 & 0.429 & 0.555 & 142.93 & 12.79 / 12.89 & 0.431 & 0.549 & 110.84 & 12.44 / 12.50  & 0.431 & 0.559 & 147.27 \\
        & SynSin~\cite{wiles2020synsin}  &  {\ul 15.38} / {\ul 15.67} & {\ul 0.530} & 0.426 & 41.75 & {\ul 14.88} / 15.46 & 0.481 & 0.436 & 43.06 & {\ul 13.96} / 14.72 & 0.464 & 0.469 & 61.67 \\
        & SynSin-6x~\cite{wiles2020synsin} & 15.17 / 15.43 & 0.525 & {\ul 0.410} & 33.72 & \textbf{14.99} / 15.54 & \textbf{0.510} & 0.442  & 37.28 & \textbf{14.26} / {\ul 14.92}  & 0.475 & 0.499 & 48.29     \\
        & PixelSynth~\cite{rockwell2021pixelsynth}   &  14.46 / 15.62  & 0.521 & 0.417         & 37.23             & 13.46 / {\ul 15.60}  & 0.490 & {\ul 0.424} & 38.39    & 14.64 / 12.28  & \textbf{0.482} & {\ul 0.467}  & 45.44 \\
        & GeoFree~\cite{rombach2021geometry} & 14.16 / 14.89  & 0.466 & 0.436 & {\ul 33.48} & 13.15 / 14.37 & 0.435 & 0.458  & {\ul 34.21}  & 12.57 / 13.60  & 0.421 & 0.491 & {\ul 35.28}      \\
        & LookOut~\cite{ren2022look}   &  12.58 / 12.78 & 0.405 & 0.506         & 44.87       & 12.72 / 13.13     & 0.411 & 0.491      & 43.17        & 12.11 / 12.54  & 0.416 & 0.508   & 43.22    \\
        &   \cellcolor{gray!25} \textbf{Ours}  &  \cellcolor{gray!25} \textbf{15.87} / \textbf{16.94} &  \cellcolor{gray!25} \textbf{0.533} &  \cellcolor{gray!25} \textbf{0.396} &  \cellcolor{gray!25} \textbf{32.42}    &  \cellcolor{gray!25} 14.65 / \textbf{15.97}   &  \cellcolor{gray!25} {\ul 0.496} &    \cellcolor{gray!25} \textbf{0.417}     &  \cellcolor{gray!25} \textbf{33.04}     &  \cellcolor{gray!25} 13.83 / \textbf{15.36}    &  \cellcolor{gray!25} {\ul 0.481} &    \cellcolor{gray!25} \textbf{0.445}   &  \cellcolor{gray!25} \textbf{35.26}    \\
        \midrule
        \multirow{9}{*}{ACID~\cite{liu2021infinite}} & MV-3D~\cite{tatarchenko2016multi}  &   14.43 / 14.53  & 0.444 & 0.512 & 148.19 & 14.20 / 14.34  & 0.426 & 0.573  & 151.24 & 14.34 / 14.62 & 0.380 & 0.579 & 150.47   \\
        & ViewApp~\cite{zhou2016view}   &    14.46 / 14.66  & 0.432 & 0.522      & 161.91 & 13.58 / 13.76  & 0.363 & 0.580 & 203.19 & 13.21 / 13.22  & 0.325 & 0.611  & 218.37    \\
        & SynSin~\cite{wiles2020synsin}  &  {\ul 17.48} / {\ul 18.05} & {\ul 0.497} & 0.463 & 55.64 & {\ul 16.49} / {\ul 17.16}  & {\ul 0.447} & 0.508 & 75.88  & \textbf{16.87} / {\ul 17.32} & {\ul 0.466} & 0.503 & 79.04            \\
        & InfNat~\cite{liu2021infinite} (1-step)   &  15.94 / 16.97  &  0.453 & 0.470 & 64.32  & 14.40 / 15.74     & 0.423 & 0.540 & 90.80 & 13.65 / 15.24  & 0.396 & 0.556 & 106.28      \\
        & InfNat~\cite{liu2021infinite} (5-step)   &  15.16 / 15.76 & 0.416 & 0.501 & 64.48 & 14.79 / 15.44 & 0.416 & 0.525 & 71.52 & 14.90 / 15.62 & 0.412 & 0.522 & 65.45        \\
        & PixelSynth~\cite{rockwell2021pixelsynth}    &  15.81 / 17.61 & 0.480 & {\ul 0.443} & 53.38   & 14.33 / 16.22    & 0.440 &  {\ul 0.489} & 63.48  & 13.53 / 15.32  & 0.404 & 0.524 & 65.60   \\
        & GeoFree~\cite{rombach2021geometry}   &  14.80 / 15.26 & 0.441 & 0.491 & {\ul 53.21} & 14.24 / 14.86 & 0.438 & 0.508  & {\ul 58.92}  & 14.22 / 14.67  & 0.436 & {\ul 0.487} & {\ul 54.78}  \\
        & InfZero~\cite{li2022infinitenature} & 12.52 / 13.51 & 0.382 & 0.558 & 106.16 & 10.53 / 11.57 & 0.336 & 0.605 & 148.63 & 10.13 / 10.94 & 0.304 & 0.622 & 180.71 \\
        &  \cellcolor{gray!25} \textbf{Ours}      &   \cellcolor{gray!25} \textbf{17.52} / \textbf{18.17}  &  \cellcolor{gray!25} \textbf{0.500} &  \cellcolor{gray!25} \textbf{0.421} &  \cellcolor{gray!25} \textbf{42.52}    &  \cellcolor{gray!25} \textbf{16.54} / \textbf{17.58}  &  \cellcolor{gray!25} \textbf{0.483} &  \cellcolor{gray!25} \textbf{0.446} &  \cellcolor{gray!25} \textbf{51.56}     &  \cellcolor{gray!25} {\ul 15.81} / \textbf{17.88}  &  \cellcolor{gray!25} \textbf{0.472} &  \cellcolor{gray!25} \textbf{0.455} &  \cellcolor{gray!25} \textbf{49.28}   \\ 
        \bottomrule
        \end{tabular}
    }
    \vspace{-0.2cm}
    \label{table:quan}
\end{table*}

\begin{figure*}[t!]
  \centering
  \includegraphics[width=0.95\linewidth]{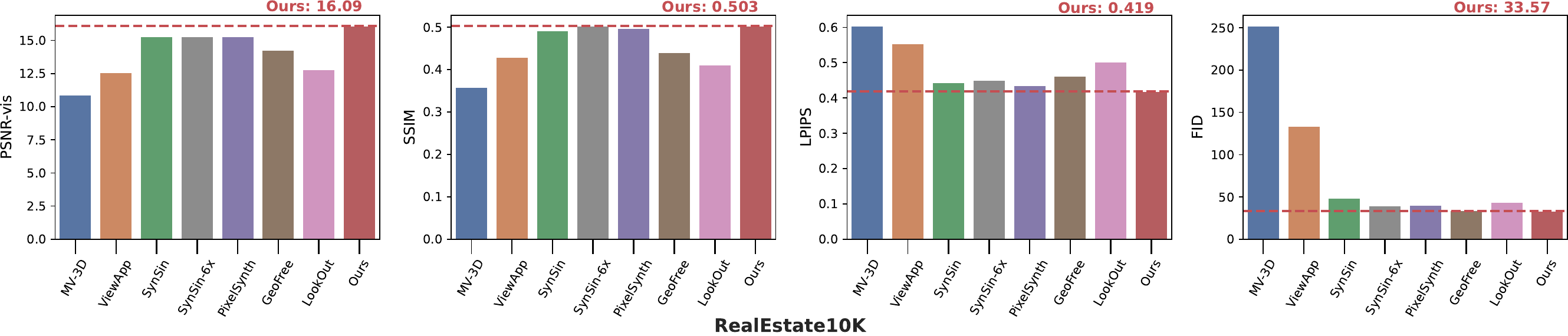}
  
  \vspace{0.2cm}
  
  \includegraphics[width=0.95\linewidth]{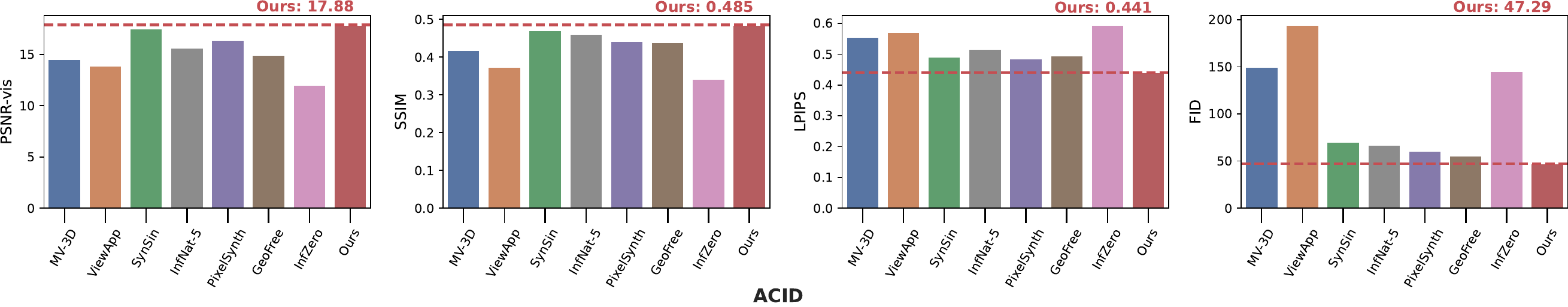}
  \vspace{-0.4cm}
  \caption{
  \textbf{Quantitative comparisons on the averaged evaluation metrics over three splits.}
  }
  \vspace{-0.4cm}
  \label{avg_quan_result}
\end{figure*}

\subsubsection{Transformation Similarity Loss}

As an extension of the existing seesaw problem, the explicit feature map $h_{e}$ may have better discriminability than the implicit feature map $h_{i}$ in reprojected regions, conversely, $h_{i}$ has better delineation of out-of-view regions than $h_{e}$.
Therefore, as shown in Fig.~\ref{fig_tsloss}, we design the transformation similarity loss between $h_{e}$ and $h_{i}$, expecting that $h_{i}$ learns to keep reprojected image contests, and $h_{e}$ also learn to generate realistic out-of-view pixels.
Specifically, we use a negative cosine similarity function $S_{c}$ for calculating the similarity between two feature maps, and the transformation similarity loss $L_{ts} = \lambda_{in} L_{ts, in} + \lambda_{out} L_{ts, out}$ is formulated as:
\begin{equation}
    L_{ts,in} = -\frac{\sum_{p}{(1-\textbf{O}(p)) \cdot S_{c}(h_{i}(p), detach(h_{e}(p)))}}{\sum_{p} (1-\textbf{O}(p))},
\end{equation}
\begin{equation}
    L_{ts,out} = -\frac{\sum_{p}{\textbf{O}(p) \cdot S_{c}(detach(h_{i}(p)), h_{e}(p))}}{\sum_{p} \textbf{O}(p)},
\end{equation}
where $\textbf{O}(p) \in \mathbb{R}^{\frac{H}{4} \times \frac{W}{4}}$ denotes an out-of-view mask derived from the depth map $D$ and the relative camera pose $T$.
Note that without detach operations, our transformation similarity loss performs the same as a simple negative cosine similarity loss between two feature maps.
Thus, we detach gradients back-propagated from $L_{ts, in}$ to $h_{e}$ and gradients from $L_{ts, out}$ to $h_{i}$, because the detach operation allows the components of $L_{ts}$ to be applied to the intended area.

\subsubsection{Final Learning Objective}
Taken together, our ViewNet is trained on the weighted sum of the $\ell_1$-loss $L_{\ell_1}$, the perceptual loss $L_{c}$, the adversarial loss $L_{adv}$ and the transformation similarity loss $L_{ts}$.
The total loss is then $L = L_{\ell_1} + \lambda_{c} L_{c} + \lambda_{adv} L_{adv} + L_{ts}$.
We fix $\lambda_{c} = 1$ and $\lambda_{adv} = 0.1$ for all experiments.

\begin{figure*}[t]
    \centering
    {\parbox{0.12\textwidth}{\centering {}}}
    {\parbox{0.12\textwidth}{\centering {\scriptsize Out-of-View(22\%)}}}
    {\parbox{0.12\textwidth}{\centering {\scriptsize PSNR-vis: 18.01}}}
    {\parbox{0.12\textwidth}{\centering {\scriptsize PSNR-vis: 18.89}}}
    {\parbox{0.12\textwidth}{\centering {\scriptsize PSNR-vis: 17.79}}}
    {\parbox{0.12\textwidth}{\centering {\scriptsize PSNR-vis: 12.67}}}
    {\parbox{0.12\textwidth}{\centering {\scriptsize PSNR-vis: 19.54}}}
    {\parbox{0.12\textwidth}{\centering {}}}

    \includegraphics[width=0.12\textwidth]{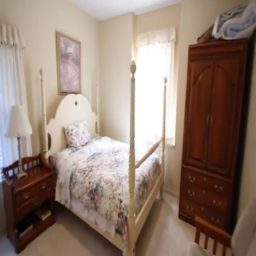}
    \includegraphics[width=0.12\textwidth]{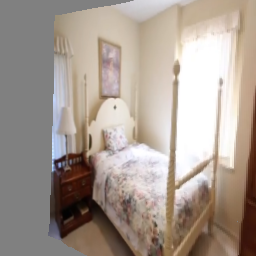}
    \includegraphics[width=0.12\textwidth]{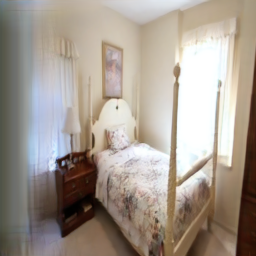}
    \includegraphics[width=0.12\textwidth]{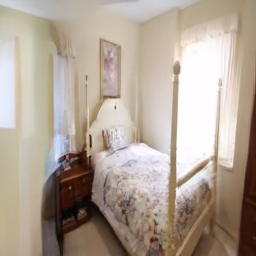}
    \includegraphics[width=0.12\textwidth]{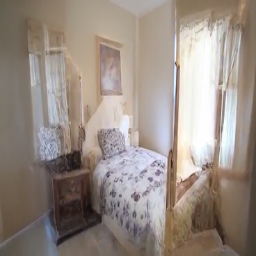}
    \includegraphics[width=0.12\textwidth]{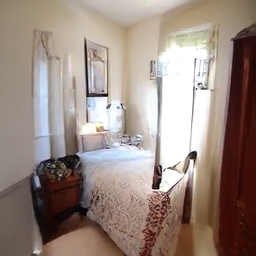}
    \includegraphics[width=0.12\textwidth]{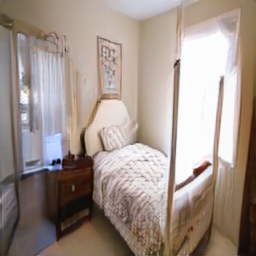}
    \includegraphics[width=0.12\textwidth]{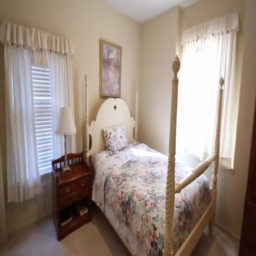}

    {\parbox{0.12\textwidth}{\centering {}}}
    {\parbox{0.12\textwidth}{\centering {\scriptsize Out-of-View(54\%)}}}
    {\parbox{0.12\textwidth}{\centering {\scriptsize PSNR-vis: 13.45}}}
    {\parbox{0.12\textwidth}{\centering {\scriptsize PSNR-vis: 13.24}}}
    {\parbox{0.12\textwidth}{\centering {\scriptsize PSNR-vis: 12.41}}}
    {\parbox{0.12\textwidth}{\centering {\scriptsize PSNR-vis: 11.22}}}
    {\parbox{0.12\textwidth}{\centering {\scriptsize PSNR-vis: 14.27}}}
    {\parbox{0.12\textwidth}{\centering {}}}

    \includegraphics[width=0.12\textwidth]{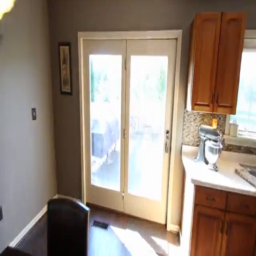}
    \includegraphics[width=0.12\textwidth]{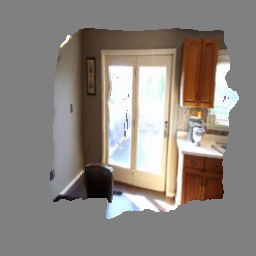}
    \includegraphics[width=0.12\textwidth]{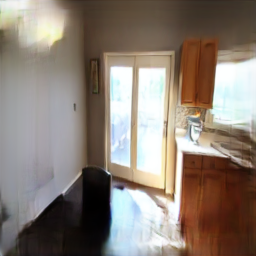}
    \includegraphics[width=0.12\textwidth]{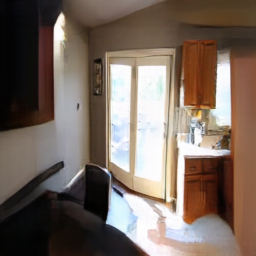}
    \includegraphics[width=0.12\textwidth]{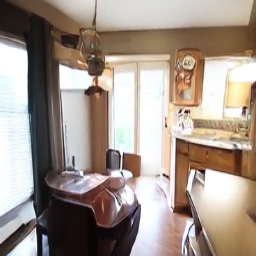}
    \includegraphics[width=0.12\textwidth]{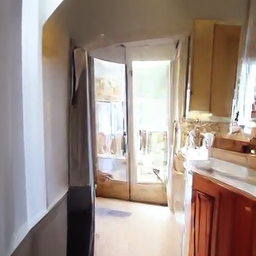}
    \includegraphics[width=0.12\textwidth]{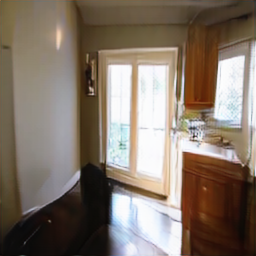}
    \includegraphics[width=0.12\textwidth]{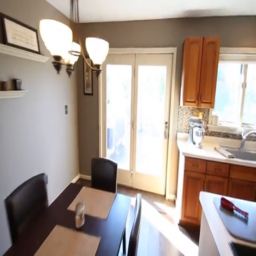}
    
    {\parbox{0.12\textwidth}{\centering {}}}
    {\parbox{0.12\textwidth}{\centering {\scriptsize Out-of-View(73\%)}}}
    {\parbox{0.12\textwidth}{\centering {\scriptsize PSNR-vis: 14.63}}}
    {\parbox{0.12\textwidth}{\centering {\scriptsize PSNR-vis: 14.64}}}
    {\parbox{0.12\textwidth}{\centering {\scriptsize PSNR-vis: 12.02}}}
    {\parbox{0.12\textwidth}{\centering {\scriptsize PSNR-vis: 9.08}}}
    {\parbox{0.12\textwidth}{\centering {\scriptsize PSNR-vis: 15.66}}}
    {\parbox{0.12\textwidth}{\centering {}}}

    \vspace{-.15cm}
    
    \subfigure[{\scriptsize Input Image}]{\includegraphics[width=0.12\textwidth]{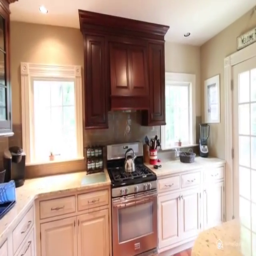}}
    \subfigure[{\scriptsize Warped Image}]{\includegraphics[width=0.12\textwidth]{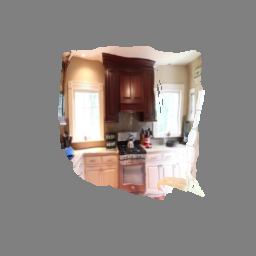}}
    \subfigure[{\scriptsize SynSin~\cite{wiles2020synsin}}]{\includegraphics[width=0.12\textwidth]{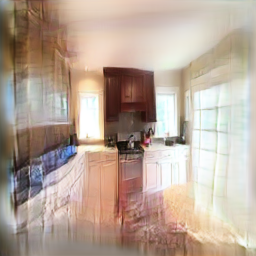}}
    \subfigure[{\scriptsize PixelSynth~\cite{rockwell2021pixelsynth}}]{\includegraphics[width=0.12\textwidth]{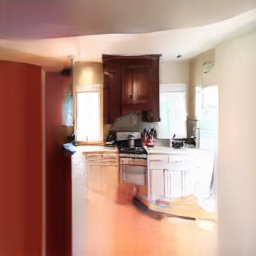}}
    \subfigure[{\scriptsize GeoFree~\cite{rombach2021geometry}}]{\includegraphics[width=0.12\textwidth]{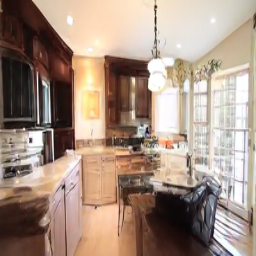}}
    \subfigure[{\scriptsize LookOut~\cite{ren2022look}}]{\includegraphics[width=0.12\textwidth]{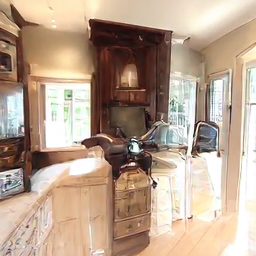}}
    \subfigure[{\scriptsize Ours}]{\includegraphics[width=0.12\textwidth]{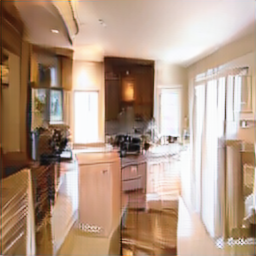}}
    \subfigure[{\scriptsize Ground Truth}]{\includegraphics[width=0.12\textwidth]{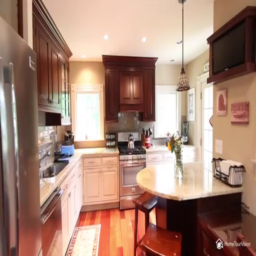}}
    
    \vspace{-0.2cm}
    
    \caption{\textbf{Comparison to baselines on RealEstate10K~\cite{zhou2018stereo}.}}

    \vspace{-0.25cm}
    
    \label{fig:comp_real}
\end{figure*}

\begin{figure*}[ht]
    \centering
    {\parbox{0.105\textwidth}{\centering {}}}
    {\parbox{0.105\textwidth}{\centering {\notsotiny Out-of-View(34\%)}}}
    {\parbox{0.105\textwidth}{\centering {\notsotiny PSNR-vis: 21.21}}}
    {\parbox{0.105\textwidth}{\centering {\notsotiny PSNR-vis: 20.30}}}
    {\parbox{0.105\textwidth}{\centering {\notsotiny PSNR-vis: 18.25}}}
    {\parbox{0.105\textwidth}{\centering {\notsotiny PSNR-vis: 19.93}}}
    {\parbox{0.105\textwidth}{\centering {\notsotiny PSNR-vis: 16.20}}}
    {\parbox{0.105\textwidth}{\centering {\notsotiny PSNR-vis: 23.69}}}
    {\parbox{0.105\textwidth}{\centering {}}}
    
    \includegraphics[width=0.105\textwidth]{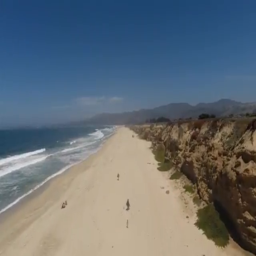}
    \includegraphics[width=0.105\textwidth]{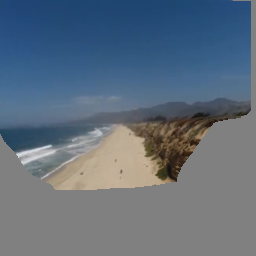}
    \includegraphics[width=0.105\textwidth]{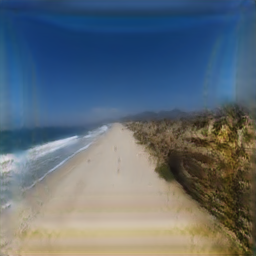}
    \includegraphics[width=0.105\textwidth]{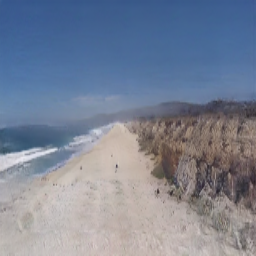}
    \includegraphics[width=0.105\textwidth]{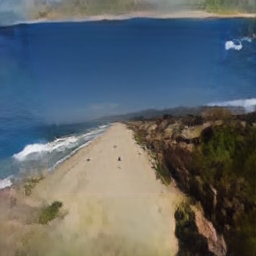}
    \includegraphics[width=0.105\textwidth]{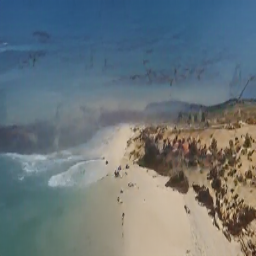}
    \includegraphics[width=0.105\textwidth]{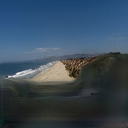}
    \includegraphics[width=0.105\textwidth]{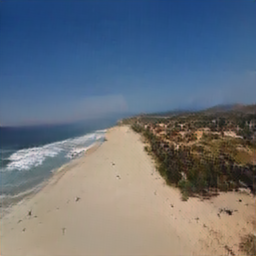}
    \includegraphics[width=0.105\textwidth]{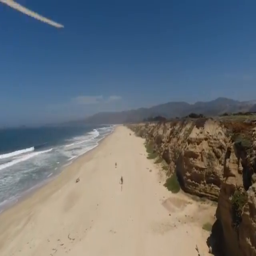}
    
    {\parbox{0.105\textwidth}{\centering {}}}
    {\parbox{0.105\textwidth}{\centering {\notsotiny Out-of-View(49\%)}}}
    {\parbox{0.105\textwidth}{\centering {\notsotiny PSNR-vis: 16.98}}}
    {\parbox{0.105\textwidth}{\centering {\notsotiny PSNR-vis: 11.83}}}
    {\parbox{0.105\textwidth}{\centering {\notsotiny PSNR-vis: 15.71}}}
    {\parbox{0.105\textwidth}{\centering {\notsotiny PSNR-vis: 11.73}}}
    {\parbox{0.105\textwidth}{\centering {\notsotiny PSNR-vis: 11.79}}}
    {\parbox{0.105\textwidth}{\centering {\notsotiny PSNR-vis: 17.13}}}
    {\parbox{0.105\textwidth}{\centering {}}}
    
    \includegraphics[width=0.105\textwidth]{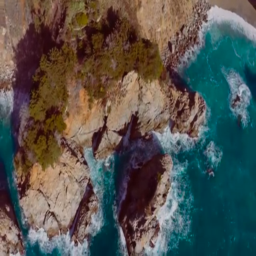}
    \includegraphics[width=0.105\textwidth]{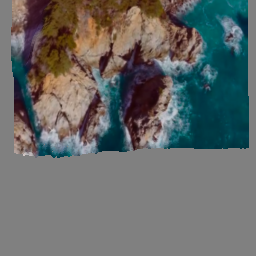}
    \includegraphics[width=0.105\textwidth]{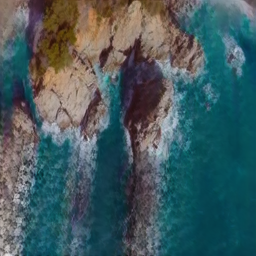}
    \includegraphics[width=0.105\textwidth]{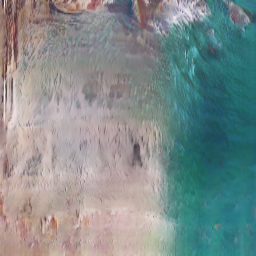}
    \includegraphics[width=0.105\textwidth]{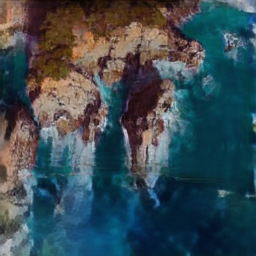}
    \includegraphics[width=0.105\textwidth]{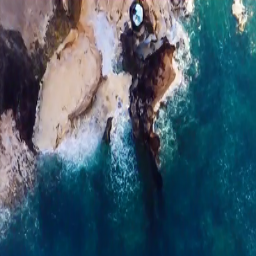}
    \includegraphics[width=0.105\textwidth]{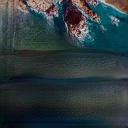}
    \includegraphics[width=0.105\textwidth]{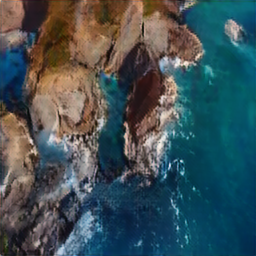}
    \includegraphics[width=0.105\textwidth]{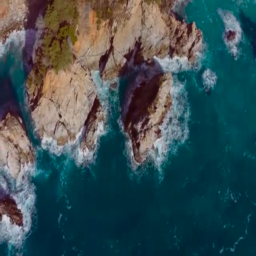}
   
    {\parbox{0.105\textwidth}{\centering {}}}
    {\parbox{0.105\textwidth}{\centering {\notsotiny Out-of-View(75\%)}}}
    {\parbox{0.105\textwidth}{\centering {\notsotiny PSNR-vis: 16.26}}}
    {\parbox{0.105\textwidth}{\centering {\notsotiny PSNR-vis: 15.67}}}
    {\parbox{0.105\textwidth}{\centering {\notsotiny PSNR-vis: 15.59}}}
    {\parbox{0.105\textwidth}{\centering {\notsotiny PSNR-vis: 15.11}}}
    {\parbox{0.105\textwidth}{\centering {\notsotiny PSNR-vis: 14.13}}}
    {\parbox{0.105\textwidth}{\centering {\notsotiny PSNR-vis: 17.38}}}
    {\parbox{0.105\textwidth}{\centering {}}}

    \vspace{-.15cm}
    
    \subfigure[{\notsotiny Input Image}]{\includegraphics[width=0.105\textwidth]{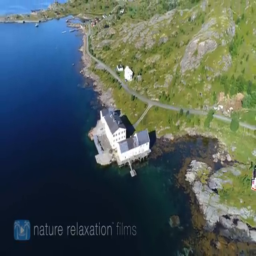}}
    \subfigure[{\notsotiny Warped Image}]{\includegraphics[width=0.105\textwidth]{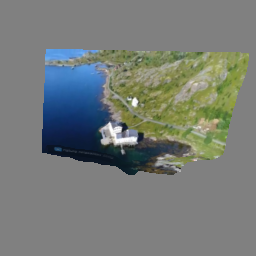}}  
    \subfigure[{\notsotiny SynSin~\cite{wiles2020synsin}}]{\includegraphics[width=0.105\textwidth]{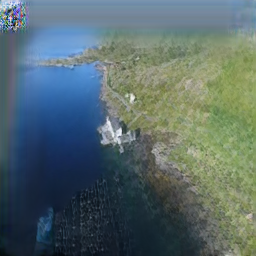}}
    \subfigure[{\notsotiny InfNat~\cite{liu2021infinite}}]{\includegraphics[width=0.105\textwidth]{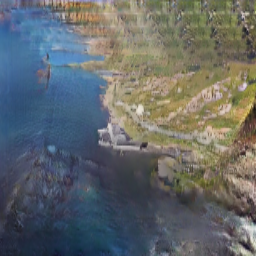}}
    \subfigure[{\notsotiny PixelSynth~\cite{rockwell2021pixelsynth}}]{\includegraphics[width=0.105\textwidth]{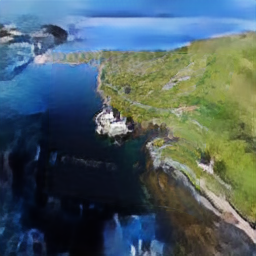}}
    \subfigure[{\notsotiny GeoFree~\cite{rombach2021geometry}}]{\includegraphics[width=0.105\textwidth]{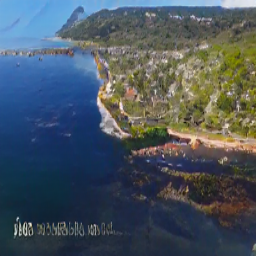}}
    \subfigure[{\notsotiny InfZero~\cite{li2022infinitenature}}]{\includegraphics[width=0.105\textwidth]{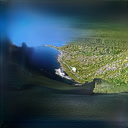}}
    \subfigure[{\notsotiny Ours}]{\includegraphics[width=0.105\textwidth]{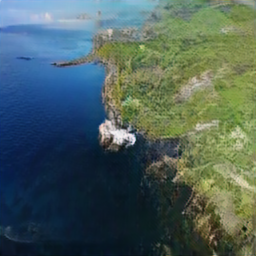}}
    \subfigure[{\notsotiny Ground Truth}]{\includegraphics[width=0.105\textwidth]{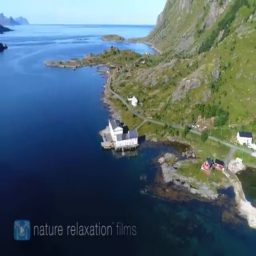}}

    \vspace{-0.2cm}
    
    \caption{\textbf{Comparison to baselines on ACID~\cite{liu2021infinite}.} For InfNat~\cite{liu2021infinite}, we report examples with higher PSNR-vis in either 1-step or 5-step.}
    \vspace{-0.5cm}
    \label{fig:comp_acid}
\end{figure*}

\section{Experimental Results}

\subsection{Experimental Settings}

\subsubsection{Dataset}
We used two standard datasets for single-image view synthesis, \textit{RealEstate10K}~\cite{zhou2018stereo} and \textit{ACID}~\cite{liu2021infinite}, which are collections of videos mostly captured in indoor and outdoor scenes, respectively.
We divided train and test sequences as in~\cite{rombach2021geometry}.
To select training image pairs from video clips in RealEstaet10K~\cite{zhou2018stereo} and ACID~\cite{liu2021infinite}, our selection protocol proceeds similarly to the previous work~\cite{wiles2020synsin}.
However, we experimentally set selection limits that allow the network to learn both small and large view changes and exclude situations of entering different rooms.
Specifically, we set the range of angle ($^{\circ}$), translation ($m$), and frame differences (frames) to $[10, 60]$, $[0, 3]$ and $[0, 100]$ for both datasets, respectively.

Moreover, we have incorporated results from the SWORD dataset, adhering to the evaluation protocol established by SIMPLI~\cite{solovev2023self}. This protocol involves training on 1800 scenes and evaluating 30 distinct scenes. The SWORD dataset is characterized by its relatively short video sequences, especially when contrasted with datasets like RealEstate10K and ACID. This feature results in a limited number of source and target image pairs, particularly when filtering out instances with minor viewpoint changes. Consequently, we opted to use the complete image sequences within each scene for training purposes.

\subsubsection{Evaluation Details}
Because explicit and implicit methods are respectively advantageous in small view change and large view change, methods should be evaluated on several sizes of viewpoint changes for a fair comparison.
Therefore,  we used a ratio of out-of-view pixels over all pixels to quantify view changes, resulting in three splits that are categorized into \textit{small} (20-40\%), \textit{medium} (40-60\%) and \textit{large} (60-80\%).
Since evaluation datasets do not have ground-truth depth maps, we used depth maps from our pre-trained DepthNet to derive the ratio of out-of-view mask pixels.
Finally, we used randomly selected 1,000 image pairs for each test split.

We use PSNR, FID~\cite{heusel2017gans}, LPIPS~\cite{zhang2018unreasonable}, and SSIM~\cite{wang2004image} as evaluation metrics.
PSNR is a traditional metric for comparing images, which is widely used to evaluate \textit{consistency}.
Nevertheless, PSNR is not suitable for verifying the image quality on large view changes~\cite{rockwell2021pixelsynth, rombach2021geometry}, so we evaluate the PSNR for entire pixels (\textit{PSNR-all}) as well as for reprojected pixels (\textit{PSNR-vis}) to clarify the performance of preserving seen contents.
For evaluating the image quality of view synthesis, FID is widely used~\cite{wiles2020synsin,rombach2021geometry,rockwell2021pixelsynth}.
Especially in the medium and large split with many out-of-view pixels, FID indicates how well the model fills out-of-view pixels and generates realistic images.
Moreover, we further evaluate the structural similarity and perceptual similarity between generated images and ground truth images with SSIM and LPIPS, respectively.
Note that these two metrics are designed to better align with human perception.

\subsubsection{Implementation Details}

We first resized all images into a resolution of $256 \times 256$, and normalized RGB value following~\cite{wiles2020synsin, rockwell2021pixelsynth}.
We trained DepthNet using the batch size of 50 for 100$k$ iterations and ViewNet using the batch size of 32 for 150$k$ iterations.
Training takes about three days on 4 NVIDIA Geforce RTX 3090 GPUs. 
We used the AdamW~\cite{loshchilov2017decoupled} optimizer (with $\beta_{1} = 0.5$ and $\beta_{2} = 0.9$) and applied weight decay of 0.01. We first linearly increased the learning rate from $10^{-6}$ to $3 \cdot 10^{-4}$ during the first 1.5$k$ steps, and then a cosine-decay learning rate schedule~\cite{loshchilov2016sgdr} was applied towards zero. 
In ViewNet, we used 8 GLSA blocks with local window size $r=5$ and set the number of transformer blocks $M=6$ in each renderer for all experiments.

\begin{figure}[t!]
    \centering
    \includegraphics[width=0.077\textwidth]{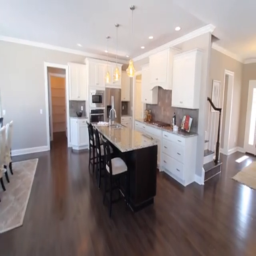}
    \includegraphics[width=0.077\textwidth]{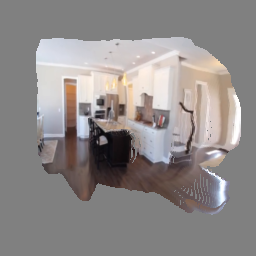}
    \includegraphics[width=0.077\textwidth]{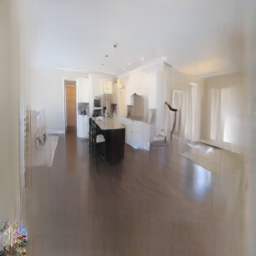}
    \includegraphics[width=0.077\textwidth]{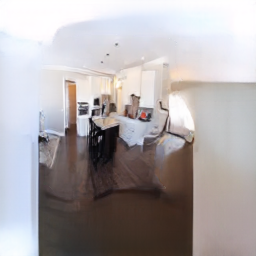}
    \includegraphics[width=0.077\textwidth]{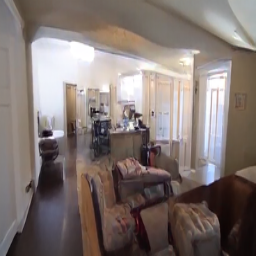}
    \includegraphics[width=0.077\textwidth]{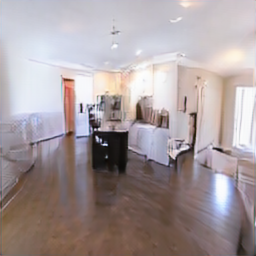}

    \vspace{.1cm}
    
    \includegraphics[width=0.077\textwidth]{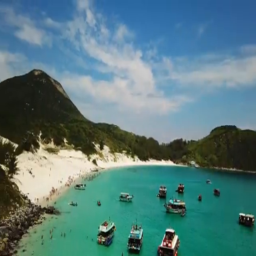}
    \includegraphics[width=0.077\textwidth]{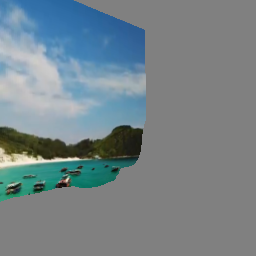}
    \includegraphics[width=0.077\textwidth]{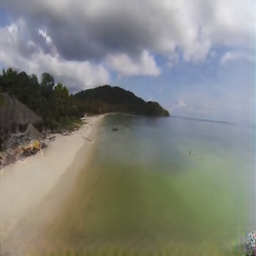}
    \includegraphics[width=0.077\textwidth]{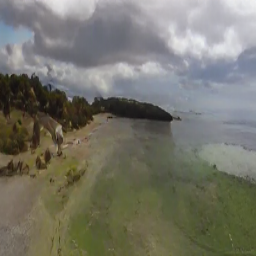}
    \includegraphics[width=0.077\textwidth]{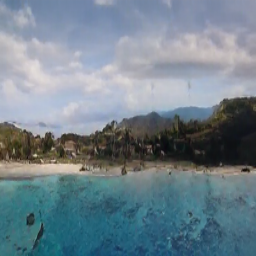}
    \includegraphics[width=0.077\textwidth]{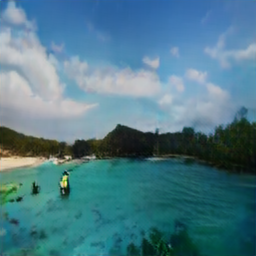}

    {\parbox{0.077\textwidth}{\centering {\scriptsize Input Img.}}}
    {\parbox{0.077\textwidth}{\centering {\scriptsize Warped Img.}}}
    {\parbox{0.077\textwidth}{\centering {\scriptsize SynSin}}}
    {\parbox{0.077\textwidth}{\centering {\scriptsize PixelSynth}}}
    {\parbox{0.077\textwidth}{\centering {\scriptsize GeoFree}}}
    {\parbox{0.077\textwidth}{\centering {\scriptsize Ours}}}
    
    \vspace{-3mm}
    
    \caption{\textbf{Qualitative Results on RealEstate10K and ACID.} We compare baselines for two datasets to our method. The first row is from RealEstate10K, and the second row is from ACID.}
    \vspace{-3mm}
    \label{fig:qual}
\end{figure}

\begin{table}[t!]
 \caption{\textbf{Average inference time.} We report averaged inference speed and how many times our method is faster than baselines as a speed-up metric.}
    \centering
    \vspace{-3mm}
    \resizebox{\linewidth}{!}{
    \begin{tabular}{lcc}
\toprule
Methods & Inference Time (sec/img) & Speed-up  \\
\midrule
SynSin~\cite{wiles2020synsin}  & 0.063 & $\times$1.23  \\ 
InfNat~\cite{liu2021infinite} (5-step)  & 1.14  & $\times$20.36 \\
PixelSynth~\cite{rockwell2021pixelsynth} & 6.22 & $\times$111.07 \\
GeoFree~\cite{rombach2021geometry}  & 9.39 & $\times$167.68 \\
LookOut~\cite{ren2022look} & 22.15 & $\times$395.54 \\
InfZero~\cite{li2022infinitenature} & 0.90 & $\times$16.07 \\
\rowcolor{gray!25} \textbf{Ours}  & \textbf{0.056} & - \\
\bottomrule
\end{tabular}
}
    \vspace{-0.2cm}
 \label{table:speed}
\end{table}

\subsection{Comparison to Baselines}

\textbf{RealEstate10K and ACID Dataset.} Table~\ref{table:quan} shows quantitative results for both datasets.
Implicit methods, LookOut~\cite{ren2022look} and InfZero~\cite{li2022infinitenature} aim at long-term novel view synthesis, so they fall short of balancing both objectives in relatively small view changes.
Also, another implicit method GeoFree~\cite{rombach2021geometry} reports a lower FID than explicit methods such as SynSin~\cite{wiles2020synsin} and PixelSynth~\cite{rockwell2021pixelsynth}, but its PSNR-vis is lower.
This shows that previous methods suffer from the seesaw problem.
However, our method consistently achieves the highest PSNR-vis in all splits on both datasets, which means our method better preserves reprojected contents than previous methods.
As observed in \cite{rockwell2021pixelsynth, ren2022look}, we note that SynSin and its variant (i.e., SynSin-6x) often produce entirely gray images in out-of-view regions, resulting they still performing competitive results in PSNR-all of the medium and large split.
This observation explains that our method performs slightly worse on PSNR-all and better on PSNR-vis for the medium and large split than SynSin~\cite{wiles2020synsin}.
Our method also achieves the lowest FID in all splits on both datasets, and this demonstrates that our method generates better quality images with filling compatible pixels regardless of view changes. 
Considering this, our method stably outperforms previous methods in all splits.
Furthermore, SSIM and LPIPS metrics validate that our method consistently generates high-quality images that align well with human perception. 
The slight performance degradation observed in comparison to explicit methods on the medium and large splits of RealEstate10K can be attributed to the fact that the SSIM tends to yield higher average values for the overall blurred images, which is the same problem in the PSNR.

We report qualitative comparisons along with PSNR-vis scores for various out-of-view ratios on RealEstate10K and ACID datasets as shown in Fig.~\ref{fig:comp_real} and \ref{fig:comp_acid}.
Each example includes a warped image, a ground-truth image, and PSNR-vis scores, making it easy to see how well the model preserves the seen contents and completes the unseen contents.
The results show that explicit methods complete blurry out-of-view regions while their PSNR-vis scores are higher than other implicit methods, and implicit methods better complete out-of-view regions.
Also, Fig.~\ref{fig:qual} illustrates that the warped regions are well-preserved and invisible parts are well-completed in our method, whereas explicit methods do not generate realistic images, and an implicit method loses the semantic information of visible contents.
Specifically, GeoFree~\cite{rombach2021geometry} does not preserve the table in the first sample and the ships floating on the sea in the second sample.
Also, explicit methods~\cite{wiles2020synsin, rockwell2021pixelsynth} either make the entire out-of-view regions in one color or produce a less realistic view than our method.
From these results, we verify that our method well preserves observed regions as well as fills unseen regions compared to baselines.

\begin{figure}[t!]
    \centering
    \includegraphics[width=0.092\textwidth]{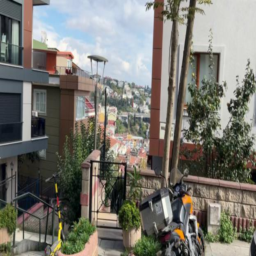}
    \includegraphics[width=0.092\textwidth]{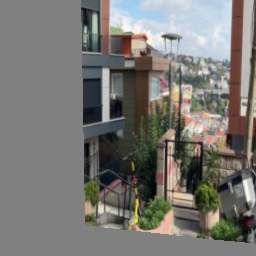}
    \includegraphics[width=0.092\textwidth]{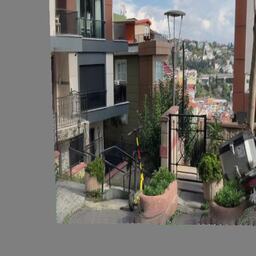}
    \includegraphics[width=0.092\textwidth]{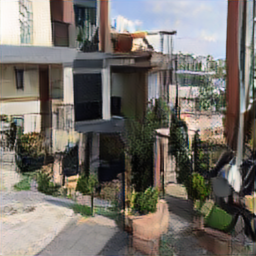}
    \includegraphics[width=0.092\textwidth]{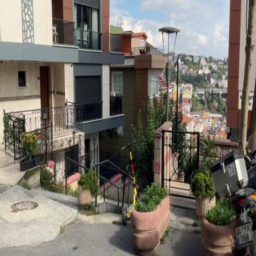}

    \vspace{.1cm}
    
    \includegraphics[width=0.092\textwidth]{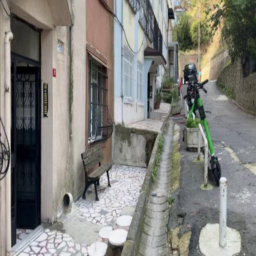}
    \includegraphics[width=0.092\textwidth]{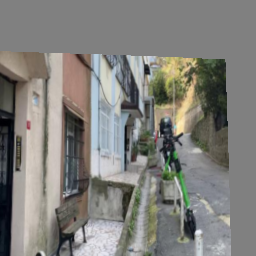}
    \includegraphics[width=0.092\textwidth]{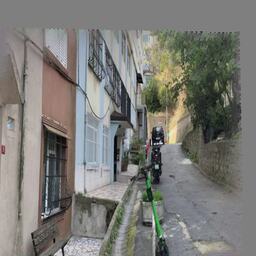}
    \includegraphics[width=0.092\textwidth]{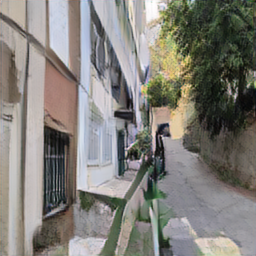}
    \includegraphics[width=0.092\textwidth]{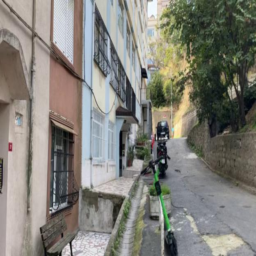}

    {\parbox{0.092\textwidth}{\centering {\scriptsize Input Image}}}
    {\parbox{0.092\textwidth}{\centering {\scriptsize Warped Image}}}
    {\parbox{0.092\textwidth}{\centering {\scriptsize SIMPLI-8L~\cite{solovev2023self}}}}
    {\parbox{0.092\textwidth}{\centering {\scriptsize Ours}}}
    {\parbox{0.092\textwidth}{\centering {\scriptsize Grount Truth}}}
    
    \vspace{-3mm}
    
    \caption{\textbf{Qualitative Results on SWORD~\cite{khakhulin2022stereo}.} We compare SIMPLI-8L~\cite{solovev2023self} to our method. Note that SIMPLI-8L utilizes two reference images, resulting in generating a novel view larger than the warped image.}
    \vspace{-3mm}
    \label{fig:qual_sword}
\end{figure}

\begin{table}[t!]
\caption{\textbf{Quantitative results on SWORD~\cite{khakhulin2022stereo}.} The results obtained with two reference views are from SIMPLI~\cite{solovev2023self}, and we report reproduced results for the single-image view synthesis performance of SIMPLI~\cite{solovev2023self}.}
    \centering
    \vspace{-3mm}
    \resizebox{\linewidth}{!}{
    \begin{tabular}{lccccc}
\toprule
Methods & $\#$ reference views & PSNR$\uparrow$ & SSIM$\uparrow$ & LPIPS$\downarrow$  \\
\midrule
IBRNet~\cite{wang2021ibrnet} & 2 & 19.02 & 0.54 & 0.35 \\ 
StereoMag~\cite{zhou2018stereo} & 2 & 18.71 & 0.53 & 0.29 \\
DeepView~\cite{flynn2019deepview} & 2 & 20.41 & 0.64 & 0.22 \\
SIMPLI-4L~\cite{solovev2023self}  & 2 & 20.78 & 0.64 & 0.23 \\
SIMPLI-8L~\cite{solovev2023self} & 2 & 20.84 & 0.64 & 0.22 \\
\arrayrulecolor{gray}\midrule
SIMPLI-8L~\cite{solovev2023self} & 1 & 15.74 & 0.35 & 0.47 \\
\rowcolor{gray!25} \textbf{Ours}  & 1 & 19.92 & 0.58 & 0.28 \\
\arrayrulecolor{black}\bottomrule
\end{tabular}
}
    \vspace{-0.4cm}
 \label{table:quan_sword}
\end{table}

\begin{figure*}[t!]
    \centering
    {\parbox{0.12\textwidth}{\centering {}}}
    {\parbox{0.12\textwidth}{\centering {\scriptsize Out-of-View(49\%)}}}
    {\parbox{0.12\textwidth}{\centering {\scriptsize PSNR-vis: 15.12} }}
    {\parbox{0.12\textwidth}{\centering {\scriptsize PSNR-vis: 11.22} }}
    {\parbox{0.12\textwidth}{\centering {\scriptsize PSNR-vis: 12.37} }}
    {\parbox{0.12\textwidth}{\centering {\scriptsize PSNR-vis: 10.21} }}
    {\parbox{0.12\textwidth}{\centering {\scriptsize PSNR-vis: 15.99} }}
    {\parbox{0.12\textwidth}{\centering {}}}
    
    \vspace{-.15cm}
    
    \subfigure[{\scriptsize Input Image}]{\includegraphics[width=0.12\textwidth]{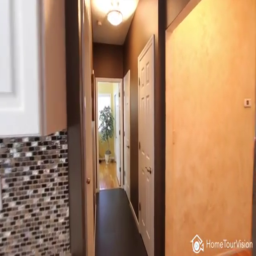}}
    \subfigure[{\scriptsize Warped Image}]{\includegraphics[width=0.12\textwidth]{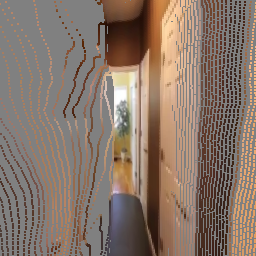}}
    \subfigure[{\scriptsize SynSin~\cite{wiles2020synsin}}]{\includegraphics[width=0.12\textwidth]{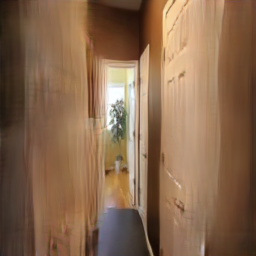}}
    \subfigure[{\scriptsize PixelSynth~\cite{rockwell2021pixelsynth}}]{\includegraphics[width=0.12\textwidth]{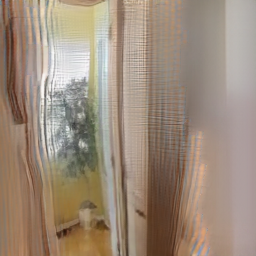}}
    \subfigure[{\scriptsize GeoFree~\cite{rombach2021geometry}}]{\includegraphics[width=0.12\textwidth]{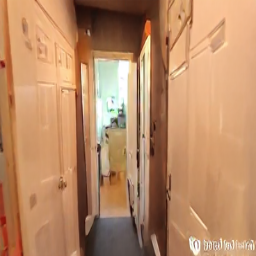}}
    \subfigure[{\scriptsize LookOut~\cite{ren2022look}}]{\includegraphics[width=0.12\textwidth]{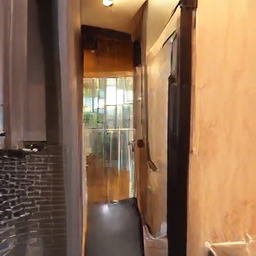}}
    \subfigure[{\scriptsize Ours}]{\includegraphics[width=0.12\textwidth]{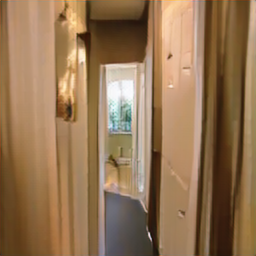}}
    \subfigure[{\scriptsize Ground Truth}]{\includegraphics[width=0.12\textwidth]{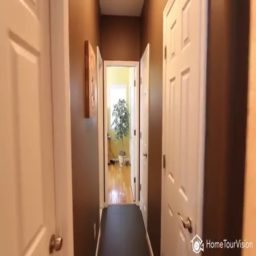}}

    \vspace{-0.35cm}
    
    \caption{\textbf{An example on the RealEstate10K~\cite{zhou2018stereo} where the camera moves forward.}}

    \vspace{-0.2cm}
    
    \label{fig:trans_hard}
\end{figure*}

\begin{figure*}[t!]
    \centering
   
    {\parbox{0.105\textwidth}{\centering {}}}
    {\parbox{0.105\textwidth}{\centering {\notsotiny Out-of-View(56\%)}}}
    {\parbox{0.105\textwidth}{\centering {\notsotiny PSNR-vis: 15.72}}}
    {\parbox{0.105\textwidth}{\centering {\notsotiny PSNR-vis: 12.18}}}
    {\parbox{0.105\textwidth}{\centering {\notsotiny PSNR-vis: 12.39}}}
    {\parbox{0.105\textwidth}{\centering {\notsotiny PSNR-vis: 12.19}}}
    {\parbox{0.105\textwidth}{\centering {\notsotiny PSNR-vis: 6.47}}}
    {\parbox{0.105\textwidth}{\centering {\notsotiny PSNR-vis: 16.37}}}
    {\parbox{0.105\textwidth}{\centering {}}}

    \vspace{-.15cm}
    
    \subfigure[{\notsotiny Input Image}]{\includegraphics[width=0.105\textwidth]{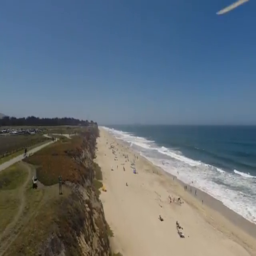}}
    \subfigure[{\notsotiny Warped Image}]{\includegraphics[width=0.105\textwidth]{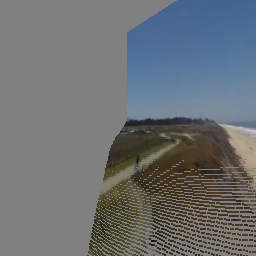}}  
    \subfigure[{\notsotiny SynSin~\cite{wiles2020synsin}}]{\includegraphics[width=0.105\textwidth]{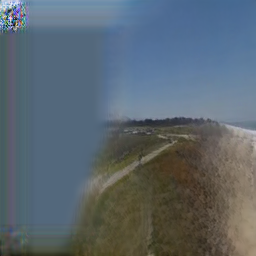}}
    \subfigure[{\notsotiny InfNat~\cite{liu2021infinite}}]{\includegraphics[width=0.105\textwidth]{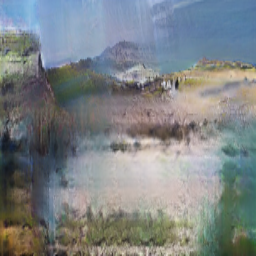}}
    \subfigure[{\notsotiny PixelSynth~\cite{rockwell2021pixelsynth}}]{\includegraphics[width=0.105\textwidth]{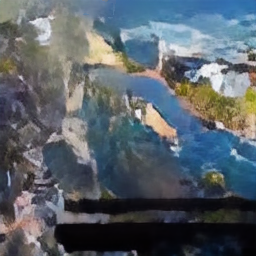}}
    \subfigure[{\notsotiny GeoFree~\cite{rombach2021geometry}}]{\includegraphics[width=0.105\textwidth]{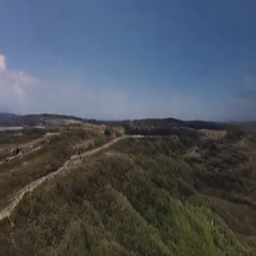}}
    \subfigure[{\notsotiny InfZero~\cite{li2022infinitenature}}]{\includegraphics[width=0.105\textwidth]{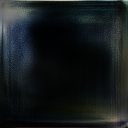}}
    \subfigure[{\notsotiny Ours}]{\includegraphics[width=0.105\textwidth]{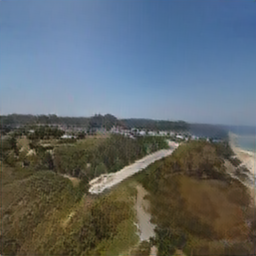}}
    \subfigure[{\notsotiny Ground Truth}]{\includegraphics[width=0.105\textwidth]{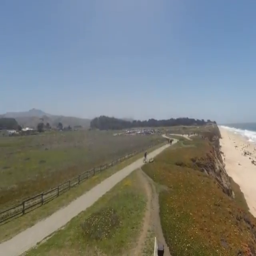}}

    \vspace{-0.35cm}
    
    \caption{\textbf{An example on ACID~\cite{liu2021infinite} where the camera moves forward.} For InfNat~\cite{liu2021infinite}, we report examples with higher PSNR-vis in either 1-step or 5-step. Autoregressive-based explicit methods fail to create realistic images.}

    \vspace{-0.4cm}
    
    \label{fig:hard_acid}
\end{figure*}

\noindent\textbf{SWORD Dataset.} Table~\ref{table:quan_sword} shows the quantitative results.
Unlike typical baselines on the SWORD dataset, which focus on synthesizing novel views of observed regions from multiple images, our approach is distinct in its use of a single-view image for synthesis.
Therefore, for a fair and relevant comparison, we configure the baseline methods to utilize two reference views and include the evaluation of single-image view synthesis performance for SIMPL~\cite{solovev2023self}.
We observe that our method achieves competitive results with previous methods that generate novel views for only visible regions through multiple images while demonstrating significantly superior performance in single-image view synthesis.
This potent capability to create out-of-view is corroborated in Fig.~\ref{fig:qual_sword}, where SIMPLI-8L~\cite{solovev2023self} utilizes two reference images to generate slightly more regions than the warped image, whereas our method produces a realistic novel view.

\noindent\textbf{Inference Speed.} We confirm that mitigating the seesaw problem by well-bridged explicit and implicit geometric transformations yields high-quality view synthesis, even acquiring a generation speed of about 110 times faster than the previous autoregressive models, as shown in Table~\ref{table:speed}.
The fast generation of novel view images allows our method to be scalable to various real-time applications.

\begin{figure}[t!]
    \centering
    \includegraphics[width=0.95\linewidth]{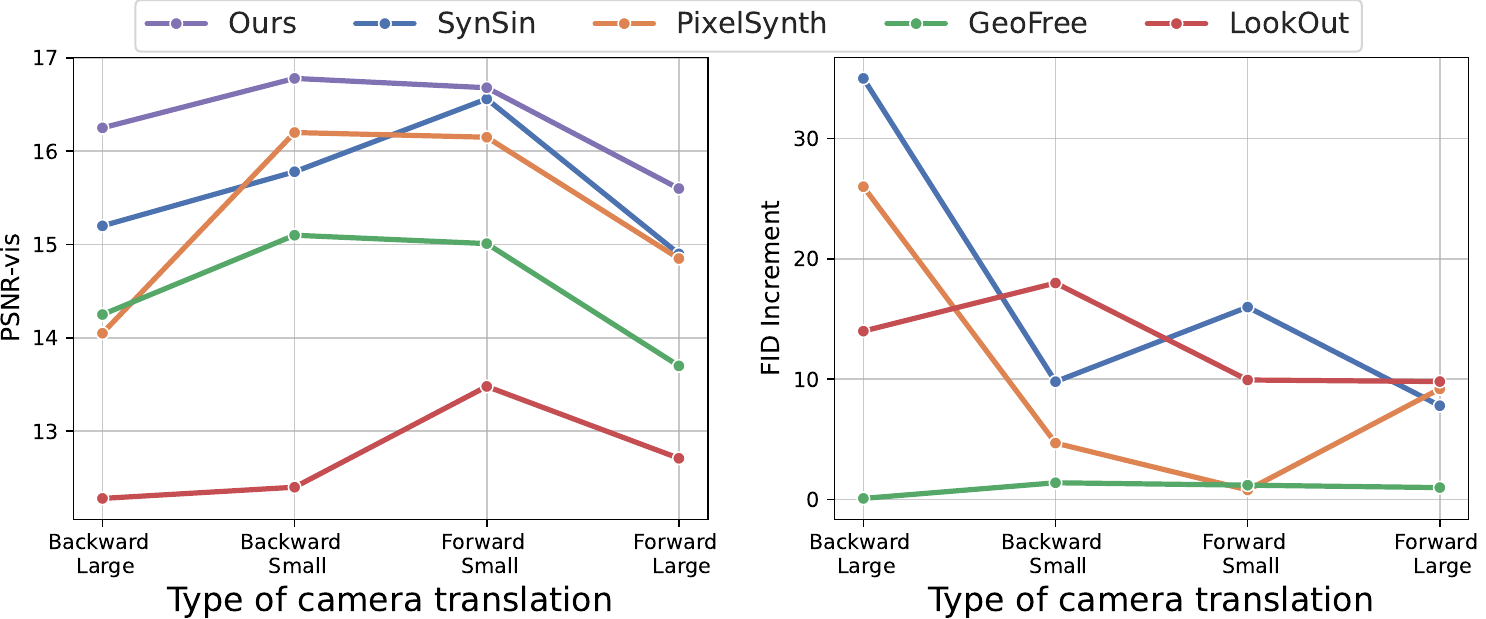}
    \vspace{-0.2cm}
    \caption{\textbf{Quantitative comparisons according to the camera translation.} We categorize the split into small and large depending on whether the camera moves more than 1$m$, and split into forward and backward depending on whether the camera moves forward and backward. We illustrate the PSNR-vis (left) and FID increment compared to our method (right) for these four splits.}
    \vspace{-0.4cm}
    \label{fig:trans_abl}
\end{figure}

\subsection{Performance for Interpolation and Extrapolation}
We analyze the generation performance for two types of out-of-view regions according to camera translations: 1) holes created by forward camera movements and 2) unseen regions created by backward camera movements.
Specifically, we categorize four splits based on the camera movements, which are defined as \textit{"Backward + Large"}, \textit{"Backward + Small"}, \textit{"Forward + Small"}, and \textit{"Forward + Large"}.
The former indicates whether a camera moves forward or backward, and the latter indicates whether the size of the camera movement is smaller or larger than $1m$.

Figure~\ref{fig:trans_abl} shows that explicit methods consistently outperform implicit methods on PSNR-vis scores, and implicit methods consistently outperform explicit methods on FID scores.
Interestingly, in terms of FID scores, explicit methods are highly degenerated for large backward motions, while they have relatively consistent performance for other splits. 
We verify that explicit methods perform better on interpolating (i.e., filling holes) than extrapolating (i.e., completing unseen regions) since localized scene representations are suitable for interpolating neighboring holes.
This result demonstrates explicit methods do not sufficiently represent the global scene representation.
Figure~\ref{fig:trans_hard} shows SynSin~\cite{wiles2020synsin} generates more blurry images for left regions with many holes to be filled, and Fig.~\ref{fig:hard_acid} also shows SynSin~\cite{wiles2020synsin} creates unseen regions as an entirely blue image while it interpolates holes well in right-down regions.
Moreover, autoregressive-based explicit methods~\cite{rockwell2021pixelsynth, liu2021infinite, li2022infinitenature} often fail to preserve visible regions in the input image.
However, our method performs well on both preserving the visible regions and creating the invisible regions regardless of the camera translation, as it sufficiently encodes the global and local scene representations.

\begin{table}[t!]
 \centering
      \caption{\textbf{Ablation Study on the Set Attention.} Checkmarks indicate whether we use local and global set attention in our encoder. S, M, and L indicate small, medium, and large split, respectively.}
       \label{table:set_ablation}
       \vspace{-3mm}
       \resizebox{\linewidth}{!}{
        \begin{tabular}{cccccccc}
        \toprule
        \multicolumn{2}{c}{Set Attention}  & \multicolumn{3}{c}{PSNR-all$\uparrow$} & \multicolumn{3}{c}{FID$\downarrow$} \\
        \arrayrulecolor{gray}\cmidrule(lr){1-2} \cmidrule(lr){3-5} \cmidrule(lr){6-8}
        $g_{local}$  &  $g_{global}$        & S & M & L                               & S & M & L              \\
        \arrayrulecolor{black}\midrule
        \checkmark     &                    & 15.69  & 14.64  & 13.78           & 34.07 & 34.81 & 37.63 \\
            &     \checkmark                & 15.74  & 14.61 & \textbf{13.88}   & 32.80  & 34.37  & 38.68   \\
        \rowcolor{gray!25} \checkmark  & \checkmark            & \textbf{15.87} & \textbf{14.65} & 13.83  & \textbf{32.42}  & \textbf{33.04} & \textbf{35.26}   \\
        \bottomrule
        \end{tabular}}
        \vspace{-0.4cm}
\end{table}

\subsection{Ablation Study}

\subsubsection{Type of Set Attention.}
We design the global and local set attention block to simultaneously extract overall contexts and detailed semantics.
Therefore, we conducted an ablation study on the RealEstate10K~\cite{zhou2018stereo} to verify each attention improves the performance of generating novel views.
Table~\ref{table:set_ablation} shows the quantitative result for the type of set attention.
Interestingly, our local set attention improves the performance relatively in large view changes, while our global set attention performs well on small view changes.
From this result, we conjecture that local and global set attention are more useful for structural reasoning of out-of-view regions and 3D scene representation of reprojected regions, respectively.
Also, significant performance improvement is achieved when both attentions are used.

\begin{table}[t!]
  \centering
        \caption{\textbf{Ablation Study on components and hyperparameters of transformation similarity loss.} * means that the detach operation is not applied to each component of the transformation similarity loss.}
        \label{table:weight_ablation}
        \vspace{-3mm}
        \resizebox{\linewidth}{!}{
        \begin{tabular}{cccccccc}
\toprule
\multicolumn{2}{c}{Loss Weight}                    & \multicolumn{3}{c}{PSNR-all$\uparrow$} & \multicolumn{3}{c}{FID$\downarrow$} \\
\arrayrulecolor{gray}\cmidrule(lr){1-2} \cmidrule(lr){3-5} \cmidrule(lr){6-8}
\multicolumn{1}{c}{$\lambda_{in}$} &$\lambda_{out}$ & S & M & L                               & S & M & L              \\
\arrayrulecolor{black}\midrule
\multicolumn{1}{c}{}    &                           & 15.41 & 14.42 & 13.57                   & 35.52 & 38.10 & 47.74  \\
\multicolumn{1}{c}{}    & $\checkmark$ (1)          & 15.72 & 14.62 & \textbf{13.85}          & 34.59 & 35.53 & 40.72  \\
\multicolumn{1}{c}{$\checkmark$ (1)}    &           & 15.63 & 14.49 & 13.73                   & 35.11 & 38.88 & 40.17  \\
\midrule
\multicolumn{1}{c}{0.1}    & 1                      & 15.78 & \textbf{14.65} &  13.81         & 33.95 & 34.10 & 37.11  \\
\multicolumn{1}{c}{10}     & 1                      & 15.48 & 14.39 & 13.56                   & 37.46 & 37.46 & 40.69  \\
\multicolumn{1}{c}{1}      & 0.1                    & 15.46 & 14.37 & 13.64                   & 34.98 & 37.51 & 39.81  \\
\multicolumn{1}{c}{1}      & 10                     & 15.70 & 14.54 & 13.77                   & 35.03 & 35.57 & 38.43  \\
\midrule
\multicolumn{1}{c}{1*} & 1*                         & 15.71 & 14.44  & 13.62                  & 35.14 & 38.26 & 47.93 \\
\rowcolor{gray!25} \multicolumn{1}{c}{1}      & 1                      & \textbf{15.87} & \textbf{14.65} & 13.83 & \textbf{32.42} & \textbf{33.04} & \textbf{35.26} \\
\bottomrule
\end{tabular}}
\vspace{-0.2cm}
\end{table}

\subsubsection{Transformation Similarity Loss.}

The transformation similarity loss $L_{ts}$ is a weighted combination of $L_{ts,in}$ and $L_{ts,out}$.
To understand the effect of each component, we conducted ablation studies of the transformation similarity loss on the RealEstate10K dataset.
Table~\ref{table:weight_ablation} reports the PSNR-all and FID of our model on three splits by changing the weights of each component for $L_{ts}$.
First, results show that combining with gradient stopping operation, $L_{ts,in}$, and $L_{ts,out}$ achieves the best results.
Either using $L_{ts,in}$ or $L_{ts,out}$ improves the performance and shows that guiding one renderer from the other renderer with the proposed loss function is effective.
Notably, our transformation similarity loss is not practical when the detach operation is not used.
We also performed an ablation study on balancing parameter $\lambda_{in}$ and $\lambda_{out}$, and results show that in the case of $\lambda_{in}=1, \lambda_{out}=1$ performs best. 
From these results, it is necessary to selectively guide unseen and seen regions by detaching the gradient and complementing each other in a balanced way.

\begin{table}[t!]
\caption{\textbf{Effects of the transformation similarity loss.} PSNR-all and FID are measured on the RealEstate10K~\cite{zhou2018stereo}. Our transformation similarity loss is more effective than just using the mask as an input of the decoder.}
\vspace{-3mm}
  \centering
    \resizebox{\linewidth}{!}{
\begin{tabular}{ccccccc}
\toprule
\multirow{2}{*}{Operation Type} & \multicolumn{3}{c}{PSNR-all$\uparrow$} & \multicolumn{3}{c}{FID$\downarrow$} \\
\arrayrulecolor{gray}\cmidrule(lr){2-4}\cmidrule{5-7}
                                & S & M & L                  & S & M & L              \\
\arrayrulecolor{black}\midrule
No $L_{ts}$                     & 15.41 & 14.42 & 13.57                   & 35.52   & 38.10  & 47.74 \\ 
$\textbf{O}(p)$ as feature      & 15.23 & 14.51 & 13.31                   & 34.74  & 36.10  & 46.43  \\ 
\rowcolor{gray!25} \textbf{Ours}  & \textbf{15.87} & \textbf{14.65} & \textbf{13.83}  & \textbf{32.42}  & \textbf{33.04}  & \textbf{35.26}   \\
\bottomrule   
\end{tabular}}
\vspace{-5mm}
    \label{table:effect_ts_loss}
\end{table}

\subsection{Effects of the Transformation Similarity Loss}

We analyze the effect of the transformation similarity loss compared to using the out-of-view mask as an additional input for the decoder.
If the out-of-view mask $\textbf{O}$ is concatenated with $h_{i}$ and $h_{e}$, the decoder can learn to fuse the rendered feature $h_{i}$ and $h_{e}$ without our transformation similarity loss.
As shown in Table~\ref{table:effect_ts_loss}, additional mask information slightly improves PSNR-vis, but the improvements in FID are negligible, considering that it takes up more parameters.
Note that two renderers without our transformation similarity loss do not sufficiently represent semantic information, although additional mask information is used.
On the other side, our method achieves significant performance improvement in both metrics while using the same number of parameters as our method trained without $L_{ts}$.

\begin{table}[t!]
\caption{\textbf{Effects of the adversarial loss.} PSNR-vis and FID are measured on RealEstate10K~\cite{zhou2018stereo}. SynSin still suffers from the seesaw problem with our adversarial loss, while our proposed method addresses the problem.}
  \centering
  \vspace{-3mm}
    \resizebox{\linewidth}{!}{
\begin{tabular}{ccccccc}
\toprule
\multirow{2}{*}{Operation Type} & \multicolumn{3}{c}{PSNR-vis$\uparrow$} & \multicolumn{3}{c}{FID$\downarrow$} \\
\arrayrulecolor{gray}\cmidrule(lr){2-4}\cmidrule{5-7}
                                & S & M & L                  & S & M & L              \\
\arrayrulecolor{black}\midrule
SynSin                  & 15.67 & 15.46 & 14.72                  & 41.75 & 43.06 & 61.67 \\ 
SynSin + our $L_{adv}$   & 15.45 & 15.31 & 14.51                   & 40.43  & 39.13  & 54.27  \\ 
\rowcolor{gray!25} \textbf{Ours} & \textbf{16.94} & \textbf{15.97} & \textbf{15.36} &  \textbf{32.42} & \textbf{33.04} & \textbf{35.26}   \\
\bottomrule         
\end{tabular}}
\vspace{-0.3cm}
    \label{table:effect_adv}
\end{table}

\begin{table}[t!]
\caption{\textbf{Effects of the DepthNet and comparison with SE3DS~\cite{koh2023simple}.} PSNR-all and FID are measured on the RealEstate10K~\cite{zhou2018stereo}.}
\vspace{-3mm}
  \centering
    \resizebox{\linewidth}{!}{
\begin{tabular}{lcccccc}
\toprule
\multirow{2}{*}{Model} & \multicolumn{3}{c}{PSNR-all$\uparrow$} & \multicolumn{3}{c}{FID$\downarrow$} \\
\arrayrulecolor{gray}\cmidrule(lr){2-4}\cmidrule{5-7}
                                & S & M & L                  & S & M & L              \\
\arrayrulecolor{black}\midrule
SE3DS~\cite{koh2023simple}                  & 15.36 & 14.21 & 12.88                   & 39.14  & 43.83  & 45.35 \\ 
Ours w/ DPT~\cite{Ranftl2021}      & 15.21 & 14.57 & 13.65                   & 37.20  & 38.15  & 42.49  \\ 
\rowcolor{gray!25} \textbf{Ours w/ DepthNet}  & \textbf{15.87} & \textbf{14.65} & \textbf{13.83}  & \textbf{32.42}  & \textbf{33.04}  & \textbf{35.26}   \\
\bottomrule
\end{tabular}}
\vspace{-3mm}
    \label{table:effect_midas}
\end{table}

\begin{figure}[t!]
  \centering
  \begin{tikzpicture}
    \draw (0, 0) node[inner sep=0] {\includegraphics[width=\linewidth]{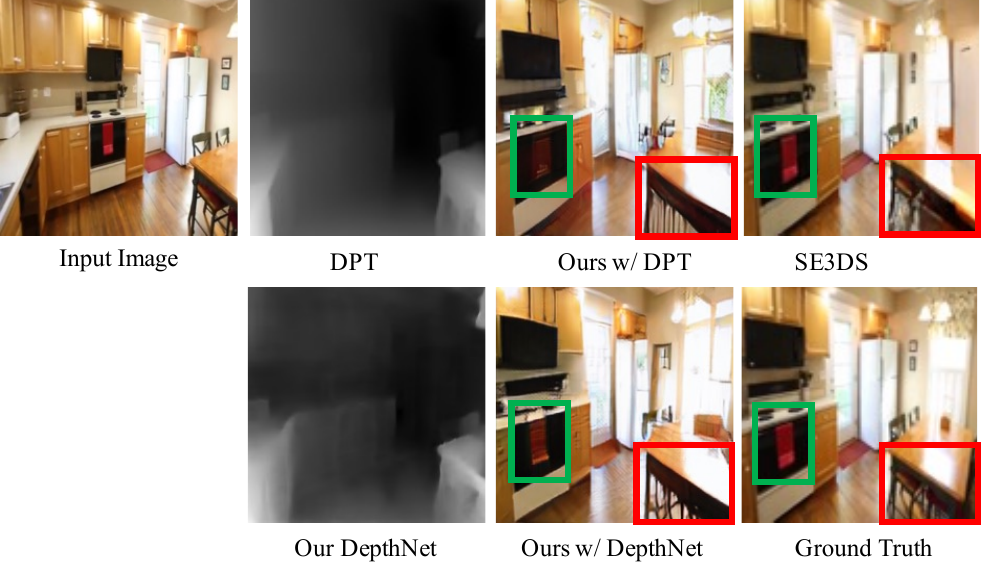}};
    \draw (-0.75, 0.22) node {\scriptsize{\cite{Ranftl2021}}};
    \draw (3.65, 0.22) node {\scriptsize{\cite{koh2023simple}}};
  \end{tikzpicture}
  
  \vspace{-0.4cm}
  
  \caption{\textbf{Qualitative comparisons based on DPT~\cite{Ranftl2021} and SE3DS~\cite{koh2023simple}.}}
  
  \vspace{-0.6cm}
  
  \label{fig_qual_midas}
\end{figure}

\subsection{Effects of the Adversarial Loss}

Since we use a different adversarial loss compared to SynSin~\cite{wiles2020synsin}, we further conducted an experiment on the effect of the adversarial loss.
Table~\ref{table:effect_adv} shows our adversarial loss improves the generation power of SynSin, but it is still a worse FID score than our method.
We confirm that our method is not just boosted with a more powerful adversarial loss.
Our architecture advances bridging explicit and implicit geometric transformations with transformation similarity loss contributes significantly to performance gain. 

Also, the new GAN loss does not solve the seesaw problem as it improves SynSin in FID by sacrificing PSNR-vis.
Explicit methods still have room for improvement in completing out-of-view regions, but more advanced generative models cannot solve the seesaw problem.
Note that our bridging scheme and the transformation similarity loss are necessary to mitigate the seesaw problem.

\begin{table*}[t!]
    \caption{\textbf{Quantitative results across different geometry usage on RealEstate10K.} Image quality is measured by PSNR, SSIM, LPIPS, and FID.}
   \vspace{-3.5mm}
    \centering
    \setlength\tabcolsep{3pt}
    \renewcommand{\arraystretch}{1.1}
    \resizebox{1.0\linewidth}{!}{
        \begin{tabular}{lcccccccccccccc}
        \toprule
        \multirow{2}{*}{Methods} & \multirow{2}{*}{\makecell{Geometry Usage\\(Explicit)}} & \multirow{2}{*}{\makecell{Inference Time \\ (sec/img)}} & \multicolumn{4}{c}{Small} & \multicolumn{4}{c}{Medium} & \multicolumn{4}{c}{Large} \\
        \arrayrulecolor{gray}\cmidrule(lr){4-7} \cmidrule(lr){8-11} \cmidrule(lr){12-15}
         & & & PSNR$\uparrow$ & SSIM$\uparrow$ & LPIPS$\downarrow$ & FID$\downarrow$  & PSNR$\uparrow$  & SSIM$\uparrow$ & LPIPS$\downarrow$ & FID$\downarrow$ & PSNR$\uparrow$ & SSIM$\uparrow$ & LPIPS$\downarrow$ & FID$\downarrow$ \\
         \arrayrulecolor{black}\midrule
        3DP~\cite{shih20203d} & Layered Depth Images & 22.1 & 11.87 & 0.423 & 0.540 & 79.06 & 12.55 & 0.436 & 0.504 & 75.42 & 12.13 & 0.450 & 0.533 & 128.79 \\
        SV-MPI~\cite{Tucker_2020_CVPR} & Multi-Plane Images  & 7.2 & 12.31  & 0.436 & 0.522 & 88.64 & 12.45 & 0.442 & 0.483 & 77.23 & 11.96  & 0.402 & 0.548 & 138.14 \\
        StableDiffusion-v1.4~\cite{rombach2022high} & Depth-Warping & 8.5 & 14.68  & 0.482 & 0.421 & 90.13 & 13.47 & 0.466 & 0.431 & 58.42 & 12.11  & 0.404 & 0.499 & 69.22 \\
        \arrayrulecolor{gray}\midrule
        \rowcolor{gray!25} \textbf{Ours} & Depth-Warping & \textbf{0.056} &  \textbf{15.87} &  \textbf{0.533} &   \textbf{0.396} &  \textbf{32.42}    & \textbf{14.65}  &   \textbf{0.496} &     \textbf{0.417}     &   \textbf{33.04}     &   \textbf{13.83}   &   \textbf{0.481} &     \textbf{0.445}   &   \textbf{35.26}    \\
        \arrayrulecolor{black}\bottomrule
        \end{tabular}
    }
    \vspace{-0.65cm}
    \label{table:quan_mpi}
\end{table*}

\subsection{Effects of the DepthNet}

We additionally compare our method with SE3DS~\cite{koh2023simple}. The results of this comparison are detailed in Table~\ref{table:effect_midas}, demonstrating that our approach significantly outperforms SE3DS across all dataset splits. Notably, our model with DepthNet achieves a higher PSNR in small (S), medium (M), and large (L) splits, and presents a more substantial improvement in FID scores, indicating a more preserving seen contents and realistic completion of the unseen contents. To be a more fair comparison, we utilized DPT~\cite{Ranftl2021} as in SE3DS and conducted a comparison in Table~\ref{table:effect_midas}. Despite the shared depth estimation technique, our method exhibits superior performance, particularly in the medium and large splits, with dramatic improvements. Even in the small split, our method maintains similar PSNR while improving upon FID, underscoring the enhanced capability of our model to synthesize high-fidelity novel views with coherent depth information. This comparison clearly evidences the strengths of our approach in handling complex scene reconstructions and view synthesis tasks.

\begin{figure}[t!]
    \centering
    \includegraphics[width=0.092\textwidth]{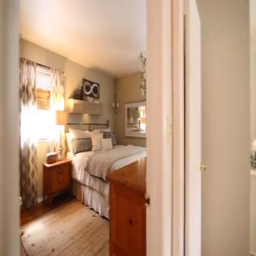}
    \includegraphics[width=0.092\textwidth]{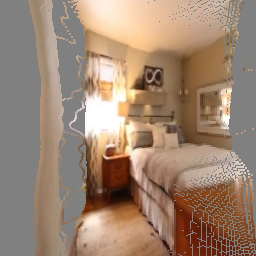}
    \includegraphics[width=0.092\textwidth]{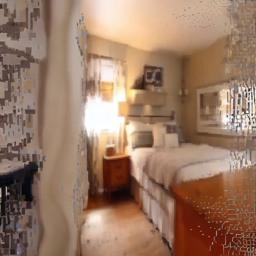}
    \includegraphics[width=0.092\textwidth]{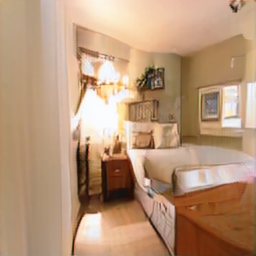}
    \includegraphics[width=0.092\textwidth]{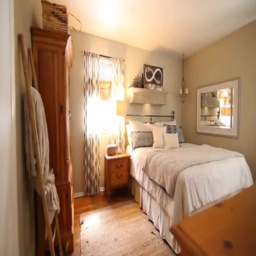}

    \vspace{.1cm}
    
    \includegraphics[width=0.092\textwidth]{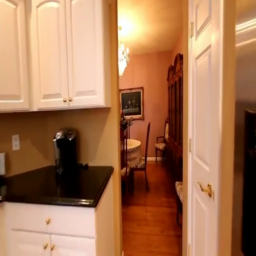}
    \includegraphics[width=0.092\textwidth]{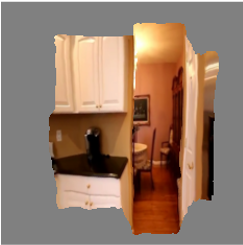}
    \includegraphics[width=0.092\textwidth]{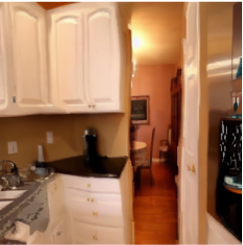}
    \includegraphics[width=0.092\textwidth]{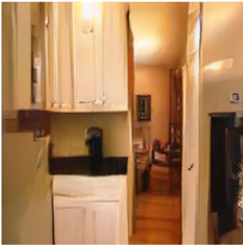}
    \includegraphics[width=0.092\textwidth]{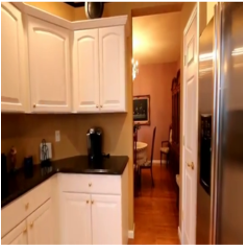}

    \vspace{.1cm}
    
    \includegraphics[width=0.092\textwidth]{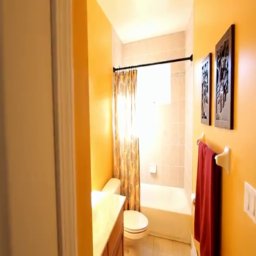}
    \includegraphics[width=0.092\textwidth]{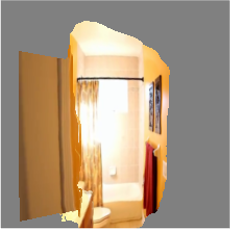}
    \includegraphics[width=0.092\textwidth]{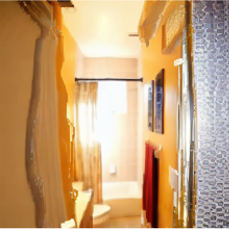}
    \includegraphics[width=0.092\textwidth]{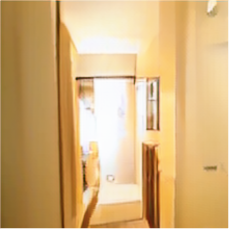}
    \includegraphics[width=0.092\textwidth]{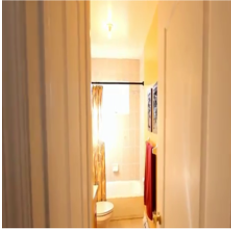}
    
    {\parbox{0.092\textwidth}{\centering {\scriptsize Input Image}}}
    {\parbox{0.092\textwidth}{\centering {\scriptsize Warped Image}}}
    {\parbox{0.092\textwidth}{\centering {\scriptsize SD-v1.4~\cite{rombach2022high}}}}
    {\parbox{0.092\textwidth}{\centering {\scriptsize Ours}}}
    {\parbox{0.092\textwidth}{\centering {\scriptsize Grount Truth}}}
    
    \vspace{-4mm}
    
  \caption{\textbf{Qualitative comparison with Diffusion-based Inpainting.}}
  \vspace{-0.6cm}
  \label{fig_qual_diffusion}
\end{figure}

To further elucidate the effectiveness of our depth estimation network, we have embraced the use of DPT~\cite{Ranftl2021, ranftl2020towards} as our depth estimation model to draw a direct comparison with SE3DS~\cite{koh2023simple}, which also utilizes DPT for single-image view synthesis. The quantitative comparisons are illustrated in Table~\ref{table:effect_midas}, where we have delineated three distinct methods: 1) our method with our DepthNet trained via a self-supervised approach, 2) our method incorporating DPT, and 3) SE3DS. The results showcase that our DepthNet is more advantageous than DPT, aiding in our method's general superiority over SE3DS, even when the same depth network is in use. Complementing these findings, Fig.~\ref{fig_qual_midas} provides a qualitative comparison that leads to two main insights: firstly, our DepthNet furnishes a more informative depth map for view synthesis compared to DPT; secondly, our method achieves more accurate view synthesis results than SE3DS, as evidenced by the precision in aligning with the ground truth.

\subsection{Comparison to different geometry usage}

For explicit methods, we further evaluate different types of representations that utilize layered depths~\cite{shih20203d, Tucker_2020_CVPR, jampani2021slide}. We report results for 3DP~\cite{shih20203d}, which uses Layered Depth Images (LDI) representation, and SV-MPI~\cite{Tucker_2020_CVPR}, which uses Multi-Plane Images (MPI) representations. These scene representations mainly focus on novel view synthesis for observed regions, addressing disocclusions and non-Lambertian effects for novel-view synthesis, and are impossible to produce realistic out-of-view regions. We also report the results of applying diffusion-based inpainting to warped images and out-of-view masks using StableDiffusion-v1.4~\cite{rombach2022high}, which has recently shown powerful generative capabilities.

Table~\ref{table:quan_mpi} shows that these methods require considerable inference time, falling short of real-time applications. Moreover, 3DP and SV-MPLI quickly degenerated on the large split where outpainting performance is critical. Also, as shown in Fig.~\ref{fig_qual_diffusion}, diffusion-based inpainting suffers from an inaccurate out-of-view mask as the camera moves forward, and does not generate pixels that are semantically compatible with the warped image.

\begin{table}[t!]
\caption{\textbf{Quantitative results across image resolution on RealEstate10K~\cite{zhou2018stereo}.}}
  \centering
  \vspace{-3mm}
    \resizebox{\linewidth}{!}{
\begin{tabular}{lccccccc}
\toprule
\multirow{2}{*}{Resolution} & \multirow{2}{*}{Patch Size} & \multicolumn{3}{c}{PSNR-vis$\uparrow$} & \multicolumn{3}{c}{LPIPS$\downarrow$} \\
\arrayrulecolor{gray}\cmidrule(lr){3-5}\cmidrule{6-8}
                           &     & S & M & L                  & S & M & L              \\
\arrayrulecolor{black}\midrule
256 $\times$ 256 & 4 $\times$ 4 & 16.94 & 15.97 & \textbf{15.36}  & 0.396  & 0.417  & \textbf{0.445} \\
512 $\times$ 512 & 8 $\times$ 8 & \textbf{17.02} & \textbf{16.05} & 15.20 & \textbf{0.389} & \textbf{0.412} & 0.466 \\ 
\bottomrule         
\end{tabular}}
\vspace{-0.4cm}
    \label{table:resolution_scalability}
\end{table}

\subsection{Analysis on Resolution Scale}
We conduct an experiment that takes doubled resolution for both the input image and patch size to maintain the overall computational cost consistent. As shown in Table~\ref{table:resolution_scalability}, quantitative results indicate a slight improvement for small and medium splits but show a minor performance degradation for large splits. These results suggest the robustness of our method over a range of image resolutions.
\section{Discussion}

\begin{figure}[t]
  \centering
  \includegraphics[width=\linewidth]{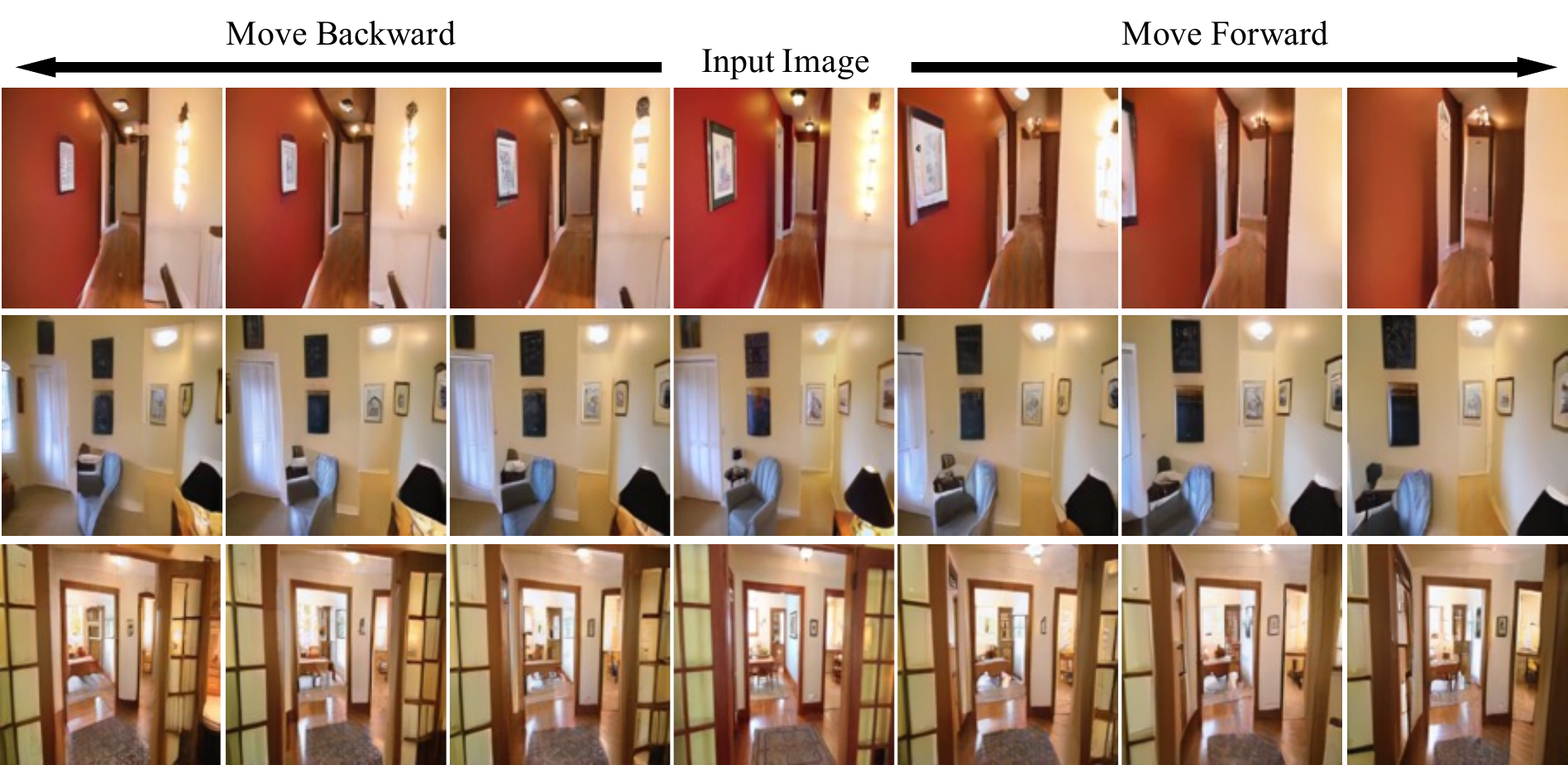}

  \vspace{-0.35cm}
  
  \caption{\textbf{Qualitative results for consistent view synthesis by sequential moving on RealEstate10K}.}
  \vspace{-0.55cm}
  \label{fig_qual_sequential}
\end{figure}

\begin{figure*}[t!]
    \centering
    \subfigure[\normalsize{Overall}]{\includegraphics[width=0.19\textwidth]{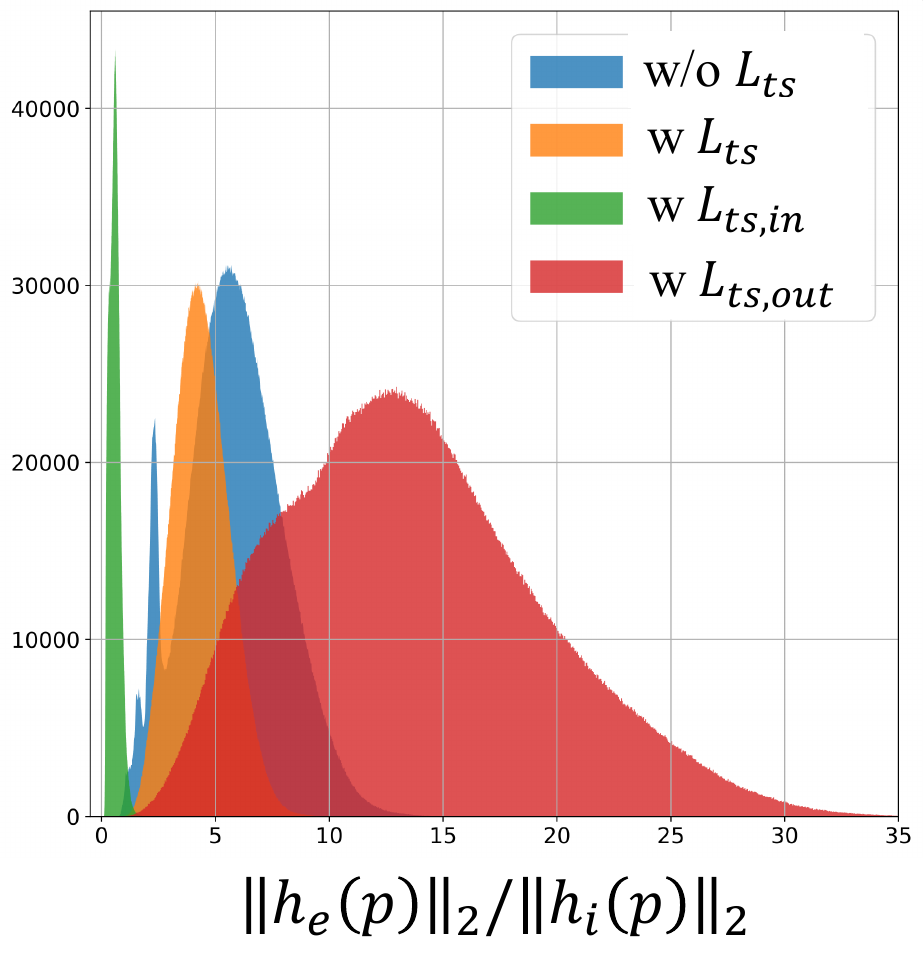} \label{fig:histogram_overall}}
    \subfigure[\normalsize{without $L_{ts}$}]{\includegraphics[width=0.19\textwidth]{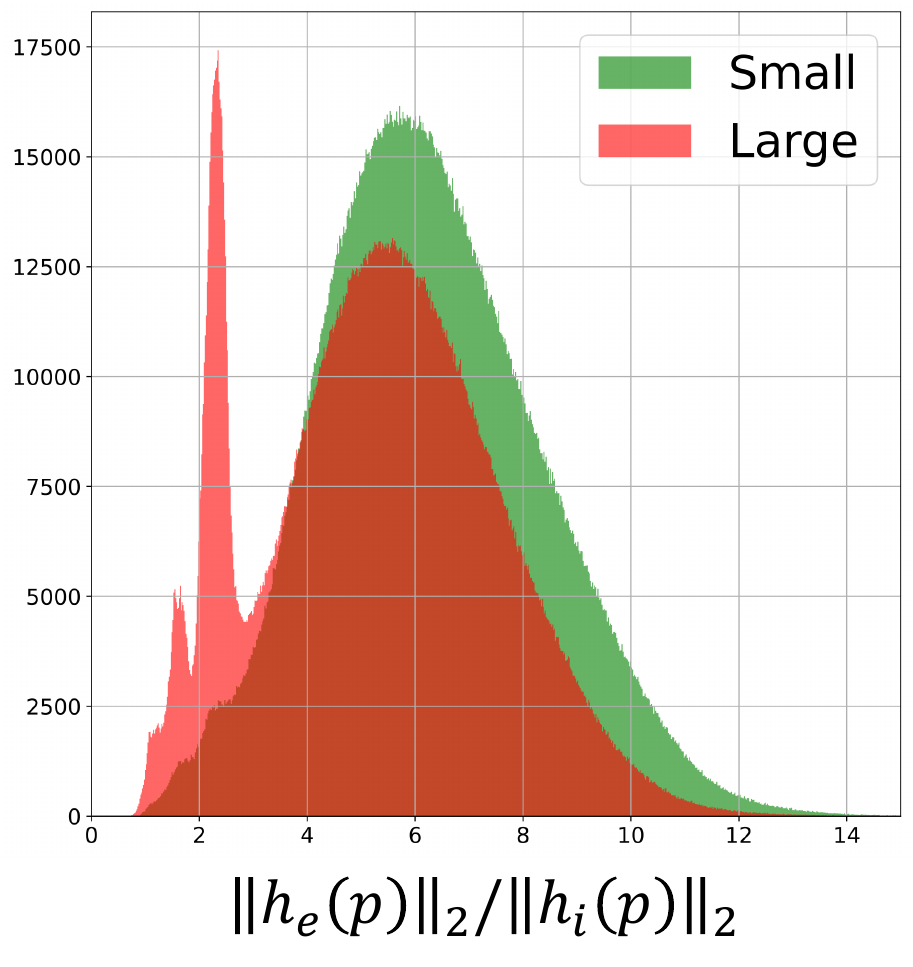} \label{fig:histogram_no}}
    \subfigure[\normalsize{with $L_{ts,in}$}]{\includegraphics[width=0.19\textwidth]{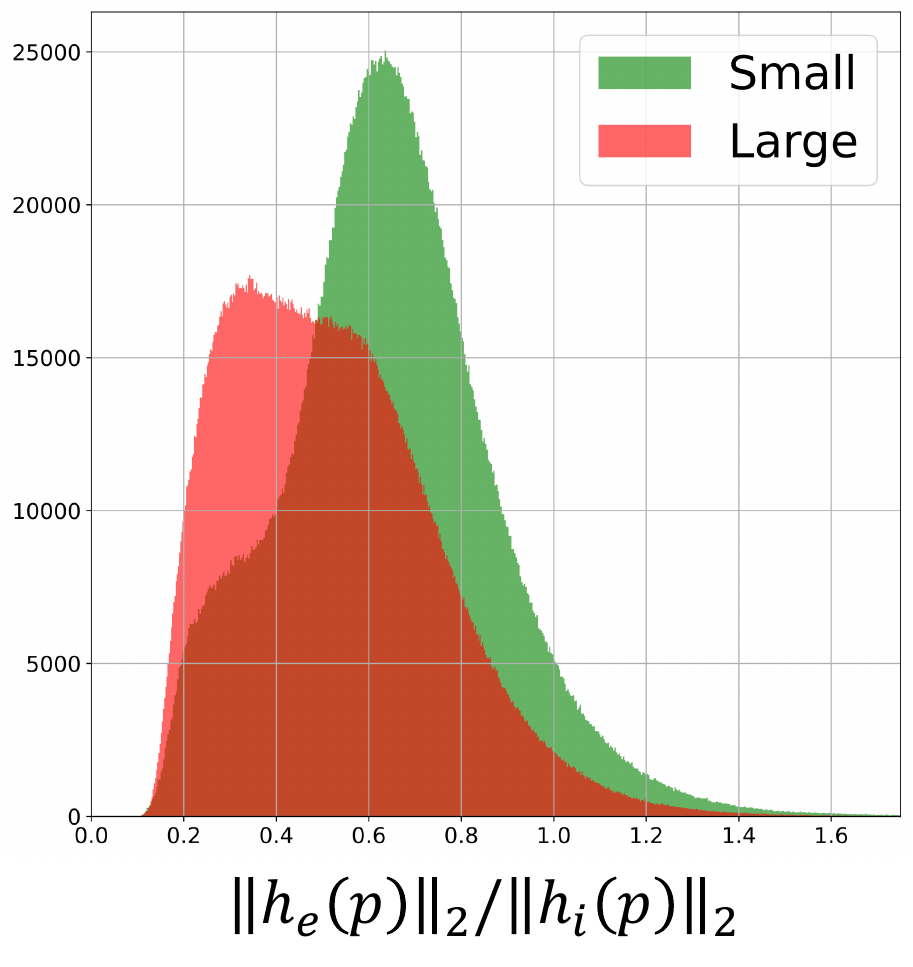} \label{fig:histogram_in}}
    \subfigure[\normalsize{with $L_{ts,out}$}]{\includegraphics[width=0.19\textwidth]{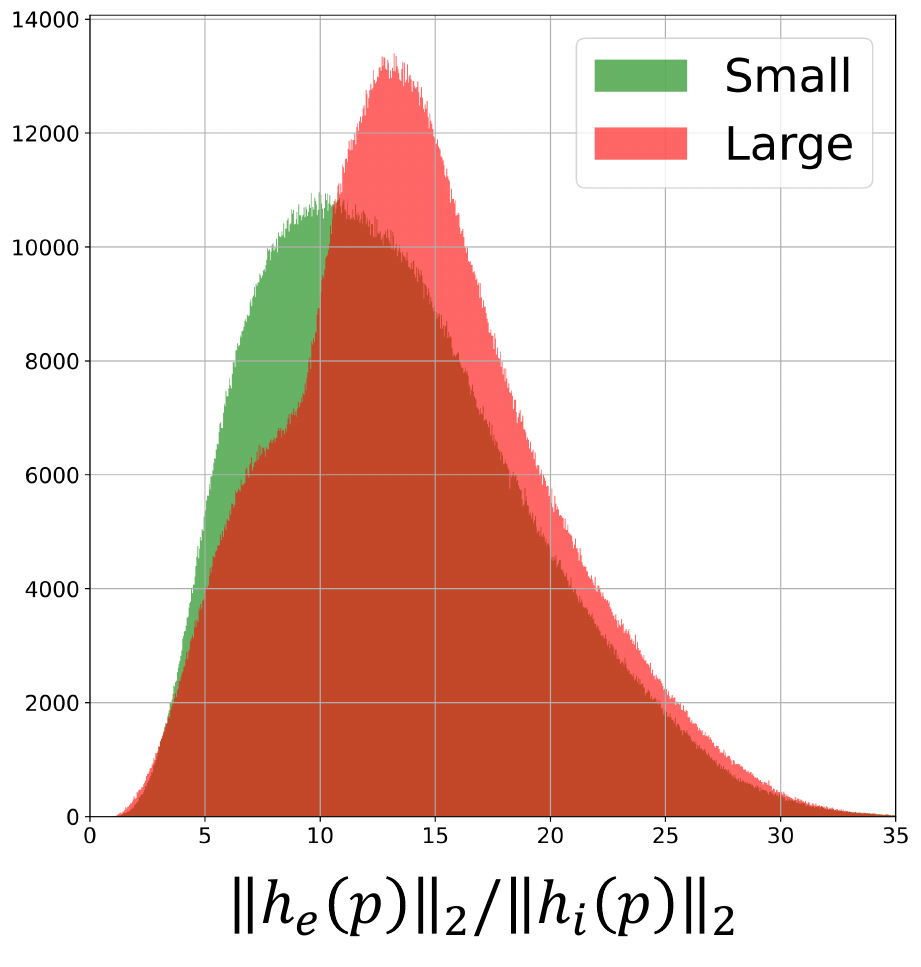} \label{fig:histogram_out}}
    \subfigure[\normalsize{with $L_{ts}$}]{\includegraphics[width=0.19\textwidth]{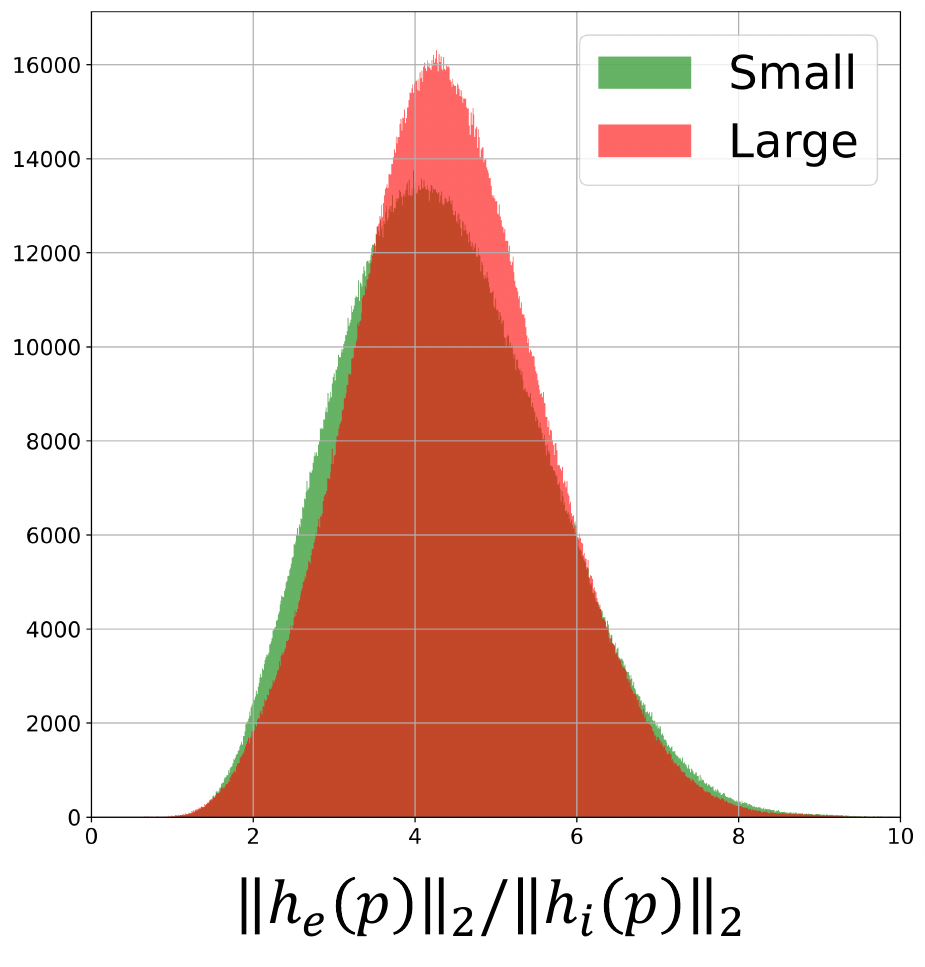} \label{fig:histogram_ours}} 

    \vspace{-0.35cm}
    
    \caption{\textbf{Histogram of $||h_e(p)||_2/ ||h_i(p)||_2$ on the small and large split of RealEstate10K dataset.}}

    \vspace{-0.45cm}
    
    \label{fig:histogram_ablation}
\end{figure*}

\subsection{Qualitative Results on Sequential Generation}

To validate the capability of our method in 3D consistent view synthesis, we illustrate qualitative results on Realestate10K with continuous camera movement in Fig.~\ref{fig_qual_sequential}. This figure specifically showcases the ability to maintain seen and unseen content consistency across different viewpoints, underlining the model's proficiency in generating coherent 3D spaces. The visual sequence highlights the seamless transition between frames, affirming the method's effectiveness in producing 3D-consistent views with high fidelity.

\subsection{Dependency Analysis between two Renderers}

Our proposed architecture exploits the implicit and explicit renderers and mixes their outputs for decoding view synthesis results.
To understand the dependency between two renderers, we analyze the norm of output feature maps.
For a spatial position $p$, the norm ratio of two spatial features $||h_e(p)||_2/ ||h_i(p)||_2$ can represent how much depends on the explicit feature $h_{e}(p)$ compared to implicit feature $h_{i}(p)$. 
For example, if the ratio is large, the model depends on the explicit renderer more than the implicit renderer at position $p$.
We compare histograms of the norm ratio by changing the components of $L_{ts}$ and data splits as shown in Fig.~\ref{fig:histogram_ablation}.

Figure~\ref{fig:histogram_overall} depicts that using $L_{ts,out}$ and $L_{ts,in}$ tends to make the model more dependent on explicit and implicit features, respectively, compared to our method trained without $L_{ts}$.
Furthermore, these tendencies are more apparent in difficult cases (i.e., large split) as shown in Fig.~\ref{fig:histogram_in}--\ref{fig:histogram_out}.
From our observations, we conjecture that guiding only a specific renderer improves the discriminability of that renderer, resulting in the model depending on the improved renderer.
Surprisingly, the model trained on all components of $L_{ts}$ makes a balanced usage of both renderers, and there is less bias in norm ratio even according to data splits as shown in Fig. \ref{fig:histogram_ours}.

\begin{figure}[t!]
    \centering
    \subfigure[\scriptsize{Input Image}]{\includegraphics[width=0.115\textwidth]{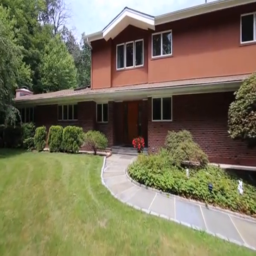}}
    \subfigure[\scriptsize{Warp Image}]{\includegraphics[width=0.115\textwidth]{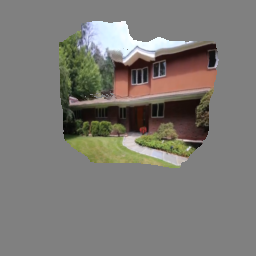}}
    \subfigure[\scriptsize{Without $L_{ts}$}]{\includegraphics[width=0.115\textwidth]{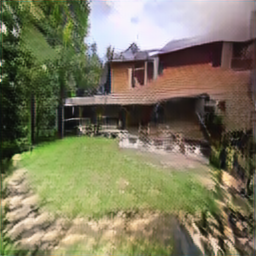}}
    \subfigure[\scriptsize{With $L_{ts}$}]{\includegraphics[width=0.115\textwidth]{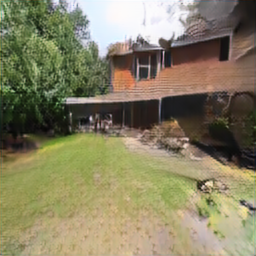}}

    \vspace{-0.3cm}
    
    \caption{\textbf{Visual ablation study.} Without the transformation similarity loss, our model complete textured out-of-view regions but not realistic enough than our model trained with the transformation similarity loss.}
    \vspace{-0.3cm}
    \label{fig:example_hist}
\end{figure}

\begin{figure}[t!]
    \centering
    
    {\parbox{0.115\textwidth}{\centering { \quad }}}
    {\parbox{0.115\textwidth}{\centering {\scriptsize Out-of-View(63\%)}}}
    {\parbox{0.115\textwidth}{\centering {\scriptsize PSNR-vis: 13.25} }}
    {\parbox{0.115\textwidth}{\centering {\scriptsize PSNR-vis: 12.84} }}

    \vspace{-.15cm}
    
    \subfigure[{\scriptsize Input Image}]{\includegraphics[width=0.115\textwidth]{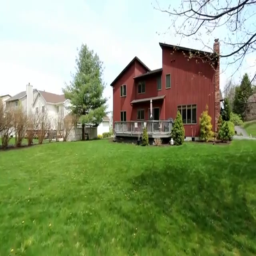}}
    \subfigure[{\scriptsize Warped Image}]{\includegraphics[width=0.115\textwidth]{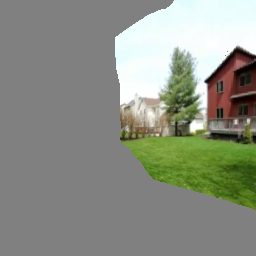}}
    \subfigure[{\scriptsize SynSin~\cite{wiles2020synsin}}]{\includegraphics[width=0.115\textwidth]{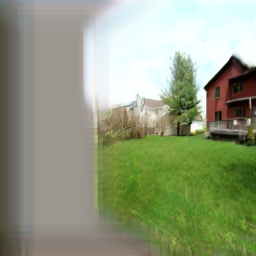}}
    \subfigure[{\scriptsize PixelSynth~\cite{rockwell2021pixelsynth}}]{\includegraphics[width=0.115\textwidth]{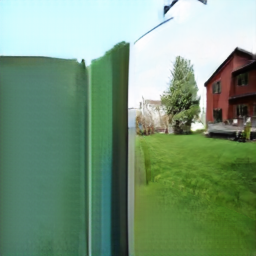}}

    \vspace{.05cm}
    
    {\parbox{0.115\textwidth}{\centering {\scriptsize PSNR-vis: 14.43} }}
    {\parbox{0.115\textwidth}{\centering {\scriptsize PSNR-vis: 11.46} }}
    {\parbox{0.115\textwidth}{\centering {\scriptsize PSNR-vis: 15.21} }}
    {\parbox{0.115\textwidth}{\centering { \quad }}}

    \vspace{-.12cm}
    
    \subfigure[{\scriptsize GeoFree~\cite{rombach2021geometry}}]{\includegraphics[width=0.115\textwidth]{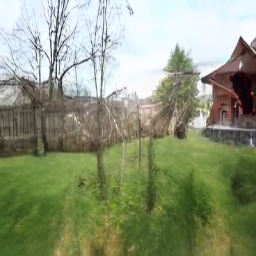}}
    \subfigure[{\scriptsize LookOut~\cite{ren2022look}}]{\includegraphics[width=0.115\textwidth]{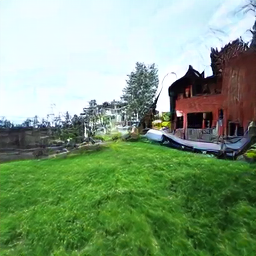}}
    \subfigure[{\scriptsize Ours}]{\includegraphics[width=0.115\textwidth]{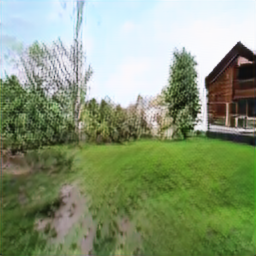}}
    \subfigure[{\scriptsize Ground Truth}]{\includegraphics[width=0.115\textwidth]{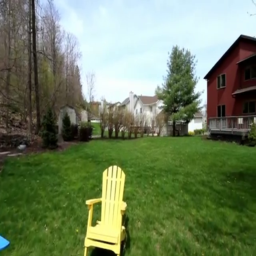}}

    \vspace{-0.3cm}
    
    \caption{\textbf{An example for synthesizing outdoor scenes on the indoor dataset RealEstate10K~\cite{zhou2018stereo}.}}
    \vspace{-0.5cm}
    \label{fig:real_outdoor}
\end{figure}

The effectiveness of our transformation similarity loss is confirmed by comparing it to our method that is trained without $L_{ts}$.
Figure~\ref{fig:histogram_no} shows that our model trained without $L_{ts}$ has some outliers for large view changes despite there being less bias according to data splits.
We observe these outliers are derived when the model fails to generate realistic out-of-view regions, especially in challenging settings, such as the network having to create novel views for both indoor and outdoor scenes, as shown in Fig.~\ref{fig:example_hist}.
We also confirm that our model trained with $L_{ts}$ performs well even in extreme cases, informing that $L_{ts}$ improves two renderers to embed discriminative features.
Figure~\ref{fig:real_outdoor} illustrates that explicit methods fail to generate the outdoor scene, but our method does as well as the implicit methods.
Collectively, $L_{ts}$ improves the discriminability of output features from two renderers and makes the behavior of the model stable, resulting in alleviating the seesaw problem.

\section{Conclusion}

We have introduced a single-image view synthesis framework by bridging explicit and implicit renderers.
Despite using autoregressive models, previous methods still suffer from the seesaw problem since they use only one explicit or implicit geometric transformation.
Thus, we design two parallel renderers to mitigate the problem and complement renderers with transformation similarity loss.
Alleviating the seesaw problem allows the network to generate novel view images better than previous methods, even with a non-autoregressive structure.
We note that the effectiveness of bridging two renderers can be applied in other tasks, such as extrapolation.
We believe that our work can prompt refocusing on non-autoregressive architecture for single-image view synthesis.

{\small
\bibliographystyle{IEEETrans}
\bibliography{egbib}

\begin{thebibliography}{10}
\providecommand{\url}[1]{#1}
\csname url@samestyle\endcsname
\providecommand{\newblock}{\relax}
\providecommand{\bibinfo}[2]{#2}
\providecommand{\BIBentrySTDinterwordspacing}{\spaceskip=0pt\relax}
\providecommand{\BIBentryALTinterwordstretchfactor}{4}
\providecommand{\BIBentryALTinterwordspacing}{\spaceskip=\fontdimen2\font plus
\BIBentryALTinterwordstretchfactor\fontdimen3\font minus
  \fontdimen4\font\relax}
\providecommand{\BIBforeignlanguage}[2]{{%
\expandafter\ifx\csname l@#1\endcsname\relax
\typeout{** WARNING: IEEEtranS.bst: No hyphenation pattern has been}%
\typeout{** loaded for the language `#1'. Using the pattern for}%
\typeout{** the default language instead.}%
\else
\language=\csname l@#1\endcsname
\fi
#2}}
\providecommand{\BIBdecl}{\relax}
\BIBdecl

\bibitem{aliev2020neural}
K.-A. Aliev, A.~Sevastopolsky, M.~Kolos, D.~Ulyanov, and V.~Lempitsky, ``Neural
  point-based graphics,'' in \emph{European Conference on Computer
  Vision}.\hskip 1em plus 0.5em minus 0.4em\relax Springer, 2020, pp. 696--712.

\bibitem{bao2019depth}
W.~Bao, W.-S. Lai, C.~Ma, X.~Zhang, Z.~Gao, and M.-H. Yang, ``Depth-aware video
  frame interpolation,'' in \emph{Proceedings of the IEEE/CVF Conference on
  Computer Vision and Pattern Recognition}, 2019, pp. 3703--3712.

\bibitem{brock2018large}
A.~Brock, J.~Donahue, and K.~Simonyan, ``Large scale gan training for high
  fidelity natural image synthesis,'' \emph{arXiv preprint arXiv:1809.11096},
  2018.

\bibitem{chen2021mvsnerf}
A.~Chen, Z.~Xu, F.~Zhao, X.~Zhang, F.~Xiang, J.~Yu, and H.~Su, ``Mvsnerf: Fast
  generalizable radiance field reconstruction from multi-view stereo,'' in
  \emph{Proceedings of the IEEE/CVF International Conference on Computer
  Vision}, 2021, pp. 14\,124--14\,133.

\bibitem{chen1993view}
S.~E. Chen and L.~Williams, ``View interpolation for image synthesis,'' in
  \emph{Proceedings of the 20th annual conference on Computer graphics and
  interactive techniques}, 1993, pp. 279--288.

\bibitem{chen2019monocular}
X.~Chen, J.~Song, and O.~Hilliges, ``Monocular neural image based rendering
  with continuous view control,'' in \emph{Proceedings of the IEEE/CVF
  International Conference on Computer Vision}, 2019, pp. 4090--4100.

\bibitem{chen2019learning}
Z.~Chen and H.~Zhang, ``Learning implicit fields for generative shape
  modeling,'' in \emph{Proceedings of the IEEE/CVF Conference on Computer
  Vision and Pattern Recognition}, 2019, pp. 5939--5948.

\bibitem{debevec1998efficient}
P.~Debevec, Y.~Yu, and G.~Borshukov, ``Efficient view-dependent image-based
  rendering with projective texture-mapping,'' in \emph{Eurographics Workshop
  on Rendering Techniques}.\hskip 1em plus 0.5em minus 0.4em\relax Springer,
  1998, pp. 105--116.

\bibitem{debevec1996modeling}
P.~E. Debevec, C.~J. Taylor, and J.~Malik, ``Modeling and rendering
  architecture from photographs: A hybrid geometry-and image-based approach,''
  in \emph{Proceedings of the 23rd annual conference on Computer graphics and
  interactive techniques}, 1996, pp. 11--20.

\bibitem{devlin2018bert}
J.~Devlin, M.-W. Chang, K.~Lee, and K.~Toutanova, ``Bert: Pre-training of deep
  bidirectional transformers for language understanding,'' \emph{arXiv preprint
  arXiv:1810.04805}, 2018.

\bibitem{diebel2006representing}
J.~Diebel, ``Representing attitude: Euler angles, unit quaternions, and
  rotation vectors,'' \emph{Matrix}, vol.~58, no. 15-16, pp. 1--35, 2006.

\bibitem{dosovitskiy2020image}
A.~Dosovitskiy, L.~Beyer, A.~Kolesnikov, D.~Weissenborn, X.~Zhai,
  T.~Unterthiner, M.~Dehghani, M.~Minderer, G.~Heigold, S.~Gelly \emph{et~al.},
  ``An image is worth 16x16 words: Transformers for image recognition at
  scale,'' \emph{arXiv preprint arXiv:2010.11929}, 2020.

\bibitem{esser2021taming}
P.~Esser, R.~Rombach, and B.~Ommer, ``Taming transformers for high-resolution
  image synthesis,'' in \emph{Proceedings of the IEEE/CVF Conference on
  Computer Vision and Pattern Recognition}, 2021, pp. 12\,873--12\,883.

\bibitem{flynn2019deepview}
J.~Flynn, M.~Broxton, P.~Debevec, M.~DuVall, G.~Fyffe, R.~Overbeck, N.~Snavely,
  and R.~Tucker, ``Deepview: View synthesis with learned gradient descent,'' in
  \emph{Proceedings of the IEEE/CVF Conference on Computer Vision and Pattern
  Recognition}, 2019, pp. 2367--2376.

\bibitem{godard2017unsupervised}
C.~Godard, O.~Mac~Aodha, and G.~J. Brostow, ``Unsupervised monocular depth
  estimation with left-right consistency,'' in \emph{Proceedings of the IEEE
  Conference on Computer Vision and Pattern Recognition}, 2017, pp. 270--279.

\bibitem{gortler1996lumigraph}
S.~J. Gortler, R.~Grzeszczuk, R.~Szeliski, and M.~F. Cohen, ``The lumigraph,''
  in \emph{Proceedings of the 23rd annual conference on Computer graphics and
  interactive techniques}, 1996, pp. 43--54.

\bibitem{guo2021pct}
M.-H. Guo, J.-X. Cai, Z.-N. Liu, T.-J. Mu, R.~R. Martin, and S.-M. Hu, ``Pct:
  Point cloud transformer,'' \emph{Computational Visual Media}, vol.~7, no.~2,
  pp. 187--199, 2021.

\bibitem{he2016deep}
K.~He, X.~Zhang, S.~Ren, and J.~Sun, ``Deep residual learning for image
  recognition,'' in \emph{Proceedings of the IEEE conference on computer vision
  and pattern recognition}, 2016, pp. 770--778.

\bibitem{hedman2018deep}
P.~Hedman, J.~Philip, T.~Price, J.-M. Frahm, G.~Drettakis, and G.~Brostow,
  ``Deep blending for free-viewpoint image-based rendering,'' \emph{ACM
  Transactions on Graphics (TOG)}, vol.~37, no.~6, pp. 1--15, 2018.

\bibitem{heusel2017gans}
M.~Heusel, H.~Ramsauer, T.~Unterthiner, B.~Nessler, and S.~Hochreiter, ``Gans
  trained by a two time-scale update rule converge to a local nash
  equilibrium,'' \emph{Advances in neural information processing systems},
  vol.~30, 2017.

\bibitem{hou2021novel}
Y.~Hou, A.~Solin, and J.~Kannala, ``Novel view synthesis via depth-guided skip
  connections,'' in \emph{Proceedings of the IEEE/CVF Winter Conference on
  Applications of Computer Vision}, 2021, pp. 3119--3128.

\bibitem{hu2021worldsheet}
R.~Hu, N.~Ravi, A.~C. Berg, and D.~Pathak, ``Worldsheet: Wrapping the world in
  a 3d sheet for view synthesis from a single image,'' in \emph{Proceedings of
  the IEEE/CVF International Conference on Computer Vision}, 2021, pp.
  12\,528--12\,537.

\bibitem{iizuka2017globally}
S.~Iizuka, E.~Simo-Serra, and H.~Ishikawa, ``Globally and locally consistent
  image completion,'' \emph{ACM Transactions on Graphics (ToG)}, vol.~36,
  no.~4, pp. 1--14, 2017.

\bibitem{jampani2021slide}
V.~Jampani, H.~Chang, K.~Sargent, A.~Kar, R.~Tucker, M.~Krainin, D.~Kaeser,
  W.~T. Freeman, D.~Salesin, B.~Curless \emph{et~al.}, ``Slide: Single image 3d
  photography with soft layering and depth-aware inpainting,'' in
  \emph{Proceedings of the IEEE/CVF International Conference on Computer
  Vision}, 2021, pp. 12\,518--12\,527.

\bibitem{khakhulin2022stereo}
T.~Khakhulin, D.~Korzhenkov, P.~Solovev, G.~Sterkin, A.-T. Ardelean, and
  V.~Lempitsky, ``Stereo magnification with multi-layer images,'' in
  \emph{Proceedings of the IEEE/CVF Conference on Computer Vision and Pattern
  Recognition}, 2022, pp. 8687--8696.

\bibitem{koh2023simple}
J.~Y. Koh, H.~Agrawal, D.~Batra, R.~Tucker, A.~Waters, H.~Lee, Y.~Yang,
  J.~Baldridge, and P.~Anderson, ``Simple and effective synthesis of indoor 3d
  scenes,'' in \emph{Proceedings of the AAAI Conference on Artificial
  Intelligence}, vol.~37, no.~1, 2023, pp. 1169--1178.

\bibitem{lee2019set}
J.~Lee, Y.~Lee, J.~Kim, A.~Kosiorek, S.~Choi, and Y.~W. Teh, ``Set transformer:
  A framework for attention-based permutation-invariant neural networks,'' in
  \emph{International Conference on Machine Learning}.\hskip 1em plus 0.5em
  minus 0.4em\relax PMLR, 2019, pp. 3744--3753.

\bibitem{levoy1996light}
M.~Levoy and P.~Hanrahan, ``Light field rendering,'' in \emph{Proceedings of
  the 23rd annual conference on Computer graphics and interactive techniques},
  1996, pp. 31--42.

\bibitem{li2022infinitenature}
Z.~Li, Q.~Wang, N.~Snavely, and A.~Kanazawa, ``Infinitenature-zero: Learning
  perpetual view generation of natural scenes from single images,'' in
  \emph{European Conference on Computer Vision}.\hskip 1em plus 0.5em minus
  0.4em\relax Springer, 2022, pp. 515--534.

\bibitem{lim2017geometric}
J.~H. Lim and J.~C. Ye, ``Geometric gan,'' \emph{arXiv preprint
  arXiv:1705.02894}, 2017.

\bibitem{liu2021infinite}
A.~Liu, R.~Tucker, V.~Jampani, A.~Makadia, N.~Snavely, and A.~Kanazawa,
  ``Infinite nature: Perpetual view generation of natural scenes from a single
  image,'' in \emph{Proceedings of the IEEE/CVF International Conference on
  Computer Vision}, 2021, pp. 14\,458--14\,467.

\bibitem{liu2019point2sequence}
X.~Liu, Z.~Han, Y.-S. Liu, and M.~Zwicker, ``Point2sequence: Learning the shape
  representation of 3d point clouds with an attention-based sequence to
  sequence network,'' in \emph{Proceedings of the AAAI Conference on Artificial
  Intelligence}, vol.~33, no.~01, 2019, pp. 8778--8785.

\bibitem{lombardi2019neural}
S.~Lombardi, T.~Simon, J.~Saragih, G.~Schwartz, A.~Lehrmann, and Y.~Sheikh,
  ``Neural volumes: Learning dynamic renderable volumes from images,''
  \emph{arXiv preprint arXiv:1906.07751}, 2019.

\bibitem{loshchilov2016sgdr}
I.~Loshchilov and F.~Hutter, ``Sgdr: Stochastic gradient descent with warm
  restarts,'' \emph{arXiv preprint arXiv:1608.03983}, 2016.

\bibitem{loshchilov2017decoupled}
------, ``Decoupled weight decay regularization,'' \emph{arXiv preprint
  arXiv:1711.05101}, 2017.

\bibitem{martin2018lookingood}
R.~Martin-Brualla, R.~Pandey, S.~Yang, P.~Pidlypenskyi, J.~Taylor, J.~Valentin,
  S.~Khamis, P.~Davidson, A.~Tkach, P.~Lincoln \emph{et~al.}, ``Lookingood:
  Enhancing performance capture with real-time neural re-rendering,''
  \emph{arXiv preprint arXiv:1811.05029}, 2018.

\bibitem{mescheder2019occupancy}
L.~Mescheder, M.~Oechsle, M.~Niemeyer, S.~Nowozin, and A.~Geiger, ``Occupancy
  networks: Learning 3d reconstruction in function space,'' in
  \emph{Proceedings of the IEEE/CVF conference on computer vision and pattern
  recognition}, 2019, pp. 4460--4470.

\bibitem{meshry2019neural}
M.~Meshry, D.~B. Goldman, S.~Khamis, H.~Hoppe, R.~Pandey, N.~Snavely, and
  R.~Martin-Brualla, ``Neural rerendering in the wild,'' in \emph{Proceedings
  of the IEEE/CVF Conference on Computer Vision and Pattern Recognition}, 2019,
  pp. 6878--6887.

\bibitem{mildenhall2020nerf}
B.~Mildenhall, P.~P. Srinivasan, M.~Tancik, J.~T. Barron, R.~Ramamoorthi, and
  R.~Ng, ``Nerf: Representing scenes as neural radiance fields for view
  synthesis,'' in \emph{European conference on computer vision}.\hskip 1em plus
  0.5em minus 0.4em\relax Springer, 2020, pp. 405--421.

\bibitem{niklaus2020softmax}
S.~Niklaus and F.~Liu, ``Softmax splatting for video frame interpolation,'' in
  \emph{Proceedings of the IEEE/CVF Conference on Computer Vision and Pattern
  Recognition}, 2020, pp. 5437--5446.

\bibitem{novotny2019perspectivenet}
D.~Novotny, B.~Graham, and J.~Reizenstein, ``Perspectivenet: A scene-consistent
  image generator for new view synthesis in real indoor environments,''
  \emph{Advances in Neural Information Processing Systems}, vol.~32, 2019.

\bibitem{olszewski2019transformable}
K.~Olszewski, S.~Tulyakov, O.~Woodford, H.~Li, and L.~Luo, ``Transformable
  bottleneck networks,'' in \emph{Proceedings of the IEEE/CVF International
  Conference on Computer Vision}, 2019, pp. 7648--7657.

\bibitem{park2021fast}
C.~Park, Y.~Jeong, M.~Cho, and J.~Park, ``Fast point transformer,'' \emph{arXiv
  preprint arXiv:2112.04702}, 2021.

\bibitem{park2019semantic}
T.~Park, M.-Y. Liu, T.-C. Wang, and J.-Y. Zhu, ``Semantic image synthesis with
  spatially-adaptive normalization,'' in \emph{Proceedings of the IEEE/CVF
  conference on computer vision and pattern recognition}, 2019, pp. 2337--2346.

\bibitem{radford2019language}
A.~Radford, J.~Wu, R.~Child, D.~Luan, D.~Amodei, I.~Sutskever \emph{et~al.},
  ``Language models are unsupervised multitask learners,'' \emph{OpenAI blog},
  vol.~1, no.~8, p.~9, 2019.

\bibitem{Ranftl2021}
R.~Ranftl, A.~Bochkovskiy, and V.~Koltun, ``Vision transformers for dense
  prediction,'' \emph{ArXiv preprint}, 2021.

\bibitem{ranftl2020towards}
R.~Ranftl, K.~Lasinger, D.~Hafner, K.~Schindler, and V.~Koltun, ``Towards
  robust monocular depth estimation: Mixing datasets for zero-shot
  cross-dataset transfer,'' \emph{IEEE transactions on pattern analysis and
  machine intelligence}, 2020.

\bibitem{razavi2019generating}
A.~Razavi, A.~Van~den Oord, and O.~Vinyals, ``Generating diverse high-fidelity
  images with vq-vae-2,'' \emph{Advances in neural information processing
  systems}, vol.~32, 2019.

\bibitem{ren2022look}
X.~Ren and X.~Wang, ``Look outside the room: Synthesizing a consistent
  long-term 3d scene video from a single image,'' in \emph{Proceedings of the
  IEEE/CVF Conference on Computer Vision and Pattern Recognition (CVPR)}, 2022.

\bibitem{rockwell2021pixelsynth}
C.~Rockwell, D.~F. Fouhey, and J.~Johnson, ``Pixelsynth: Generating a
  3d-consistent experience from a single image,'' in \emph{Proceedings of the
  IEEE/CVF International Conference on Computer Vision}, 2021, pp.
  14\,104--14\,113.

\bibitem{rombach2022high}
R.~Rombach, A.~Blattmann, D.~Lorenz, P.~Esser, and B.~Ommer, ``High-resolution
  image synthesis with latent diffusion models,'' in \emph{Proceedings of the
  IEEE/CVF conference on computer vision and pattern recognition}, 2022, pp.
  10\,684--10\,695.

\bibitem{rombach2021geometry}
R.~Rombach, P.~Esser, and B.~Ommer, ``Geometry-free view synthesis:
  Transformers and no 3d priors,'' in \emph{Proceedings of the IEEE/CVF
  International Conference on Computer Vision}, 2021, pp. 14\,356--14\,366.

\bibitem{sauer2021projected}
A.~Sauer, K.~Chitta, J.~M{\"u}ller, and A.~Geiger, ``Projected gans converge
  faster,'' \emph{Advances in Neural Information Processing Systems}, vol.~34,
  2021.

\bibitem{seitz2006comparison}
S.~M. Seitz, B.~Curless, J.~Diebel, D.~Scharstein, and R.~Szeliski, ``A
  comparison and evaluation of multi-view stereo reconstruction algorithms,''
  in \emph{2006 IEEE computer society conference on computer vision and pattern
  recognition (CVPR'06)}, vol.~1.\hskip 1em plus 0.5em minus 0.4em\relax IEEE,
  2006, pp. 519--528.

\bibitem{shepperd1978quaternion}
S.~W. Shepperd, ``Quaternion from rotation matrix,'' \emph{Journal of guidance
  and control}, vol.~1, no.~3, pp. 223--224, 1978.

\bibitem{shih20203d}
M.-L. Shih, S.-Y. Su, J.~Kopf, and J.-B. Huang, ``3d photography using
  context-aware layered depth inpainting,'' in \emph{Proceedings of the
  IEEE/CVF Conference on Computer Vision and Pattern Recognition}, 2020, pp.
  8028--8038.

\bibitem{sitzmann2019deepvoxels}
V.~Sitzmann, J.~Thies, F.~Heide, M.~Nie{\ss}ner, G.~Wetzstein, and
  M.~Zollhofer, ``Deepvoxels: Learning persistent 3d feature embeddings,'' in
  \emph{Proceedings of the IEEE/CVF Conference on Computer Vision and Pattern
  Recognition}, 2019, pp. 2437--2446.

\bibitem{solovev2023self}
P.~Solovev, T.~Khakhulin, and D.~Korzhenkov, ``Self-improving
  multiplane-to-layer images for novel view synthesis,'' in \emph{Proceedings
  of the IEEE/CVF Winter Conference on Applications of Computer Vision}, 2023,
  pp. 4309--4318.

\bibitem{srinivasan2019pushing}
P.~P. Srinivasan, R.~Tucker, J.~T. Barron, R.~Ramamoorthi, R.~Ng, and
  N.~Snavely, ``Pushing the boundaries of view extrapolation with multiplane
  images,'' in \emph{Proceedings of the IEEE/CVF Conference on Computer Vision
  and Pattern Recognition}, 2019, pp. 175--184.

\bibitem{srinivasan2017learning}
P.~P. Srinivasan, T.~Wang, A.~Sreelal, R.~Ramamoorthi, and R.~Ng, ``Learning to
  synthesize a 4d rgbd light field from a single image,'' in \emph{Proceedings
  of the IEEE International Conference on Computer Vision}, 2017, pp.
  2243--2251.

\bibitem{tatarchenko2016multi}
M.~Tatarchenko, A.~Dosovitskiy, and T.~Brox, ``Multi-view 3d models from single
  images with a convolutional network,'' in \emph{European Conference on
  Computer Vision}.\hskip 1em plus 0.5em minus 0.4em\relax Springer, 2016, pp.
  322--337.

\bibitem{Tucker_2020_CVPR}
R.~Tucker and N.~Snavely, ``Single-view view synthesis with multiplane
  images,'' in \emph{Proceedings of the IEEE/CVF Conference on Computer Vision
  and Pattern Recognition (CVPR)}, June 2020.

\bibitem{vaswani2017attention}
A.~Vaswani, N.~Shazeer, N.~Parmar, J.~Uszkoreit, L.~Jones, A.~N. Gomez,
  {\L}.~Kaiser, and I.~Polosukhin, ``Attention is all you need,''
  \emph{Advances in neural information processing systems}, vol.~30, 2017.

\bibitem{wang2018learning}
C.~Wang, J.~Miguel~Buenaposada, R.~Zhu, and S.~Lucey, ``Learning depth from
  monocular videos using direct methods,'' in \emph{Proceedings of the IEEE
  Conference on Computer Vision and Pattern Recognition}, 2018, pp. 2022--2030.

\bibitem{wang2021ibrnet}
Q.~Wang, Z.~Wang, K.~Genova, P.~P. Srinivasan, H.~Zhou, J.~T. Barron,
  R.~Martin-Brualla, N.~Snavely, and T.~Funkhouser, ``Ibrnet: Learning
  multi-view image-based rendering,'' in \emph{Proceedings of the IEEE/CVF
  Conference on Computer Vision and Pattern Recognition}, 2021, pp. 4690--4699.

\bibitem{wang2004image}
Z.~Wang, A.~C. Bovik, H.~R. Sheikh, and E.~P. Simoncelli, ``Image quality
  assessment: from error visibility to structural similarity,'' \emph{IEEE
  transactions on image processing}, vol.~13, no.~4, pp. 600--612, 2004.

\bibitem{watson2023novel}
\BIBentryALTinterwordspacing
D.~Watson, W.~Chan, R.~M. Brualla, J.~Ho, A.~Tagliasacchi, and M.~Norouzi,
  ``Novel view synthesis with diffusion models,'' in \emph{The Eleventh
  International Conference on Learning Representations}, 2023. [Online].
  Available: \url{https://openreview.net/forum?id=HtoA0oT30jC}
\BIBentrySTDinterwordspacing

\bibitem{wiles2020synsin}
O.~Wiles, G.~Gkioxari, R.~Szeliski, and J.~Johnson, ``Synsin: End-to-end view
  synthesis from a single image,'' in \emph{Proceedings of the IEEE/CVF
  Conference on Computer Vision and Pattern Recognition}, 2020, pp. 7467--7477.

\bibitem{xie2021segformer}
E.~Xie, W.~Wang, Z.~Yu, A.~Anandkumar, J.~M. Alvarez, and P.~Luo, ``Segformer:
  Simple and efficient design for semantic segmentation with transformers,''
  \emph{Advances in Neural Information Processing Systems}, vol.~34, 2021.

\bibitem{xie2018attentional}
S.~Xie, S.~Liu, Z.~Chen, and Z.~Tu, ``Attentional shapecontextnet for point
  cloud recognition,'' in \emph{Proceedings of the IEEE Conference on Computer
  Vision and Pattern Recognition}, 2018, pp. 4606--4615.

\bibitem{xu2022point}
Q.~Xu, Z.~Xu, J.~Philip, S.~Bi, Z.~Shu, K.~Sunkavalli, and U.~Neumann,
  ``Point-nerf: Point-based neural radiance fields,'' in \emph{Proceedings of
  the IEEE/CVF Conference on Computer Vision and Pattern Recognition}, 2022,
  pp. 5438--5448.

\bibitem{yang2019modeling}
J.~Yang, Q.~Zhang, B.~Ni, L.~Li, J.~Liu, M.~Zhou, and Q.~Tian, ``Modeling point
  clouds with self-attention and gumbel subset sampling,'' in \emph{Proceedings
  of the IEEE/CVF Conference on Computer Vision and Pattern Recognition}, 2019,
  pp. 3323--3332.

\bibitem{yu2021pixelnerf}
A.~Yu, V.~Ye, M.~Tancik, and A.~Kanazawa, ``pixelnerf: Neural radiance fields
  from one or few images,'' in \emph{Proceedings of the IEEE/CVF Conference on
  Computer Vision and Pattern Recognition}, 2021, pp. 4578--4587.

\bibitem{zhang2018unreasonable}
R.~Zhang, P.~Isola, A.~A. Efros, E.~Shechtman, and O.~Wang, ``The unreasonable
  effectiveness of deep features as a perceptual metric,'' in \emph{Proceedings
  of the IEEE conference on computer vision and pattern recognition}, 2018, pp.
  586--595.

\bibitem{zhao2021point}
H.~Zhao, L.~Jiang, J.~Jia, P.~H. Torr, and V.~Koltun, ``Point transformer,'' in
  \emph{Proceedings of the IEEE/CVF International Conference on Computer
  Vision}, 2021, pp. 16\,259--16\,268.

\bibitem{zhou2018stereo}
T.~Zhou, R.~Tucker, J.~Flynn, G.~Fyffe, and N.~Snavely, ``Stereo magnification:
  Learning view synthesis using multiplane images,'' \emph{arXiv preprint
  arXiv:1805.09817}, 2018.

\bibitem{zhou2016view}
T.~Zhou, S.~Tulsiani, W.~Sun, J.~Malik, and A.~A. Efros, ``View synthesis by
  appearance flow,'' in \emph{European conference on computer vision}.\hskip
  1em plus 0.5em minus 0.4em\relax Springer, 2016, pp. 286--301.

\bibitem{zitnick2004high}
C.~L. Zitnick, S.~B. Kang, M.~Uyttendaele, S.~Winder, and R.~Szeliski,
  ``High-quality video view interpolation using a layered representation,''
  \emph{ACM transactions on graphics (TOG)}, vol.~23, no.~3, pp. 600--608,
  2004.

\end{thebibliography}
}

\vspace{-10mm}

\begin{IEEEbiography}[{\includegraphics[width=1in,height=1.25in,clip,keepaspectratio]{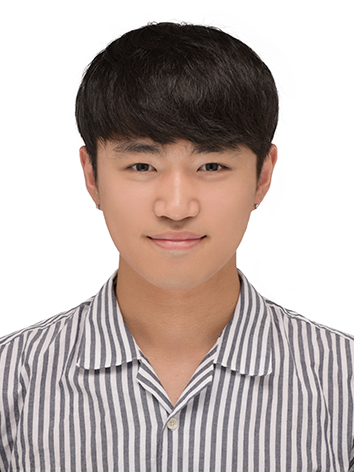}}]{Byeongjun Park}
received the B.S. degree in the School of Electrical Engineering from Korea Advanced Institute of Science and Technology (KAIST), Daejeon, South Korea, in 2020. Since 2020, he has been pursuing the Ph.D. degree in electrical engineering with the School of Electrical Engineering from KAIST. His current research interests include neural rendering, novel view synthesis, and diffusion models.
\end{IEEEbiography}

\vspace{-10mm}

\begin{IEEEbiography}[{\includegraphics[width=1in,height=1.25in,clip,keepaspectratio]{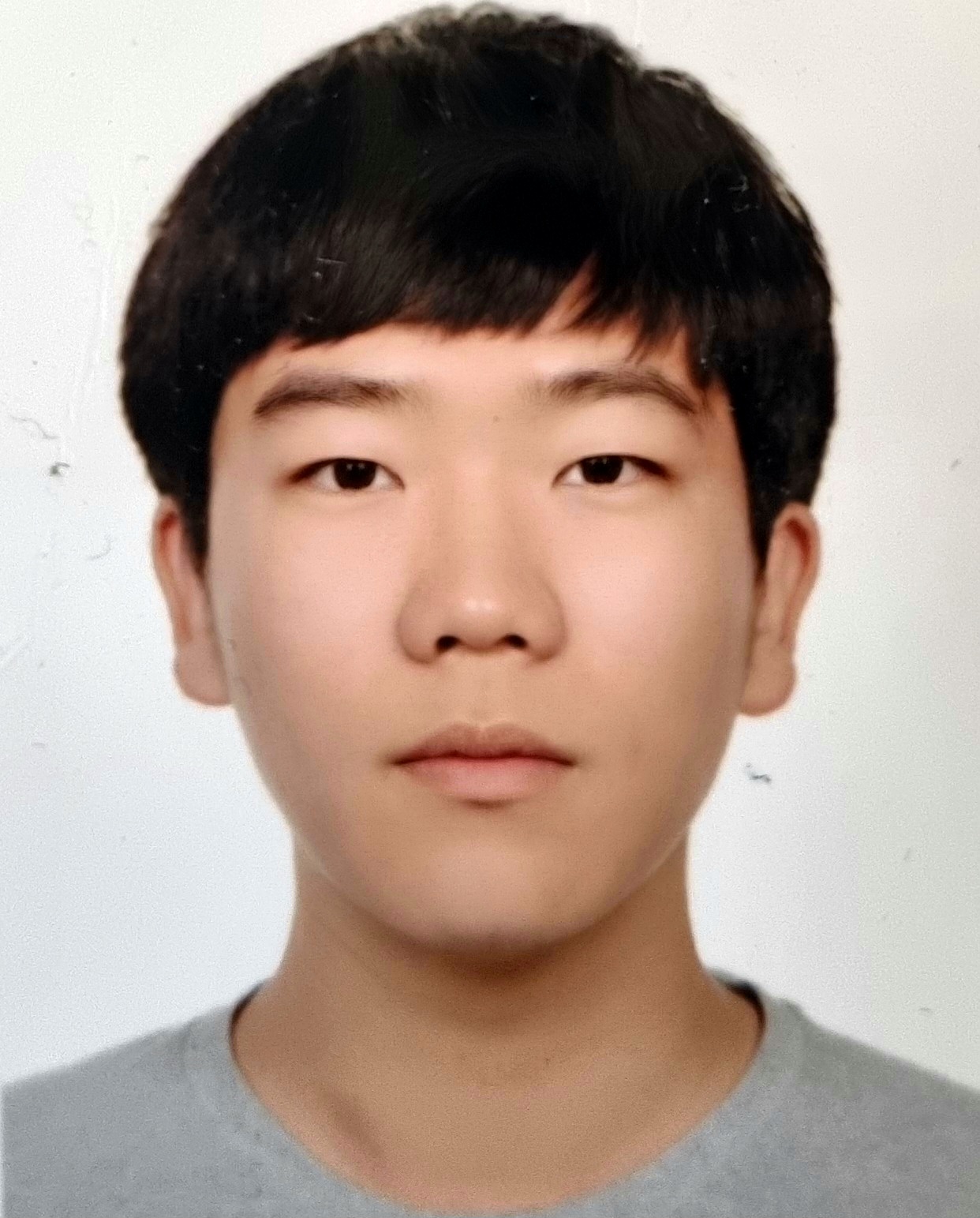}}]{Hyojun Go}
received the B.S. degree in the Department of Electrical Engineering from Hanyang University, Seoul, South Korea, in 2019, and the M.S. degree in the School of Electrical Engineering from Korea Advanced Institute of Science and Technology (KAIST), Daejeon, South Korea, in 2022. Since 2023, he has been working with  Twelvelabs, in Seoul, South Korea. His current research interests include generative models and Diffusion Models. 
\end{IEEEbiography}

\vspace{-10mm}

\begin{IEEEbiography}[{\includegraphics[width=1in,height=1.25in,clip,keepaspectratio]{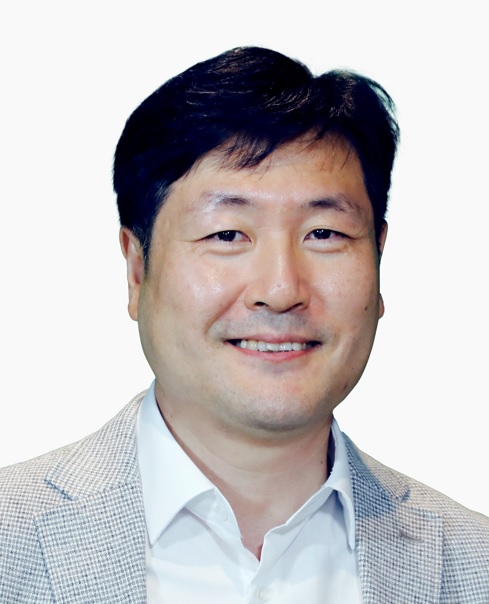}}]{Changick Kim}
received the B.S. degree in electrical engineering from Yonsei University, Seoul, South Korea, in 1989, the M.S. degree in electronics and electrical engineering from the Pohang University of Science and Technology (POSTECH), Pohang, South Korea, in 1991, and the Ph.D. degree in electrical engineering from the University of Washington, Seattle, WA, USA, in 2000. From 2000 to 2005, he was a Senior Member of Technical Staff with Epson Research and Development, Inc., Palo Alto, CA, USA. From 2005 to 2009, he was an Associate Professor at the School of Engineering, Information and Communications University, Daejeon, South Korea. Since March 2009, he has been with the School of Electrical Engineering, Korea Advanced Institute of Science and Technology (KAIST), Daejeon, Korea, where he is currently a Professor. He is also in charge of the center for Security Technology Research, KAIST. His research interests include few-shot learning, adversarial attack, and 3D reconstruction.
\end{IEEEbiography}

\appendices
\begin{table}[ht]
    \caption{\textbf{Notations.}}

    \vspace{-0.3cm}
    
    \resizebox{0.9\linewidth}{!}{
        \begin{tabular}{ll}
        \multicolumn{2}{l}{{\ul Images and Feature maps: }} \\ \\
            $I_{ref}$      & Input image \\
            $I_{tgt}$      & Generated target image \\
            $I_{gt}$       & Ground-truth target image \\
            $D$                                             & The estimated Depth Map from DepthNet\\
            $f_{i} $           & The $i$-th output point feature of the encoder \\
            $l^{i}_{local}$    & The continuous positional encodings of the $i$-th LSA layer \\
            $g^{i}_{global}$  & The $i$-th global set attention of the encoder \\
            $g^{i}_{local}$   & The $i$-th local set attention of the encoder \\
            $h_{i}$   & The output feature map of the implicit renderer \\
            $h_{e}$   & The output feature map of the explicit renderer \\     
            $\textbf{O}$   & The out-of-view mask \\
            
        \\ \multicolumn{2}{l}{{\ul  Camera parameters and Coordinates: }} \\ \\
            $K$ & Input camera intrinsic matrix for a resolution of $H \times W$ \\ 
            $T$ & Input relative camera pose matrix \\
            $R$ & The rotation matrix of $T$ \\
            $t$ & The translation vector of $T$ \\
            $\textbf{u}/\norm{\textbf{u}}$ & The normalized axis that is not changed by $R$ \\
            $\theta$ & The amount of rotated angle of $R$ \\
            $X_{img}$    & A set of normalized image coordinates \\
            $X_{w}$      & A set of 3D world coordinates \\
            $\mathcal{N}(p)$                & A set of neighbor homogeneous coordinates of $p$ \\
        \end{tabular}
    }
\vspace{-0.2cm}
    
\end{table}

\begin{table*}[!t]
\caption{\textbf{Types of baselines and our method.} Note that InfNat~\cite{liu2021infinite} varies according to the number of steps, so we mark it as \trianbox0{cyellow}.}
\vspace{-0.4cm}
    \centering
    \resizebox{0.9\linewidth}{!}{
    \begin{tabular}{cccccccccc}
\toprule
\multirow{2}{*}{Types} & \multicolumn{8}{c}{Methods} \\
\arrayrulecolor{gray}\cmidrule(lr){2-10}
                  & MV-3D~\cite{tatarchenko2016multi} & ViewApp~\cite{zhou2016view} &  SynSin~\cite{wiles2020synsin} &
                  InfNat~\cite{liu2021infinite} & PixelSynth~\cite{rockwell2021pixelsynth} & GeoFree~\cite{rombach2021geometry} & 
                  LookOut~\cite{ren2022look} & InfZero~\cite{li2022infinitenature} & Ours \\
\arrayrulecolor{black}\midrule
Explicit   & \xmark    & \cmark    & \cmark        & \cmark      & \cmark     & \xmark & \xmark & \cmark  & \cmark   \\
Implicit   &  \cmark   &   \xmark  & \xmark      &  \xmark   & \xmark  & \cmark & \cmark & \xmark & \cmark    \\ 
Autoregressive   &  \xmark    &    \xmark  & \xmark     &   \trianbox0{cyellow}   & \cmark     & \cmark & \cmark & \cmark & \xmark  \\
\bottomrule
\end{tabular}
}
\vspace{-0.3cm}
\label{table:method_types}
\end{table*}

\section{Baselines}

\crefformat{footnote}{#2\footnotemark[#1]#3}

We compared our method to previous single-image view synthesis methods, and
Table~\ref{table:method_types} briefly shows whether each method is an explicit, implicit, and autoregressive model.
In contrast to the previous methods, we employ a combination of explicit and implicit geometric transformations within a non-autoregressive model.

\subsection{SynSin~\cite{wiles2020synsin}}
SynSin~\cite{wiles2020synsin} uses a point cloud representation for single-image view synthesis.
Similar to our method, it does not require any ground-truth 3D information and uses a differentiable point cloud renderer.
The point cloud representation projected by the renderer is refined to generate novel view images.
We use the official code for implementation~\footnote{\label{synsin_git}\url{https://github.com/facebookresearch/synsin}}.
SynSin-6x, which is a variant of SynSin trained on large viewpoint changes, is introduced in~\cite{rockwell2021pixelsynth}.
For implementation of SynSin-6x, we adopt the official code of PixelSynth~\cite{rockwell2021pixelsynth}~\footnote{\label{pixelsynth_git} \url{https://github.com/crockwell/pixelsynth}}.

\subsection{PixelSynth~\cite{rockwell2021pixelsynth}}
SynSin achieves remarkable view synthesis results in small viewpoint changes, but it fails to fill the unseen region of novel view images realistically.
PixelSynth utilizes the outpainting strategy to supplement the ability to complete the unseen region of SynSin.
Although a slow autoregressive model is used for outpainting, PixelSynth still performs poorly in filling the out-of-view pixels.
The official code is publicly available, and we utilize it for implementation~\cref{pixelsynth_git}.

\subsection{GeoFree~\cite{rombach2021geometry}}
With the powerful transformer and autoregressive model, GeoFree~\cite{rombach2021geometry} shows that the model can learn the 3D transformation needed for the single-image view synthesis. 
Its view synthesis results are realistic, but it fails to maintain the seen contents.
We adopt the official code for implementation~\footnote{\url{https://github.com/CompVis/geometry-free-view-synthesis}}.

\subsection{MV-3D~\cite{tatarchenko2016multi} and ViewApp~\cite{zhou2016view}}
MV-3D~\cite{tatarchenko2016multi} uses a convolutional neural network to predict an RGB image and a depth map for arbitrary viewpoint. 
ViewApp~\cite{zhou2016view} predicts the flow and warps the reference image to the target view with this flow.
For both methods, we adopt the implementation of SynSin~\cite{wiles2020synsin}~\cref{synsin_git}.

\subsection{InfNat~\cite{liu2021infinite}}
Infinite Nature~\cite{liu2021infinite} focuses on nature scenes and generates a video from an image and a camera trajectory.
InfNat uses a pretrained MiDAS~\cite{Ranftl2021} to estimate depth maps, and novel views are generated based on explicit geometric transformations.
We evaluate the performance for 1-step (i.e., direct generation) and 5-step (i.e., gradual generation for target view).
We adopt the official code for implementation~\footnote{\label{infnat_git}\url{https://github.com/google-research/google-research/tree/master/infinite_nature}}.

\subsection{LookOut~\cite{ren2022look}}
Ren~\etal~\cite{ren2022look} focus on long-term view synthesis with the autoregressive model.
Novel views are generated time-sequentially, which takes more generation time than GeoFree~\cite{rombach2021geometry}.
LookOut utilizes a pretrained encoder-decoder in GeoFree~\cite{rombach2021geometry} for mapping the images to tokens.
We adopt the official code for implementation~\footnote{\label{lookout_git}\url{https://github.com/xrenaa/Look-Outside-Room}}.

\subsection{InfZero~\cite{li2022infinitenature}}
Li~\etal~\cite{li2022infinitenature} focus on long-term view synthesis for nature scenes.
Similar to InfNat~\cite{liu2021infinite}, novel views are generated based on explicit geometric transformations.
We evaluate the performance of the 5-step generation for the target view.
We adopt the official code for implementation~\footnote{\label{infzero_git}\url{https://github.com/google-research/google-research/tree/master/infinite_nature_zero}}.

\begin{figure*}[t!]
    \centering
    {\parbox{0.12\textwidth}{\centering {}}}
    {\parbox{0.12\textwidth}{\centering {\scriptsize Out-of-View(31\%)}}}
    {\parbox{0.12\textwidth}{\centering {\scriptsize PSNR-vis: 17.20}}}
    {\parbox{0.12\textwidth}{\centering {\scriptsize PSNR-vis: 17.34}}}
    {\parbox{0.12\textwidth}{\centering {\scriptsize PSNR-vis: 16.79}}}
    {\parbox{0.12\textwidth}{\centering {\scriptsize PSNR-vis: 11.15}}}
    {\parbox{0.12\textwidth}{\centering {\scriptsize PSNR-vis: 18.95}}}
    {\parbox{0.12\textwidth}{\centering {}}}
    
    \includegraphics[width=0.12\textwidth]{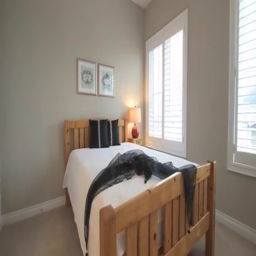}
    \includegraphics[width=0.12\textwidth]{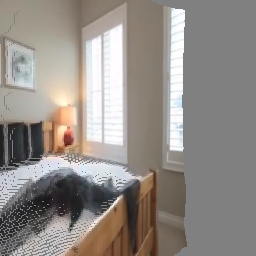}
    \includegraphics[width=0.12\textwidth]{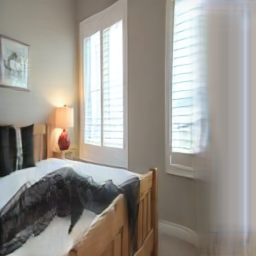}
    \includegraphics[width=0.12\textwidth]{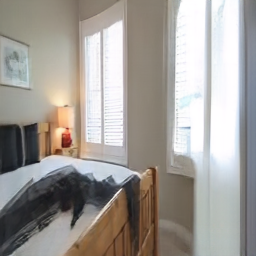}
    \includegraphics[width=0.12\textwidth]{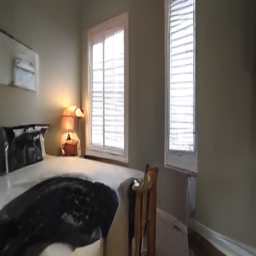}
    \includegraphics[width=0.12\textwidth]{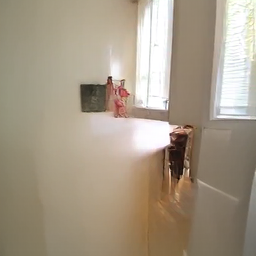}
    \includegraphics[width=0.12\textwidth]{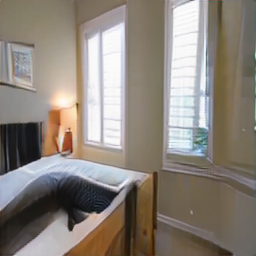}
    \includegraphics[width=0.12\textwidth]{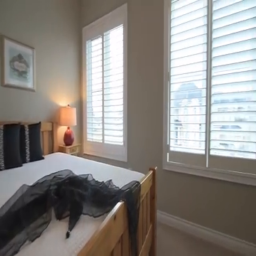}
     
    {\parbox{0.12\textwidth}{\centering {}}}
    {\parbox{0.12\textwidth}{\centering {\scriptsize Out-of-View(35\%)}}}
    {\parbox{0.12\textwidth}{\centering {\scriptsize PSNR-vis: 16.27}}}
    {\parbox{0.12\textwidth}{\centering {\scriptsize PSNR-vis: 16.34}}}
    {\parbox{0.12\textwidth}{\centering {\scriptsize PSNR-vis: 14.23}}}
    {\parbox{0.12\textwidth}{\centering {\scriptsize PSNR-vis: 11.83}}}
    {\parbox{0.12\textwidth}{\centering {\scriptsize PSNR-vis: 16.98}}}
    {\parbox{0.12\textwidth}{\centering {}}}

    \includegraphics[width=0.12\textwidth]{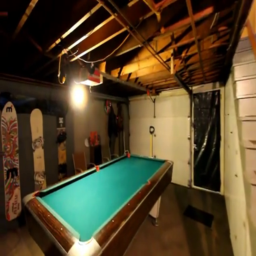}
    \includegraphics[width=0.12\textwidth]{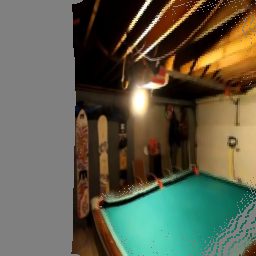}
    \includegraphics[width=0.12\textwidth]{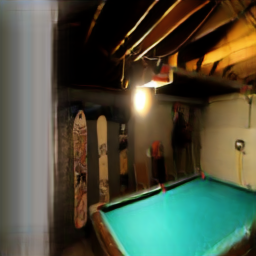}
    \includegraphics[width=0.12\textwidth]{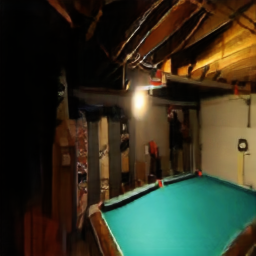}
    \includegraphics[width=0.12\textwidth]{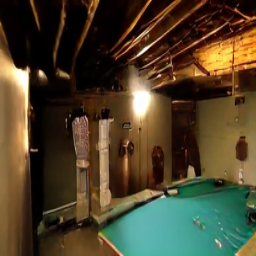}
    \includegraphics[width=0.12\textwidth]{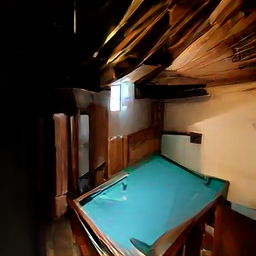}
    \includegraphics[width=0.12\textwidth]{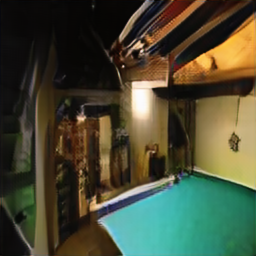}
    \includegraphics[width=0.12\textwidth]{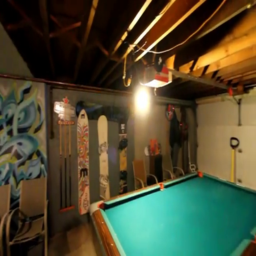}
      
    {\parbox{0.12\textwidth}{\centering {}}}
    {\parbox{0.12\textwidth}{\centering {\scriptsize Out-of-View(46\%)}}}
    {\parbox{0.12\textwidth}{\centering {\scriptsize PSNR-vis: 14.61}}}
    {\parbox{0.12\textwidth}{\centering {\scriptsize PSNR-vis: 14.68}}}
    {\parbox{0.12\textwidth}{\centering {\scriptsize PSNR-vis: 11.67}}}
    {\parbox{0.12\textwidth}{\centering {\scriptsize PSNR-vis: 10.29}}}
    {\parbox{0.12\textwidth}{\centering {\scriptsize PSNR-vis: 15.92}}}
    {\parbox{0.12\textwidth}{\centering {}}}

    \includegraphics[width=0.12\textwidth]{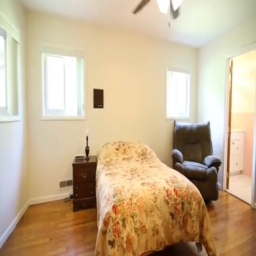}
    \includegraphics[width=0.12\textwidth]{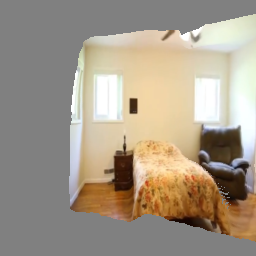}
    \includegraphics[width=0.12\textwidth]{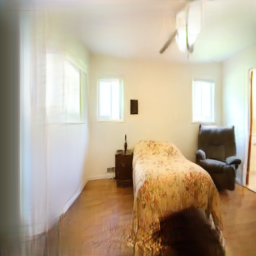}
    \includegraphics[width=0.12\textwidth]{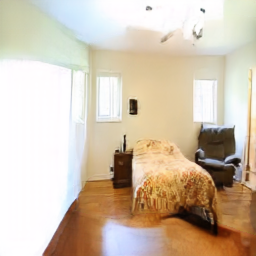}
    \includegraphics[width=0.12\textwidth]{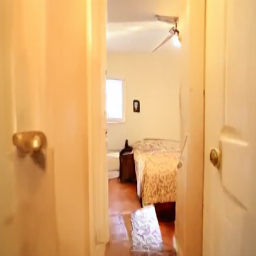}
    \includegraphics[width=0.12\textwidth]{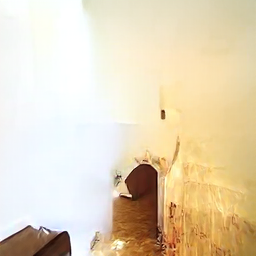}
    \includegraphics[width=0.12\textwidth]{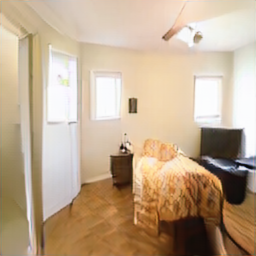}
    \includegraphics[width=0.12\textwidth]{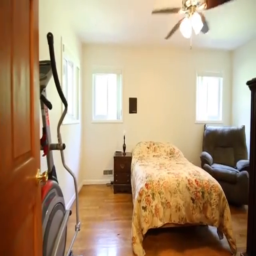}

    {\parbox{0.12\textwidth}{\centering {}}}
    {\parbox{0.12\textwidth}{\centering {\scriptsize Out-of-View(69\%)}}}
    {\parbox{0.12\textwidth}{\centering {\scriptsize PSNR-vis: 15.37}}}
    {\parbox{0.12\textwidth}{\centering {\scriptsize PSNR-vis: 15.22}}}
    {\parbox{0.12\textwidth}{\centering {\scriptsize PSNR-vis: 13.54}}}
    {\parbox{0.12\textwidth}{\centering {\scriptsize PSNR-vis: 13.33}}}
    {\parbox{0.12\textwidth}{\centering {\scriptsize PSNR-vis: 15.47}}}
    {\parbox{0.12\textwidth}{\centering {}}}

    \vspace{-.15cm}
    
    \subfigure[{\scriptsize Input Image}]{\includegraphics[width=0.12\textwidth]{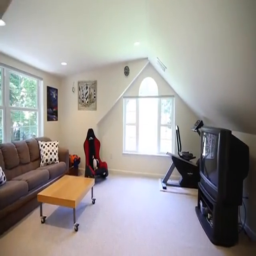}}
    \subfigure[{\scriptsize Warped Image}]{\includegraphics[width=0.12\textwidth]{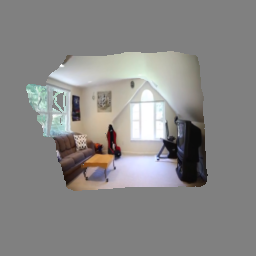}}
    \subfigure[{\scriptsize SynSin~\cite{wiles2020synsin}}]{\includegraphics[width=0.12\textwidth]{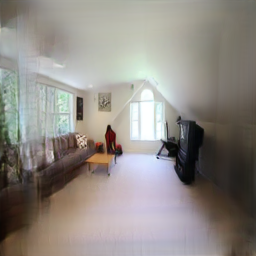}}
    \subfigure[{\scriptsize PixelSynth~\cite{rockwell2021pixelsynth}}]{\includegraphics[width=0.12\textwidth]{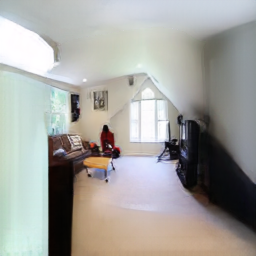}}
    \subfigure[{\scriptsize GeoFree~\cite{rombach2021geometry}}]{\includegraphics[width=0.12\textwidth]{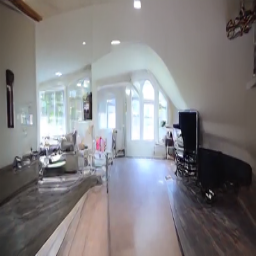}}
    \subfigure[{\scriptsize LookOut~\cite{ren2022look}}]{\includegraphics[width=0.12\textwidth]{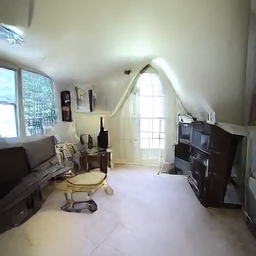}}
    \subfigure[{\scriptsize Ours}]{\includegraphics[width=0.12\textwidth]{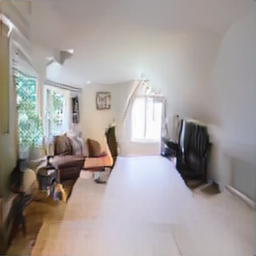}}
    \subfigure[{\scriptsize Ground Truth}]{\includegraphics[width=0.12\textwidth]{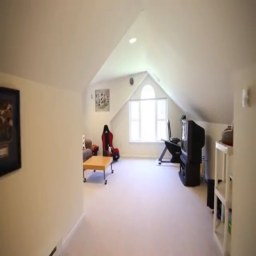}} 
    
     \vspace{-.3cm}
     
    \caption{\textbf{Qualitative Results on RealEstate10K~\cite{zhou2018stereo}.}}
    \label{fig:qual_real}
    \vspace{-0.3cm}
\end{figure*}

\begin{figure*}[t!]
    \centering
    
    {\parbox{0.105\textwidth}{\centering {}}}
    {\parbox{0.105\textwidth}{\centering {\notsotiny Out-of-View(29\%)}}}
    {\parbox{0.105\textwidth}{\centering {\notsotiny PSNR-vis: 21.13}}}
    {\parbox{0.105\textwidth}{\centering {\notsotiny PSNR-vis: 21.61}}}
    {\parbox{0.105\textwidth}{\centering {\notsotiny PSNR-vis: 14.63}}}
    {\parbox{0.105\textwidth}{\centering {\notsotiny PSNR-vis: 18.45}}}
    {\parbox{0.105\textwidth}{\centering {\notsotiny PSNR-vis: 9.31}}}
    {\parbox{0.105\textwidth}{\centering {\notsotiny PSNR-vis: 22.23}}}
    {\parbox{0.105\textwidth}{\centering {}}}
    
    \includegraphics[width=0.105\textwidth]{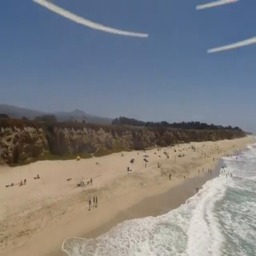}
    \includegraphics[width=0.105\textwidth]{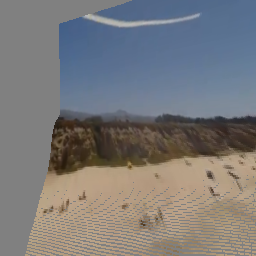}
    \includegraphics[width=0.105\textwidth]{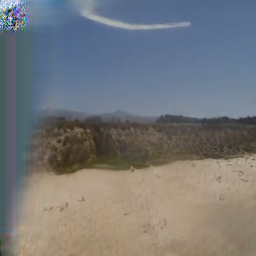}
    \includegraphics[width=0.105\textwidth]{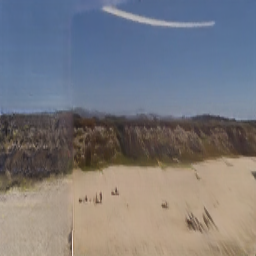}
    \includegraphics[width=0.105\textwidth]{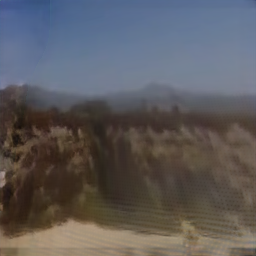}
    \includegraphics[width=0.105\textwidth]{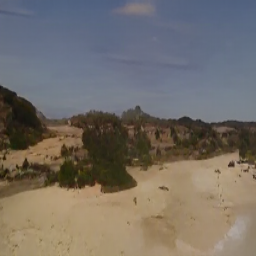}
    \includegraphics[width=0.105\textwidth]{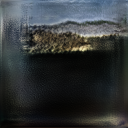}
    \includegraphics[width=0.105\textwidth]{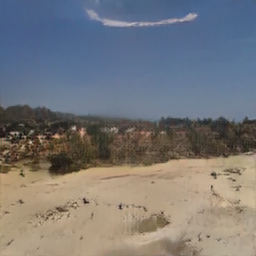}
    \includegraphics[width=0.105\textwidth]{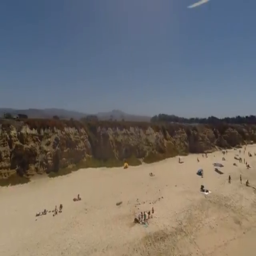}
   
    {\parbox{0.105\textwidth}{\centering {}}}
    {\parbox{0.105\textwidth}{\centering {\notsotiny Out-of-View(43\%)}}}
    {\parbox{0.105\textwidth}{\centering {\notsotiny PSNR-vis: 20.55}}}
    {\parbox{0.105\textwidth}{\centering {\notsotiny PSNR-vis: 13.87}}}
    {\parbox{0.105\textwidth}{\centering {\notsotiny PSNR-vis: 18.97}}}
    {\parbox{0.105\textwidth}{\centering {\notsotiny PSNR-vis: 18.94}}}
    {\parbox{0.105\textwidth}{\centering {\notsotiny PSNR-vis: 13.56}}}
    {\parbox{0.105\textwidth}{\centering {\notsotiny PSNR-vis: 20.91}}}
    {\parbox{0.105\textwidth}{\centering {}}}
    
    \includegraphics[width=0.105\textwidth]{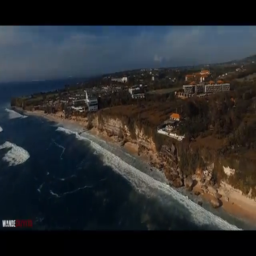}
    \includegraphics[width=0.105\textwidth]{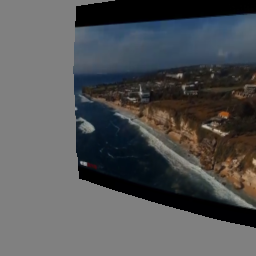}
    \includegraphics[width=0.105\textwidth]{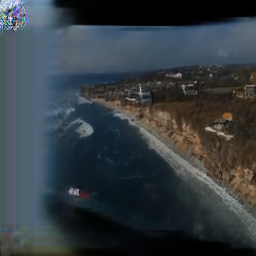}
    \includegraphics[width=0.105\textwidth]{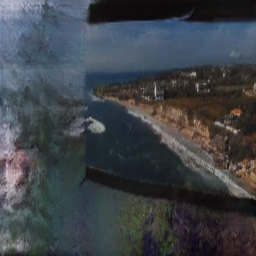}
    \includegraphics[width=0.105\textwidth]{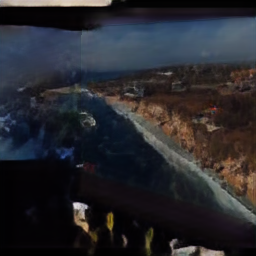}
    \includegraphics[width=0.105\textwidth]{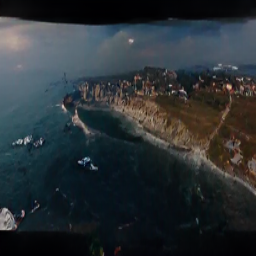}
    \includegraphics[width=0.105\textwidth]{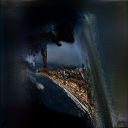}
    \includegraphics[width=0.105\textwidth]{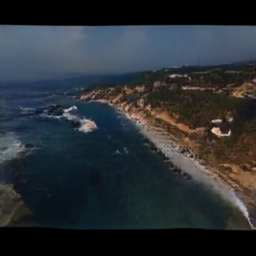}
    \includegraphics[width=0.105\textwidth]{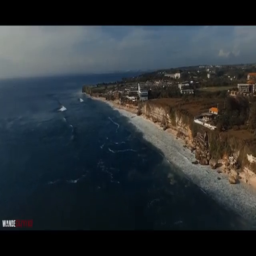}   

    {\parbox{0.105\textwidth}{\centering {}}}
    {\parbox{0.105\textwidth}{\centering {\notsotiny Out-of-View(64\%)}}}
    {\parbox{0.105\textwidth}{\centering {\notsotiny PSNR-vis: 18.64}}}
    {\parbox{0.105\textwidth}{\centering {\notsotiny PSNR-vis: 17.64}}}
    {\parbox{0.105\textwidth}{\centering {\notsotiny PSNR-vis: 18.46}}}
    {\parbox{0.105\textwidth}{\centering {\notsotiny PSNR-vis: 17.26}}}
    {\parbox{0.105\textwidth}{\centering {\notsotiny PSNR-vis: 6.79}}}
    {\parbox{0.105\textwidth}{\centering {\notsotiny PSNR-vis: 18.89}}}
    {\parbox{0.105\textwidth}{\centering {}}}
    
    \includegraphics[width=0.105\textwidth]{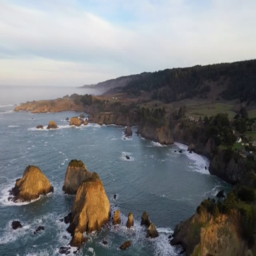}
    \includegraphics[width=0.105\textwidth]{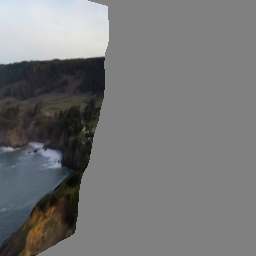}
    \includegraphics[width=0.105\textwidth]{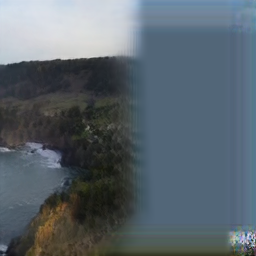}
    \includegraphics[width=0.105\textwidth]{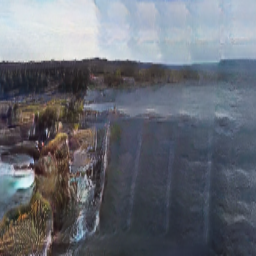}
    \includegraphics[width=0.105\textwidth]{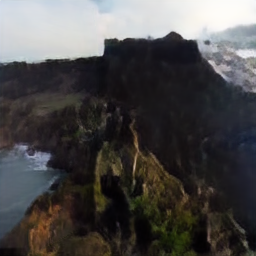}
    \includegraphics[width=0.105\textwidth]{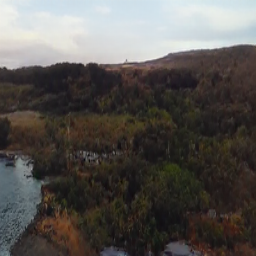}
    \includegraphics[width=0.105\textwidth]{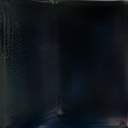}
    \includegraphics[width=0.105\textwidth]{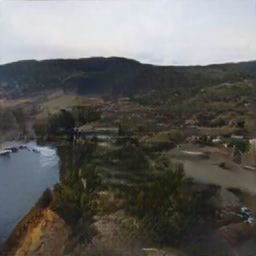}
    \includegraphics[width=0.105\textwidth]{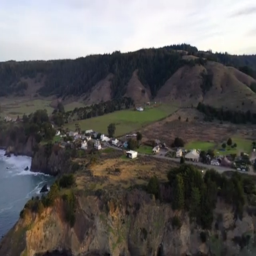} 

    {\parbox{0.105\textwidth}{\centering {}}}
    {\parbox{0.105\textwidth}{\centering {\notsotiny Out-of-View(69)}}}
    {\parbox{0.105\textwidth}{\centering {\notsotiny PSNR-vis: 16.78}}}
    {\parbox{0.105\textwidth}{\centering {\notsotiny PSNR-vis: 13.31}}}
    {\parbox{0.105\textwidth}{\centering {\notsotiny PSNR-vis: 14.77}}}
    {\parbox{0.105\textwidth}{\centering {\notsotiny PSNR-vis: 15.73}}}
    {\parbox{0.105\textwidth}{\centering {\notsotiny PSNR-vis: 6.10}}}
    {\parbox{0.105\textwidth}{\centering {\notsotiny PSNR-vis: 17.92}}}
    {\parbox{0.105\textwidth}{\centering {}}}
    
    \vspace{-.15cm}
    
    \subfigure[{\notsotiny Input Image}]{\includegraphics[width=0.105\textwidth]{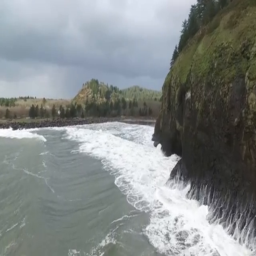}}
    \subfigure[{\notsotiny Warped Image}]{\includegraphics[width=0.105\textwidth]{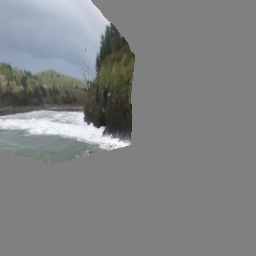}}  
    \subfigure[{\notsotiny SynSin~\cite{wiles2020synsin}}]{\includegraphics[width=0.105\textwidth]{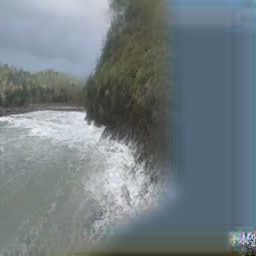}}
    \subfigure[{\notsotiny InfNat~\cite{liu2021infinite}}]{\includegraphics[width=0.105\textwidth]{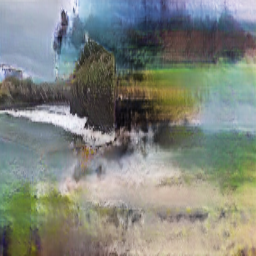}}
    \subfigure[{\notsotiny PixelSynth~\cite{rockwell2021pixelsynth}}]{\includegraphics[width=0.105\textwidth]{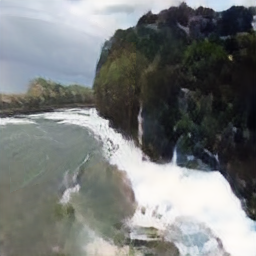}}
    \subfigure[{\notsotiny GeoFree~\cite{rombach2021geometry}}]{\includegraphics[width=0.105\textwidth]{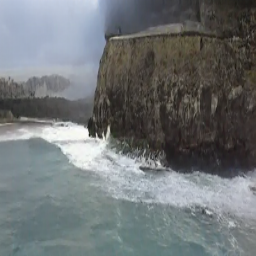}}
    \subfigure[{\notsotiny InfZero~\cite{li2022infinitenature}}]{\includegraphics[width=0.105\textwidth]{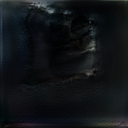}}
    \subfigure[{\notsotiny Ours}]{\includegraphics[width=0.105\textwidth]{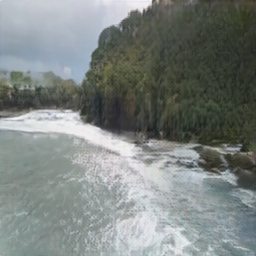}}
    \subfigure[{\notsotiny Ground Truth}]{\includegraphics[width=0.105\textwidth]{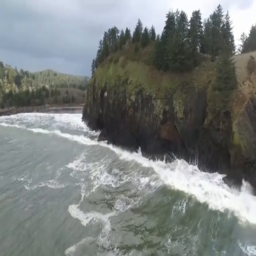}}

 \vspace{-.3cm}
    
    \caption{\textbf{Qualitative Results on ACID~\cite{liu2021infinite}.} For InfNat~\cite{liu2021infinite}, we report examples with higher PSNR-vis scores in either 1-step or 5-step variants.}
    \label{fig:qual_acid}
\end{figure*}

\begin{figure*}
\centering
\resizebox{1.0\textwidth}{!}{

\begin{tabular}{@{}cccccc@{}}
    \includegraphics[width=.4\linewidth, height=.4\linewidth]{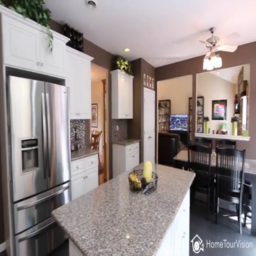} &
    \includegraphics[width=.4\linewidth, height=.4\linewidth]{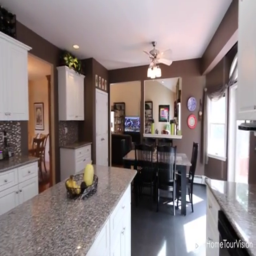} & 
    \multicolumn{1}{c|}{\includegraphics[width=.4\linewidth, height=.4\linewidth]{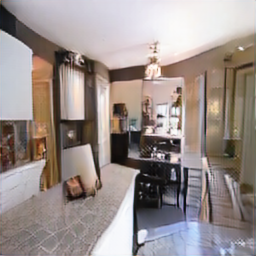}} & 
    \includegraphics[width=.4\linewidth, height=.4\linewidth]{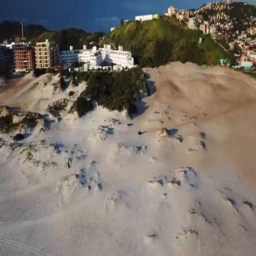} &
    \includegraphics[width=.4\linewidth, height=.4\linewidth]{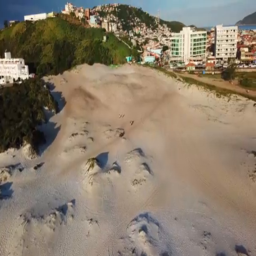} & 
    \includegraphics[width=.4\linewidth, height=.4\linewidth]{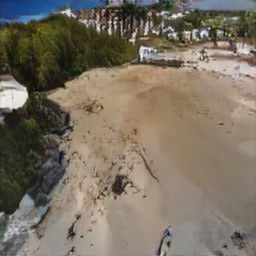} \\
        \includegraphics[width=.4\linewidth, height=.4\linewidth]{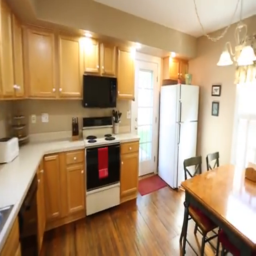} &
    \includegraphics[width=.4\linewidth, height=.4\linewidth]{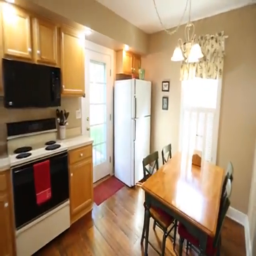} & 
    \multicolumn{1}{c|}{\includegraphics[width=.4\linewidth, height=.4\linewidth]{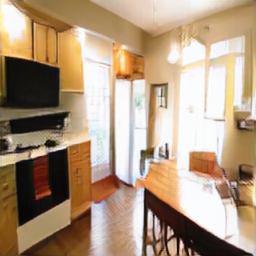}} & 
    \includegraphics[width=.4\linewidth, height=.4\linewidth]{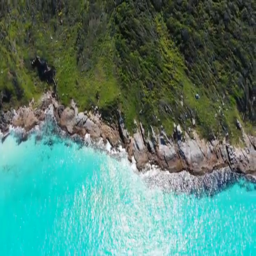} &
    \includegraphics[width=.4\linewidth, height=.4\linewidth]{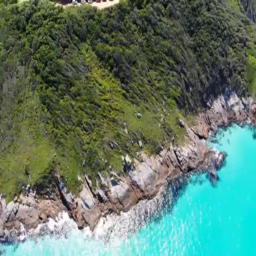} & 
    \includegraphics[width=.4\linewidth, height=.4\linewidth]{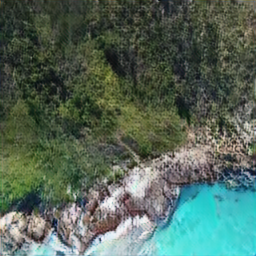} \\
        \includegraphics[width=.4\linewidth, height=.4\linewidth]{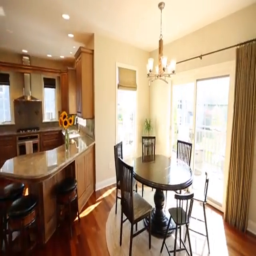} &
    \includegraphics[width=.4\linewidth, height=.4\linewidth]{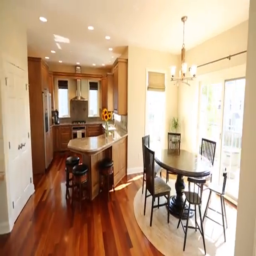} & 
    \multicolumn{1}{c|}{\includegraphics[width=.4\linewidth, height=.4\linewidth]{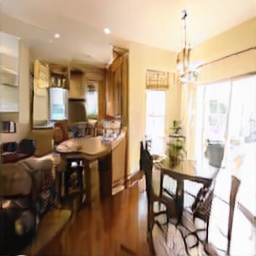}} & 
    \includegraphics[width=.4\linewidth, height=.4\linewidth]{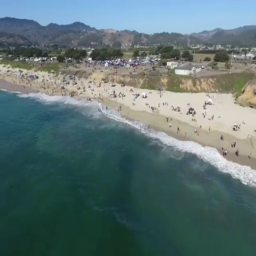} &
    \includegraphics[width=.4\linewidth, height=.4\linewidth]{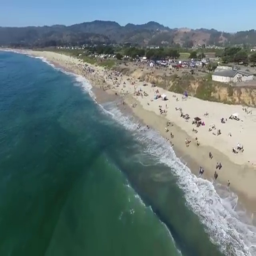} & 
    \includegraphics[width=.4\linewidth, height=.4\linewidth]{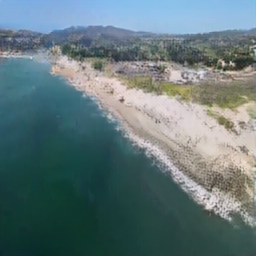} \\
        \includegraphics[width=.4\linewidth, height=.4\linewidth]{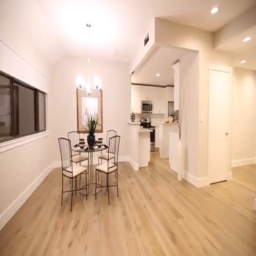} &
    \includegraphics[width=.4\linewidth, height=.4\linewidth]{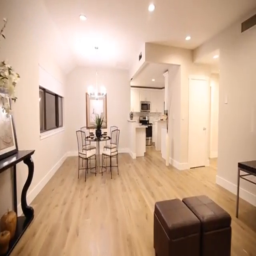} & 
    \multicolumn{1}{c|}{\includegraphics[width=.4\linewidth, height=.4\linewidth]{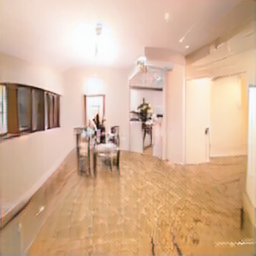}} & 
    \includegraphics[width=.4\linewidth, height=.4\linewidth]{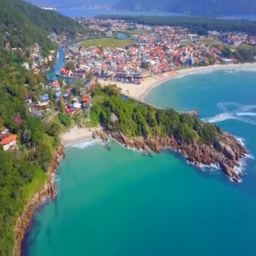} &
    \includegraphics[width=.4\linewidth, height=.4\linewidth]{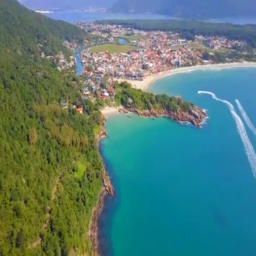} & 
    \includegraphics[width=.4\linewidth, height=.4\linewidth]{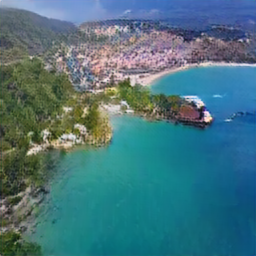} \\
        \includegraphics[width=.4\linewidth, height=.4\linewidth]{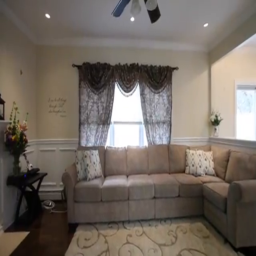} &
    \includegraphics[width=.4\linewidth, height=.4\linewidth]{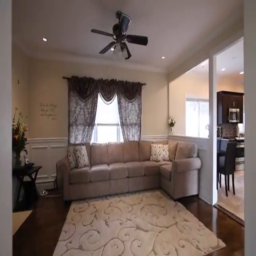} & 
    \multicolumn{1}{c|}{\includegraphics[width=.4\linewidth, height=.4\linewidth]{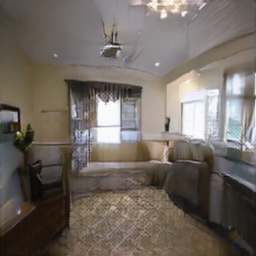}} & 
    \includegraphics[width=.4\linewidth, height=.4\linewidth]{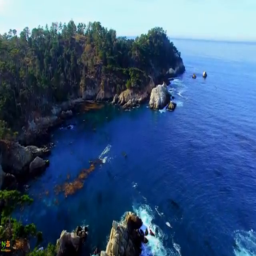} &
    \includegraphics[width=.4\linewidth, height=.4\linewidth]{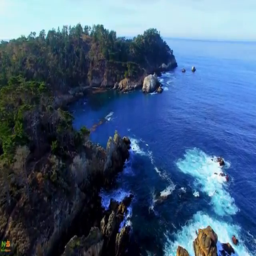} & 
    \includegraphics[width=.4\linewidth, height=.4\linewidth]{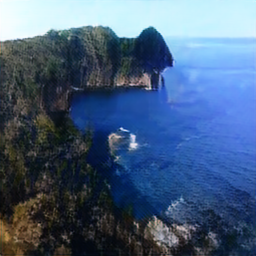} \\
        \includegraphics[width=.4\linewidth, height=.4\linewidth]{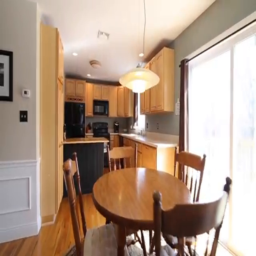} &
    \includegraphics[width=.4\linewidth, height=.4\linewidth]{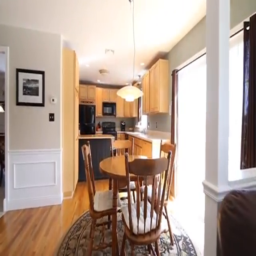} & 
    \multicolumn{1}{c|}{\includegraphics[width=.4\linewidth, height=.4\linewidth]{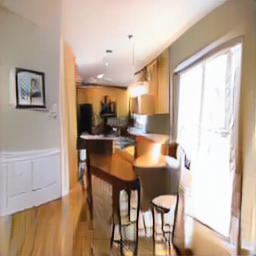}} & 
    \includegraphics[width=.4\linewidth, height=.4\linewidth]{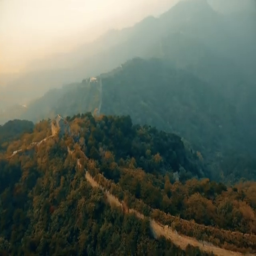} &
    \includegraphics[width=.4\linewidth, height=.4\linewidth]{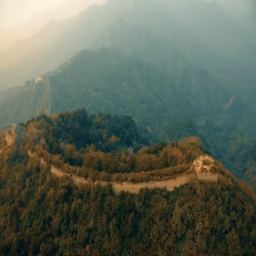} & 
    \includegraphics[width=.4\linewidth, height=.4\linewidth]{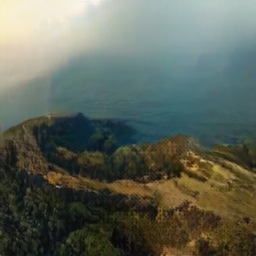} \\
            \includegraphics[width=.4\linewidth, height=.4\linewidth]{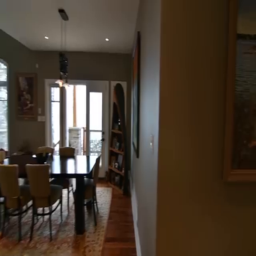} &
    \includegraphics[width=.4\linewidth, height=.4\linewidth]{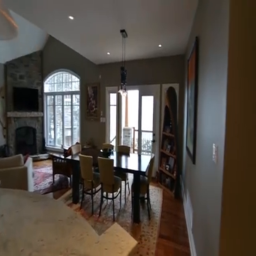} & 
    \multicolumn{1}{c|}{\includegraphics[width=.4\linewidth, height=.4\linewidth]{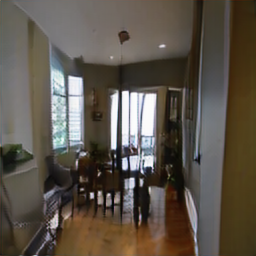}} & 
    \includegraphics[width=.4\linewidth, height=.4\linewidth]{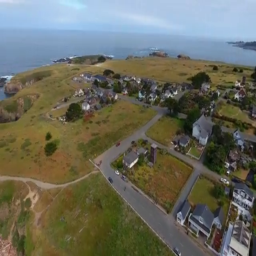} &
    \includegraphics[width=.4\linewidth, height=.4\linewidth]{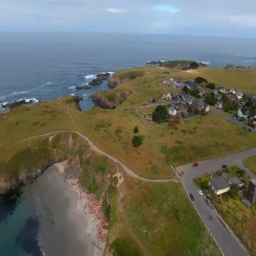} & 
    \includegraphics[width=.4\linewidth, height=.4\linewidth]{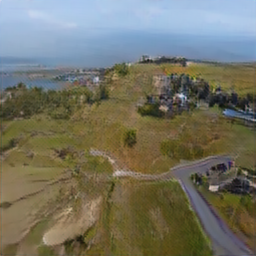} \\
            \includegraphics[width=.4\linewidth, height=.4\linewidth]{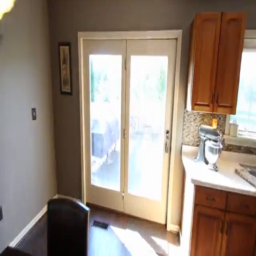} &
    \includegraphics[width=.4\linewidth, height=.4\linewidth]{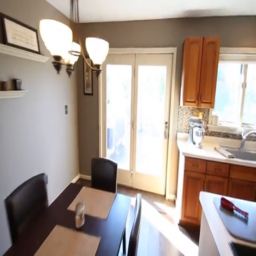} & 
    \multicolumn{1}{c|}{\includegraphics[width=.4\linewidth, height=.4\linewidth]{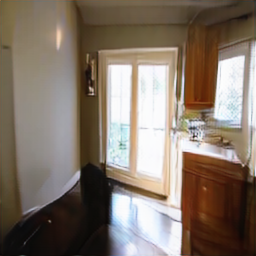}} & 
    \includegraphics[width=.4\linewidth, height=.4\linewidth]{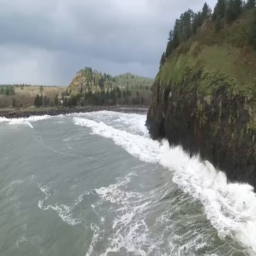} &
    \includegraphics[width=.4\linewidth, height=.4\linewidth]{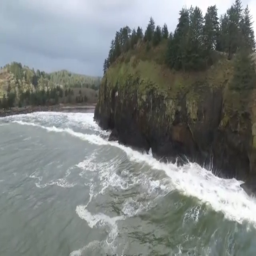} & 
    \includegraphics[width=.4\linewidth, height=.4\linewidth]{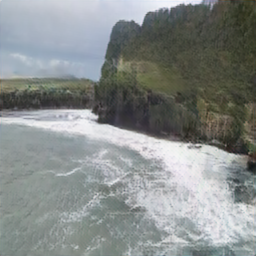} \\
    
    {\parbox{.4\linewidth}{\centering {\LARGE Input Images} }} &
    {\parbox{.4\linewidth}{\centering {\LARGE GT Images} }} &
    {\parbox{.4\linewidth}{\centering {\LARGE Generated Images}}} &
    {\parbox{.4\linewidth}{\centering {\LARGE Input Images} }} &
    {\parbox{.4\linewidth}{\centering {\LARGE GT Images} }} &
    {\parbox{.4\linewidth}{\centering {\LARGE Generated Images}}} \\ \\
    
    \multicolumn{3}{c}{{\parbox{1.0\linewidth}{\centering {\huge (a) RealEstate10K~\cite{zhou2018stereo}} }}} &
    \multicolumn{3}{c}{{\parbox{1.0\linewidth}{\centering {\huge (b) ACID~\cite{liu2021infinite}} }}} \\
  \end{tabular}
 }
\caption{\textbf{Additional Qualitative Results.}}
\label{fig:add_qual}
\end{figure*}

\section{Qualitative Results}

We further evaluate our method on different sizes of viewpoint changes as shown in Fig.~\ref{fig:qual_real} and Fig.~\ref{fig:qual_acid}.
We also visualize additional qualitative results in Fig.~\ref{fig:add_qual}.
Note that our method synthesizes novel views consistent with $I_{ref}$ and realistic out-of-view regions, regardless of the view change.

\newpage


\end{document}